\pdfoutput=1
\documentclass[11pt]{article}
\usepackage{authblk}
\usepackage[]{acl}

\usepackage{times}
\usepackage{latexsym}

\usepackage[T2A,T1]{fontenc}
\usepackage[utf8]{inputenc}
\usepackage[hindi,russian,english]{babel}
\usepackage{booktabs}
\usepackage{microtype}
\usepackage{amsmath}
\usepackage{graphicx}
\usepackage{subfigure}
\usepackage{amsfonts}
\usepackage{multirow}
\usepackage{listings}
\usepackage{url}
\usepackage[ruled,linesnumbered]{algorithm2e}  
\usepackage{CJKutf8}
\usepackage{graphicx}
\usepackage{subfigure}
\usepackage{supertabular}
\usepackage{subfigure}
\usepackage{dcolumn,booktabs}
\usepackage{tikz}
\usepackage[right]{rotating}

\title{Crosslingual Transfer Learning for Low-Resource Languages \\
Based on Multilingual Colexification Graphs}

\author[*$\diamond$]{\bf Yihong Liu}
\author[*$\diamond$]{\bf Haotian Ye}
\author[*$\diamond$]{\bf Leonie Weissweiler}
\author[*]{\bf Renhao Pei}
\author[*$\diamond$]{\bf Hinrich Sch\"utze}

\affil[*]{Center for Information and Language Processing, LMU Munich} \affil[$\diamond$]{Munich Center for Machine Learning (MCML)
 \protect\\ \texttt{\{yihong, yehao, weissweiler\}@cis.lmu.de}} 

\definecolor{eva}{HTML}{B990D4}
\definecolor{asuka}{HTML}{FC0006}
\definecolor{rei}{HTML}{84A8F1}

\definecolor{orange}{HTML}{F56600}
\definecolor{pink}{HTML}{F51893}
\definecolor{green}{HTML}{00A300}
\definecolor{blue}{HTML}{005BE0}
\definecolor{purple}{HTML}{A51CC4}
\definecolor{yellow}{HTML}{F5D000}
\definecolor{red}{HTML}{A30021}

\def\networkone{ColexNet\xspace}
\def\networktwo{ColexNet+\xspace}

\def\embname{$\overrightarrow{\mbox{\networktwo}}$\xspace}

\newcounter{notecounter}
\newcommand{\enotesoff}{\long\gdef\enote##1##2{}}
\newcommand{\enoteson}{\long\gdef\enote##1##2{{
\stepcounter{notecounter}
{\large\bf \hspace{1cm}\arabic{notecounter} $<<<$ ##1: ##2 $>>>$\hspace{1cm}}}}}
\enoteson
 \enotesoff

\def\secref#1{\S\ref{sec:#1}}
\def\seclabel#1{\label{sec:#1}}

\begin{document}
\maketitle

\begin{abstract}

% didn't use macro in the abstract because it is easier to paste in submission.
In comparative linguistics, colexification refers to the phenomenon of a lexical form conveying two or more distinct meanings. Existing work on colexification patterns relies on annotated word lists, limiting scalability and usefulness in NLP. In contrast, we identify colexification patterns of more than 2,000 concepts across 1,335 languages directly from an unannotated parallel corpus. We then propose simple and effective methods to build multilingual graphs from the colexification patterns: \textbf{ColexNet} and \textbf{ColexNet+}. ColexNet's nodes are concepts and its edges are colexifications. In ColexNet+, concept nodes are additionally linked through intermediate nodes, each representing an ngram in one of 1,334 languages. We use ColexNet+ to train $\overrightarrow{\mbox{ColexNet+}}$, high-quality multilingual embeddings that are well-suited for transfer learning. In our experiments, we first show that ColexNet achieves high recall on CLICS, a dataset of crosslingual colexifications. We then evaluate $\overrightarrow{\mbox{ColexNet+}}$ on roundtrip translation, sentence retrieval and sentence classification and show that our embeddings surpass several transfer learning baselines. This demonstrates the benefits of using colexification as a source of information in multilingual NLP.

\end{abstract}

\begin{figure}
    \setlength{\belowcaptionskip}{-0.5cm}
  \centering
  \includegraphics[width=0.48\textwidth]{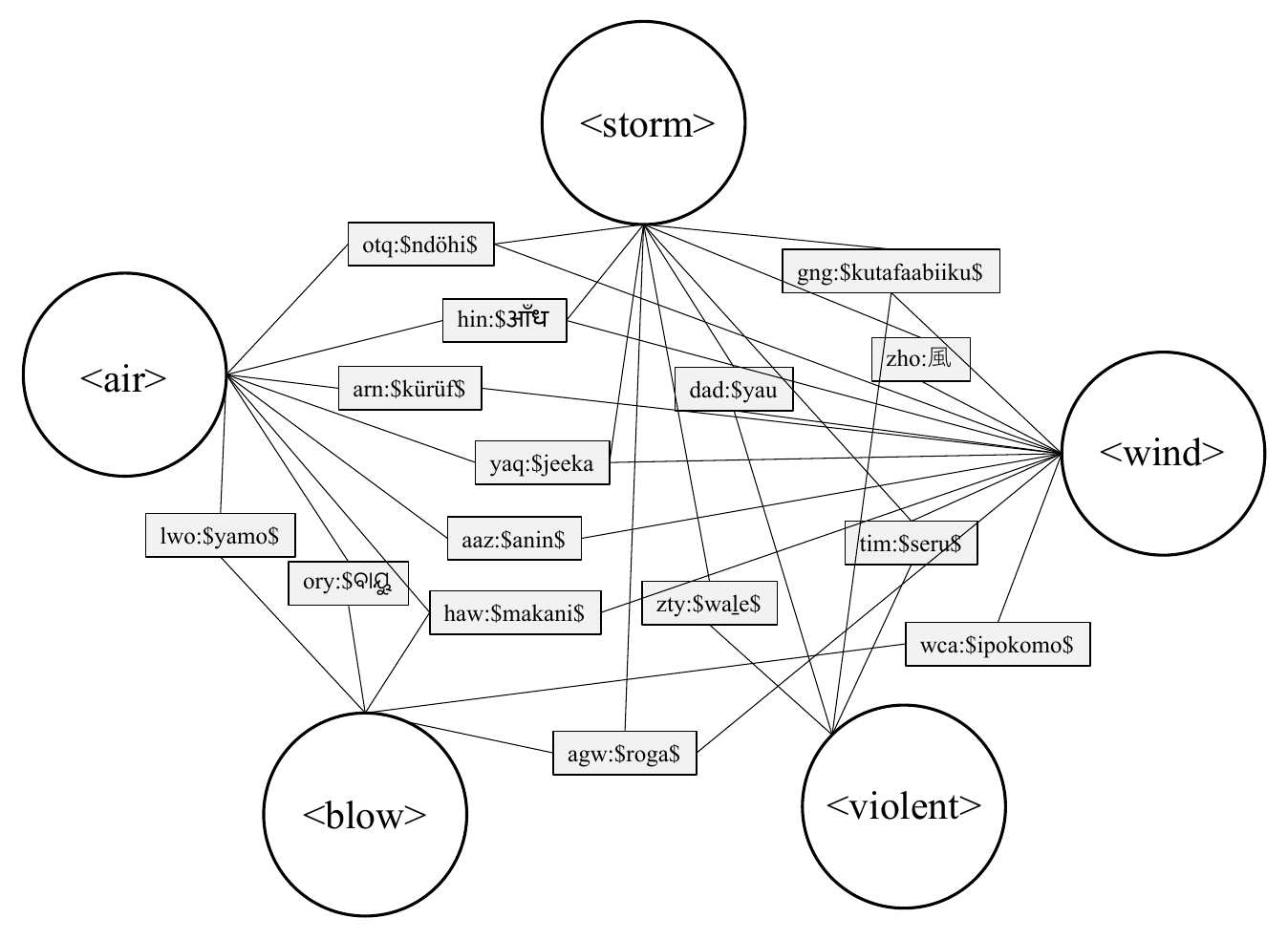}
  \caption{A subgraph of \networktwo. Circles (<air>,
  <storm>, \ldots): concept nodes. Rectangles
  (otq:\$nd\"{o}hi\$, lwo:\$yamo\$, \ldots): ngram nodes (each prefixed by its ISO
  693-3 code). Ngram nodes
  realize colexifications, e.g., <air> and <storm> are
  linked, through translations in the parallel corpus, to
the  Querétaro Otom\'{\i} ngram ``nd\"{o}hi'' (i.e., otq:\$nd\"{o}hi\$).}
  \label{fig:first_page}
\end{figure}

\section{Introduction}\seclabel{intro}

Multilingual representations are beneficial
in natural language processing (NLP) due to
their ability to transfer knowledge across
languages \citep{artetxe-schwenk-2019-massively,conneau-etal-2020-unsupervised,fan2021beyond}. Typically,
such representations are learned through pre-training Large
Language Models
(LLMs) \citep{brown2020language,chowdhery2022palm,touvron2023llama}
or multilingual word
embeddings \citep{ammar2016massively,lample2018muse,dufter-etal-2018-embedding}.
However, LLMs require enormous amounts of data to train,
limiting their use mostly to high-resource and
medium-resource languages \citep{zhou2023comprehensive}.
Alternatively, multilingual word embeddings are widely used
in NLP because of their simplicity and good
performance \citep{ammar2016massively,lample2018muse,jawanpuria-etal-2019-learning}.
However, most existing multilingual embeddings are learned
through \emph{word-context} information, without leveraging
global cooccurrence information in individual languages or
across languages, which can help distinguish distinct meanings
conveyed by a lexical form. Therefore, we see a pressing
need in NLP for massively multilingual word embeddings that
span a large number of languages (1,335 in our case)
and that specifically account for
global occurrence and are a good basis for
crosslingual transfer learning.

\emph{Colexification} has gained increasing attention in
comparative linguistics and crosslingual NLP.  According
to \citet{franccois2008semantic}, a
language \emph{colexifies} two distinct meanings if 
it expresses them with the same lexical
form. Different languages have
different \emph{colexification patterns}. For example,
while English
has separate words for <hand> and <arm>\footnote{We
represent a concept (from any language) by surrounding an English word that refers to the concept with ``<>''.},
Russian `\foreignlanguage{russian}{рукa}' colexifies these two
concepts.
Most prior work
explores colexification
\citep{floyd2021modeling,brochhagen2022languages,list2023inference}
using
manually curated crosslingual datasets
that consist of multilingual word lists such as
CLICS \citep{list_johann_mattis_2018_1194088,
list2018clics2,rzymski2020database}. However, 
relying on these datasets has several limitations: 
extension to more languages and more concepts can be challenging; 
these datasets contain lists of lemmata and (in a corpus-based
approach for low-resource languages without morphological
resources)
cannot easily be used for the processing of
occurrences in context.

To overcome these limitations and boost 
crosslingual transfer learning especially 
for low-resource languages, we use the
\textbf{P}arallel \textbf{B}ible \textbf{C}orpus
(PBC) \citep{mayer-cysouw-2014-creating}, which has
verse-level aligned translations of the Bible in 1,335 languages, 
to identify colexification patterns (a verse in PBC roughly
corresponds to a sentence). 
With the identified patterns between a wide
range of concepts, we propose novel  algorithms
that efficiently build large-scale multilingual graphs.
To the best of our knowledge, this is the first work that
constructs graphs of colexification
and trains multilingual representations for crosslingual transfer learning
directly from a parallel corpus on a large scale.
We show that the
graphs capture the links between concepts across
languages  and that
the derived multilingual representations
considerably improve crosslingual transfer
on downstream tasks. 
Previous work on building monolingual
graphs \citep{jauhar-etal-2015-ontologically,ustalov-etal-2017-watset}
or multilingual
graphs \citep{harvill-etal-2022-syn2vec,jafarinejad2023synset2node,chen2023colexlang}
is different
as it (1) does not consider words in context and only uses
lemmata, (2) is based on external sense
inventories such as WordNet \citep{miller1995wordnet} and
BabelNet \citep{navigli2012babelnet,ijcai2021p620}, which
are  not available for many low-resource languages,
and (3) does not investigate the crosslingual
transferability of the multilingual
representations on NLP downstream tasks such as
sentence retrieval or classification in a crosslingual scenario.

The contributions of this work are as follows: (i) We
present \networkone, a graph of concepts based on
colexification patterns that are directly extracted from a
parallel corpus.  (ii) By extending \networkone, we further
present \networktwo, a large-scale multilingual graph that
additionally contains ngrams in 1,334 languages 
that instantiate those patterns. (iii) We
contribute to crosslingual transfer learning, by
using \networktwo to generate multilingual
embeddings: \embname. We show that \embname outperforms
several baselines on roundtrip translation, verse retrieval,
and classification.  (iv) We evaluate
\networkone on CLICS and show that we
identify a large portion of the ground-truth
colexifications. (v) Going beyond many works on crosslingual
transfer that focus on transfer from English,
we systematically investigate the effect of the source
language on successful transfer with \embname:
we use 1,245 languages as sources and experiment on 1,245 $\times$
1,245 transfer directions.  (vi) We make our code, graphs, and
embeddings publicly available.\footnote{\url{https://github.com/cisnlp/ColexificationNet}}

% (v) We conduct a detailed
% graph-based analysis of \networkone and show that
% language family and geographic area influence the
% community structure.

\section{Related Work}\seclabel{related_works}
% \subsection{Multilingual word embeddings}
There are many ways to learn multilingual word
embeddings. One common way is to first learn monolingual
embeddings on each language separately through, e.g.,
Word2Vec \citep{word2vec2013miklov},
GloVe \cite{pennington-etal-2014-glove}, or
fastText \citep{bojanowski-etal-2017-enriching}, and then
map them into the same
space \citep{artetxe-etal-2017-learning,lample2018muse,artetxe-etal-2018-robust}. Another
group of methods uses parallel corpora 
% (sentence-level aligned)
to directly learn bilingual
embeddings \citep{hermann-blunsom-2014-multilingual,chandar2014bilingual-word-representations,levy-etal-2017-strong}.
Our work is related
to that of \citet{dufter-etal-2018-embedding}, which also learns
embeddings on the PBC, but does not take advantage of
colexification, i.e., 
the explicit modeling of relations between
colexified concepts/ngrams. We use
S-ID \citep{levy-etal-2017-strong} and embeddings
from \citet{dufter-etal-2018-embedding} as baselines.

One of the best-known and widely used multilingual resources
is BabelNet \citep{navigli2012babelnet,ijcai2021p620}. BabelNet has been used for learning or enhancing
embeddings \citep{iacobacci-etal-2015-sensembed,jose2018senseembeddings,conia-navigli-2020-conception,levine-etal-2020-sensebert,harvill-etal-2022-syn2vec,chen2023colexlang} for lexical-level tasks such as semantic word similarity and word sense
disambiguation \citep{speer-lowry-duda-2017-conceptnet,conia-navigli-2020-conception,Procopio2021MultiMirror,navigli2022babelnet}. Our
focus is on the coverage of many more languages 
(i.e., larger scale in terms of languages) for crosslingual transfer learning. While
hand-curated lexica often have better quality
than automatically learned resources, they are not available
for most of our languages. Ultimately, the two approaches
should be combined.

Colexification was introduced
by \citet{haspelmath2003geometry} in the context of
grammatical semantics. \citet{franccois2008semantic} then
used colexification as the foundation for studying semantic
change crosslinguistically.
CLICS \citep{list_johann_mattis_2018_1194088,
list2018clics2,rzymski2020database} is a crosslingual database
that facilitates research on colexification. Languages can differ in their colexification patterns, which are influenced by many factors such as human cognition, language family, and geographic area \citep{jackson2019emotion,xu2020conceptual,segerer2022areal}. An empirical study by \citet{bao-etal-2021-universal} indicates that no pair of concepts is colexified in every language. On the other hand, a recent investigation on conceptualization from the PBC shows some concepts are more likely to be involved in colexification than others \cite{liu-etal-2023-crosslingual}. Such universal colexification patterns across languages reflect crosslinguistic similarities \citep{youn2016universal,georgakopoulos2022universal}. Therefore, by integrating colexification patterns of as many languages as possible, we can generate multilingual representations that are suitable for massively crosslingual transfer learning.

\section{Methodology}\seclabel{method}

\subsection{Data}
We use 1,335 Bible translations from the PBC corpus \citep{mayer-cysouw-2014-creating}. Each translation is from a different language (identified by its ISO 639-3 code). 
Prior work \citep{asgari-schutze-2017-past,
dufter-etal-2018-embedding, weissweiler-etal-2022-camel} has
used subsets of the corpus. In contrast,
we follow Conceptualizer \citep{liu-etal-2023-crosslingual} and
use all parallel verses between English and other languages.
This gives us better coverage of concepts and the contexts in which they occur. 

\subsection{Colexification pattern identification}

\paragraph{Concept Pool.} Conceptualizer \citep{liu-etal-2023-crosslingual} uses a small manually selected group of \textbf{focal concepts}, i.e., concepts of interest (83 in total) and constructs a set of strings to represent each concept. For example, it uses \{\texttt{\$belly\$}, \texttt{\$bellies\$}\} to represent the focal concept <belly>, where \texttt{\$} is the word boundary. Manually defining the sets is not feasible when a large number of concepts are to be explored. Thus, in this work, we lemmatize the English corpus and regard each lemma as a concept. The set of all lemmata forms the concept pool $F$. 

\paragraph{Conceptualizer.}
Conceptualizer \citep{liu-etal-2023-crosslingual} creates a
bipartite directed alignment graph between source language
concepts and target language strings. It
consists of a forward pass (FP) and a backward pass (BP).
This kind of two-step workflow is also used in extracting semantic relations \citep{Dyvik2004semantic} and paraphrases \citep{bannard-callison-burch-2005-paraphrasing}
from bilingual parallel corpora. A key difference compared with
this prior work is that Conceptualizer
works on the ngram level instead of the token level; this
facilitates the extraction of any associations hidden inside words. 
In Conceptualizer, 
FP first searches for target language ngrams highly associated
with a given focal concept; BP then searches for English
ngrams highly correlated with the target ngrams identified
in FP. The association is measured using $\chi^2$ score.
The process can detect if the conceptualization of
the focal concept diverges in any language. For example,
starting from concept <hand>, FP finds the Russian ngram
`\foreignlanguage{russian}{рук}', and BP then finds two
English ngrams `hand' and `arm'. This indicates that the
conceptualization of these concepts diverges in English and
Russian. The divergence of
conceptualization in the lexical forms indicates
a difference in their colexification patterns: Russian
colexifies the concepts <hand> and <arm> (in the word
`\foreignlanguage{russian}{рук}') while English does not.

\paragraph{Forward Pass.} Let $f$ be a focal concept
 in $F$ and $V_f$ the set of verses in which $f$ occurs.
 FP identifies a set
 of ngrams $T$ in target-language $l$ where each ngram can
 refer to concept $f$,
 i.e., $T = \text{FP}(f, l)$. We exhaustively search all
 ngrams $t$ within all tokens\footnote{Similar to the setting in Conceptualizer \citep{liu-etal-2023-crosslingual}, we use \texttt{\$} to
 denote token boundaries;
 e.g., \texttt{\$k}, \texttt{\$ke}, \texttt{\$ke\$},
 \texttt{k}, \texttt{ke}, \texttt{ke\$}, \texttt{e}, 
 \texttt{e\$}
 are all valid ngrams of token $\texttt{\$ke\$}$.} 
 in the
 parallel corpus in target language $l$ for high correlation
 with $V_f$. This procedure is similar to
\citet{ostling2023language}'s  subword-level
 alignment, but
we align concepts in English
 and subwords in other target languages. E.g., we
 start from <hand> and find that the Russian
 ngram `\foreignlanguage{russian}{рук}' has the highest
 correlation with $V_{\text{<hand>}}$, which indicates
 `\foreignlanguage{russian}{рук}' can refer to 
 <hand>. Like Conceptualizer, we use $\chi^2$ as a measure
 of correlation and iterate FP
 until the cumulative coverage  $T = \bigcup t$ of focal
 concept $f$ exceeds
 a threshold $\alpha=0.9$, but for a
 maximum of
   $M=3$ iterations.
 See \secref{hyperparam} for a discussion of these hyperparameters.

 \paragraph{Backward Pass.} BP is essentially
 the same as FP, but the search direction is reversed. Let
 $V_T$ be the set of verses in which at least one ngram in
 $T$ (identified in FP for target language $l$ and concept
 $f$) from target language $l$ occurs. We exhaustively
 search all concepts $c$ from the concept pool $F$ for high
 correlations with $V_T$. Let $C = \text{BP}(T, l)$ be the final set 
 of identified concepts. If
 $|C|=1$ and $c \in C \wedge c = f$, this indicates the
 ngrams can only refer to the concept $f$ according to the
 bilingual context. Alternatively, if $|C| > 1$, this indicates language $l$
 colexifies concepts in $C$ through ngrams $T$. For example,
 by performing BP on ngram `\foreignlanguage{russian}{рук}',
 we get <hand> and <arm> as the result, which
 indicates Russian colexifies <hand> and <arm>. 
 Notably,
 since we consider ngrams instead of tokens on the target
 language side, this allows us to also identify \emph{partial
 colexification} patterns in BP, i.e., patterns that do not 
 involve an entire word, but rather part of it. 
 We show such examples in \secref{investigation}. 
 % `\begin{CJK*}{UTF8}{bsmi}
 % 天\end{CJK*}' is identified in FP when initializing the
 % search for the concept <day> and we find <heaven> in BP
 % since Chinese \emph{partially} colexifies <day> and
 % <heaven> through the word `\begin{CJK*}{UTF8}{bsmi}天
 % 堂\end{CJK*}' (heaven).
 
\subsection{\networkone}                
We run FP and BP for
all 1,806 focal concepts in the English concept pool $F$ that have frequency
between 5 and 2000 and for
every language $l$ in our set of
1,334 target languages $L$ (excluding English). 
This allows us to uncover the colexification
patterns in 1,334 languages. We formalize the relations of
the colexification patterns as an undirected graph, where each node
is a concept represented by an English lemma and each edge
indicates that at least one language colexifies the two connected 
concepts. Formally, let $\mathcal{G}(F, \mathcal{E}, w_c, w_n)$ be
a weighted  undirected graph on vertices $F$, i.e., the
concept pool, where $\mathcal{E}$ is a set of undirected
edges; $w_c$ is an edge weighting (counting) function:
$F \times F \rightarrow \mathbb{Z}_{+}$, which returns,
for a pair of concepts,
the number of languages colexifying them; 
$w_n$ is an edge record function,
which returns all ngrams that colexify a given pair of concepts.
We show the graph construction in
Algorithm \ref{alg:concept}.

\begin{algorithm}
    \caption{\networkone \& \networktwo} \footnotesize \label{alg:concept} \KwIn{set of languages $L$, concept pool $F$, minimum
    number of languages threshold
    $\lambda$\;} \KwOut{\networkone
    $\mathcal{G}_1$, \networktwo $\mathcal{G}_2$\;}
    $\mathcal{G}_1 \leftarrow$ graph with $F$ as nodes and
    no edges $\mathcal{E}_1$\; $\mathcal{G}_2 \leftarrow$
    graph with no nodes $V$ and no edges $\mathcal{E}_2$\;
    $V \leftarrow F$\;
    set $w_c(\cdot) = 0 $ and $w_n(\cdot) = \emptyset $ by default\;
    \For{$l \in L $} { \For{$f \in F$} {
    $T \leftarrow \text{FP}(f,l)$\;
    $C \leftarrow \text{BP}(T,l)$\; $V \leftarrow V \cup
    T$\; \For{$c \in C$} {
    $\mathcal{E}_1 \leftarrow \mathcal{E}_1 \cup (f,
    c)$\;  $w_c((f, c)) += 1$\;
    $w_n((f, c)) \leftarrow w_n((f, c)) \cup T$\; } } }
    \For{$e \in \mathcal{E}_1$}
    { \If{$w_c(e) < \lambda$} { remove $e$ from
    $\mathcal{E}_1$\;
    } }
    \For{$e \in \mathcal{E}_1$}
    {
    $(f_1, f_2) \leftarrow e$\;
    \For{$v \in w_n((f_1, f_2))$}
    { $\mathcal{E}_2 \leftarrow \mathcal{E}_2 \cup (f_1, v) \cup
    (v, f_2)$\;}
    }
    remove nodes in
    $\mathcal{G}_1$ and $\mathcal{G}_2$ that have zero
    degree\; \KwRet{$\mathcal{G}_1$, $\mathcal{G}_2$}
    %\end{algorithmic}  
\end{algorithm}
% \vspace{-0.2cm}

In this study, we use a threshold $\lambda$ to control the
confidence of the colexification edges:
we remove an edge $e$ if $w_c(e) < \lambda$.
The intuition is that: if two concepts 
$f_1$ and $f_2$ are colexified in many languages,
we can be more certain that the edge between $f_1$ and $f_2$
is correctly identified. Looking at it the other way around,
if two concepts are only colexified in a few languages, this
might be a wrongly identified pattern because of verse-level
misalignment, free translation, or other errors in the data. See
Table \ref{tab:basic_stats} for the influence of different
$\lambda$ on graph statistics. In addition, we remove
zero-degree nodes to filter out isolated concepts. 

\subsection{\networktwo}

\networkone
only contains concepts that are expressed in English lemmata
and cannot be directly used to learn
multilingual representations for the target
languages. Therefore, we propose \networktwo, a large
multilingual graph expanded from \networkone by including
target language ngrams  that instantiate the
colexification patterns identified
in \networkone. Specifically, we replace each edge $(f_1,
f_2)$ with a set of new pairs of edges: (1) find the set of
ngrams $w_n((f_1, f_2))$ that colexify concepts $f_1$ and
$f_2$ (in any language) and (2) for each ngram $v$ in the set, insert new
edges $(f_1, v)$ and $(v, f_2)$.
To define a clean bipartite structure,
we do not include the
original $(f_1, f_2)$; this guarantees that
only concept-ngram
edges and no
concept-concept edges occur. In addition, any two related
concepts (i.e., there is
an edge connecting the two concepts in \networkone) are always implicitly
connected through ngram nodes in \networktwo that
associate the two concepts. 
Figure \ref{fig:first_page} shows a subnetwork
of \networktwo consisting of a few concepts and ngrams in
different languages that colexify them.
The graph construction is shown
in Algorithm \ref{alg:concept}.

As \networktwo is expanded from \networkone, this allows us
to only include pairs of edges 
expanded from reliable edges 
($w_c(e) \ge \lambda$) 
in \networkone.
% Similar to \networkone, we include the new edges between
% two concepts if and only if the two concepts are
% colexified in  at least $\lambda$ languages. 
The number of nodes and
edges included in \networktwo is thus influenced by
$\lambda$. The higher $\lambda$, the fewer nodes and edges
will be in \networktwo. \secref{hyperparam} presents
statistics and performance for different values of
$\lambda$.

\subsection{Multilingual Embedding learning}

% \networktwo is a large multilingual graph of concepts and ngrams in
% 1,335 languages, so it is natural to use representation
% learning methods on graphs to generate multilingual
% embeddings that benefit crosslingual transfer. 
To capture the semantic relations among the nodes and the structure of
\networktwo, we use Node2Vec \citep{grover2016node2vec} to
generate node representations.
Let $v$ be the node that a random walk currently resides in,
$t$ the node that the walk has traversed in the last step, 
and $x$ the node that the walk will visit in the next step.
Node2Vec calculates the
unnormalized transition probability from $v$ to $x$ 
as $\pi_{vx} = \alpha_{pq}(t, x) \cdot w((v, x))$ for
sampling the next node $x$ in the graph, where $w((v, x))$ is the
weight of the undirected edge $(v, x)$,
\begin{equation*}
\alpha_{pq}(t, x)=
\begin{cases}
\frac{1}{p} & \text{if } d_{tx} = 0\\
1 & \text{if } d_{tx} = 1 \\
\frac{1}{q} & \text{if } d_{tx} = 2\\
\end{cases}
\end{equation*}
and $d_{tx}$ is the shortest path distance between $t$ and $x$. 
The transition probability determines 
if either a new node or 
an already-visited node (regardless of a concept or ngram node) 
will be sampled. 
In \networktwo, $d_{tx} \neq 1$ for any nodes $t$ and $x$,
because a concept (resp. ngram) node will not connect 
with other concept (resp. ngram) nodes.
We set return parameter $p=.5$ and in-out parameter $q=2$ in the hope of encoding more ``local'' information, 
as this setting approximates breadth-first sampling according to \citet{grover2016node2vec}.

Below, we show that the multilingual representations trained this way
have some desirable properties, e.g., representations of ngrams from different languages that refer to the same concept can be highly cosine-similar, which is important for zero-shot crosslingual transfer learning. 

\section{Experiments}\seclabel{experiments}

To evaluate our proposed methods, we conduct the following
experiments: (1) colexification identification; (2) roundtrip translation; (3) verse retrieval;
and (4) verse classification. Experiment (1)
evaluates the colexification patterns we identify
in \networkone. Experiments (2), (3) and (4) evaluate the learned
multilingual embeddings \embname.

\subsection{Baselines}
To evaluate the effectiveness of
\embname,
our multilingual embeddings, we consider several previously proposed strong multilingual embeddings
as baselines for downstream tasks. The dimension of all
embeddings (ours and the baselines) is set to 200 for a fair
comparison. In addition, we consider three non-embedding
baselines: bag-of-words (BOW),
XLM-R \citep{conneau-etal-2020-unsupervised} and
Glot500-m \citep{imanigooghari-etal-2023-glot500}. The first
is a random baseline and is expected to perform the worst
because a BOW model is only trained on the English corpus,
which does not directly transfer to other languages. The
latter two are strong multilingual pretrained models. XLM-R is pretrained on 100 languages while
Glot500-m is a continued-pretrained version of XLM-R on the Glot500-c corpus \citep{imanigooghari-etal-2023-glot500} that includes more than 500 languages. We choose the base version of these multilingual pretrained models. We introduce the embedding baselines below.

\paragraph{S-ID embedding.}
\citet{levy-etal-2017-strong}  show that S-ID embeddings,
which leverage the sentence ID feature, are effective in
learning good multilingual embeddings from parallel
corpora. We use pairs of a verse ID identifier and a
token in this verse as input to
Word2Vec \citep{word2vec2013miklov} to train S-ID
embeddings. For example,, the pairs  (01049027, \texttt{Wolf}) and
(01049027, \begin{CJK*}{UTF8}{bsmi}狼\end{CJK*}) will be
presented in the data because
`\begin{CJK*}{UTF8}{bsmi}狼\end{CJK*}' (resp.\ `Wolf')
occurs in Chinese (resp.\ German) in verse number
01049027. This is a strong baseline because the verse number
is an abstract representation of the context. Therefore it
encourages words occurring across languages in the same verse to have similar embeddings.

\paragraph{CLIQUE \& $N(t)$ embedding.}
CLIQUE embeddings \citep{dufter-etal-2018-embedding} are
learned on cliques extracted from  PBC. Each clique is a set
of tokens from different languages that refer to the same
concept. The embeddings are then learned from token-clique
pairs.
Additionally, to take the connections between concepts into account, \citet{dufter-etal-2018-embedding} consider the neighbors (tokens that are connected with the current node in the dictionary graph) of each token and train embeddings on those pairs of neighbors, which we refer to as $N(t)$ embedding.

\paragraph{Eflomal-aligned embedding.} 
We construct an alignment graph of words by using
Eflomal \citep{ostling2016efficient} and learn embeddings on
the graph as another strong baseline. Specifically, we align
the English Bible with Bibles in all other target
languages.
We define the edge set of the graph  as the set of all
edges that connect an English word with its aligned target
language word (if there are at least two such alignments).
Finally, we use Node2Vec (same hyperparameters as for \networktwo) 
to learn multilingual embeddings. 

\subsection{Colexification identification}\seclabel{colex_identification}
We first evaluate how well \networkone performs at
identifying colexification patterns. 
We use CLICS \citep{list_johann_mattis_2018_1194088,
list2018clics2,rzymski2020database},
a database of colexifications, as the gold
standard. Each node in CLICS is a concept expressed in
English. In \networkone,
we use English lemmata as expressions of concepts whereas
CLICS also includes short noun phrases.
We only consider the common concepts, i.e., concepts that
are expressed as English words and occur in both CLICS
and \networkone. For each start concept $s$ in the common
concepts $P$, let $T(s)$ be the neighbors in CLICS, i.e., a
set of concepts that have a colexification relation with $s$ and $C(s)$ be the neighbors in \networkone. Then we compute the recall for $s$ as ${| T(s) \cap  C(s)|}/{| T(s)|}$. To have a global view of the performance, we report the micro average recall (MicroRec.): 
\setlength{\abovecaptionskip}{0.1cm}
$$\text{MicroRec.} = \frac{\sum_{s \in P} |T(s) \cap C(s)|}{\sum_{s \in P} |T(s)|}$$
\setlength{\belowcaptionskip}{-0.2cm}
macro average recall (MacroRec.): 
\setlength{\abovecaptionskip}{0.1cm}
$$\text{MacroRec.} = \sum_{s \in P} \frac{|T(s) \cap C(s)|}{|T(s)| |P|}$$
\setlength{\belowcaptionskip}{-0.2cm}
and the average number of incorrect (not present in CLICS)
colexifications per concept (\#aw\_colex): 
\setlength{\abovecaptionskip}{0.1cm}
$$\text{\#aw\_colex} = \sum_{s \in P}\frac{ |C(s) - T(s)|}{|P|}$$
\setlength{\belowcaptionskip}{-0.2cm}
\#aw\_colex has a similar function to precision. We do not
use precision as a measure because it can underestimate the
performance, as many patterns included in \networkone can
actually be correct (see \secref{investigation} for
examples). The \#aw\_colex measure can better and more
directly reflect how the exact number of ``incorrect''
patterns per concept changes with respect to the value of  $\lambda$.
Results are shown in Table \ref{tab:colexification_identification}.

\begin{table}
\centering
\footnotesize
\begin{tabular}{@{}rrccr@{}}
\toprule
$\lambda$ & $P$ & MicroRec. & MacroRec. & \#aw\_colex \\ \midrule
1                  & 1220             & 0.71          & 0.80          & 580.87          \\
5                  & 1056             & 0.63          & 0.77          & 84.34           \\
10                 & 1001             & 0.58          & 0.73          & 42.04           \\
20                 & 935              & 0.54          & 0.70          & 22.91           \\
50                 & 833              & 0.48          & 0.66          & 10.69           \\
100                & 761              & 0.42          & 0.62          & 5.78            \\ \bottomrule
\end{tabular}
\caption{\label{tab:colexification_identification}
The number of common concepts ($P$) in \networkone and CLICS,
Micro and Macro average recall, 
as well as the average number of colexification
patterns per concept that are not in CLICS (\#aw\_colex)
for different thresholds $\lambda$ (the minimum number of languages for keeping a colexification edge).}
\end{table}

If the constraint $\lambda$ is stricter, fewer concepts and
fewer edges (both colexification
edges contained and not contained in CLICS) will be included in \networkone. 
Thus, we
observe a consistent drop in both micro and macro recall.
On the other hand, we observe a decrease
in \#aw\_colex when we increase $\lambda$, as CLICS
edges are less likely to be removed than edges missing from
CLICS: many languages can share the same colexification
patterns for some concepts whereas edges not in CLICS will
not be shared across many languages. This can be
verified by the steepness of the decrease
in \#aw\_colex. From $\lambda = 1$ to 5, around 500 edges
not in CLICS are removed for each concept. When $\lambda >
5$, the speed decreases, suggesting the identified
colexification edges are more reliable. In summary, high
recall indicates that we successfully identify
many ground-truth colexifications directly from PBC.
% 500 edges per concept are normal because 1220 common concepts are just a subset of concepts in ColexNet
It is important to note that CLICS' coverage is far from
complete for low-resource languages:
for many of them, 
fewer than 100 concepts are included in
CLICS. Therefore, \#aw\_colex gives some indication of performance or discrepancy between CLICS and \networkone,
but many of the edges not in CLICS are actually correct. 
On the other hand, \networkone is not immune to semantic errors \citep{pairsman2008semantic}, such as antonyms, hypernyms, or hyponyms, due to co-occurrence or free translation.
See \secref{investigation} for a detailed analysis of the identified colexifications.

\begin{table*}
    \footnotesize
    \centering
    \begin{tabular}{lccccccc}
        \toprule
        & \multicolumn{3}{c}{roundtrip translation} & \multicolumn{3}{c}{verse retrieval} & verse classification \\
        \cmidrule(lr){2-4} \cmidrule(lr){5-7} \cmidrule(lr){8-8}
        & top-1 & top-5 & top-10 & top-1 & top-5 & top-10 & \\
        \midrule
        BOW & - & - & - & 0.02 & 0.05 & 0.06 & 0.09 \\
        \midrule
        S-ID \citep{levy-etal-2017-strong} & 0.10 & 0.21 & 0.25 & 0.17 & 0.29 & 0.35 & 0.32 \\
        CLIQUE \citep{dufter-etal-2018-embedding}& 0.22 & 0.63 & 0.79 & 0.41 & 0.62 & 0.70 & 0.44 \\
        $N(t)$ \citep{dufter-etal-2018-embedding} & 0.22 & 0.53 & 0.65 & 0.28 & 0.46 & 0.55 & 0.47 \\
        \midrule
        XLM-R \citep{conneau-etal-2020-unsupervised} & - & - & - & 0.04 & 0.07 & 0.09 & 0.15 \\
        Glot500-m \citep{imanigooghari-etal-2023-glot500} & - & - & - & 0.11 & 0.17 & 0.21 & 0.22 \\
        \midrule
        Eflomal-aligned & 0.24 & 0.58 & 0.70 & 0.61 & 0.76 & 0.81 & 0.48 \\
        \embname & \textbf{0.44} & \textbf{0.85} & \textbf{0.93} & \textbf{0.65} &  \textbf{0.80} &  \textbf{0.84} &  \textbf{0.49} \\
        \bottomrule
    \end{tabular}
    
    \caption{Results of different multilingual embeddings on
    roundtrip translation, verse retrieval, and verse
    classification tasks. Each number for roundtrip
    translation (top-$k$ accuracy, $k \in \{1, 5, 10\}$) is
    the average of 10 runs with 3 randomly selected
    intermediate languages. Each number in verse retrieval
    (top-$k$ accuracy, $k \in \{1, 5, 10\}$) and verse
    classification (macro $F_1$) is the average over all
    available languages (1,250 for verse retrieval, 1,245 for verse classification). We also report the performance of BOW, XLM-R, and Glot500-m. The first serves as a random baseline whereas the latter two are strong multilingual model baselines. 
    The performance reported for Glot500-m is different from the original paper as it was evaluated on a subset of languages the model supports, while we evaluate the model on all languages that \embname supports, making comparison easier.
    \textbf{Bold}: best result per column.}
    \label{tab:results}
\end{table*}

\subsection{Roundtrip translation}
We additionally use
roundtrip translation \citep{dufter-etal-2018-embedding} to
assess the quality of multilingual representations. Let
$[l_0,l_1, l_2, ..., l_R]$ be a sequence of languages where
$l_0=l_R$ is the source language and 
$l_i\neq l_0 \, \forall 1\leq i \leq R-1$ 
different intermediate languages.
Roundtrip
translation
starts with a word $w_0$ in $l_0$ and
iteratively finds the word $w_r$ in language $l_r$
($1 \leq r \leq R$)
that is
closest to word $w_{r-1}$ in language $l_{r-1}$ in the
embedding space.  If $w_0 =
w_R$, this indicates that the $R-1$ ``intermediate'' words
have representations similar to $w_0$ and
represent
the meaning of $w_0$.
We compute the percentage of roundtrips
for $w_0$ 
that are successful, i.e., $w_0 = w_R$ (top-1
accuracy). In addition, we also report top-5 and top-10
accuracies (i.e., $w_0$ is in the $k$ ($k=$ 5 or $k=$ 10)
nearest neighbors). We set $R=4$,
$l_0 = $ English and take
1,654 English words that occur in all embedding spaces as the
starting point $w_0$.
For each trial, we
randomly select 
three intermediate languages and then compute results for
each of the 1,654 $w_0$. 
We run this experiment ten times
and report averages.
We ensure
that the intermediate languages are different in each trial.

\subsection{Verse retrieval}

Similarly to Glot500-m, we use 500 English-aligned
verses from PBC for verse retrieval. 1,250 languages are used (we remove 85 languages that
cover fewer than 400 out of the 500 verses).
We represent each verse as
the average of the embeddings of its units.
Given a verse
$v_e$ in
English, we find the most cosine-similar verses
$v_l$ in target language $l$.
We then compute top-1, top-5 and top-10 accuracy for the
returned ranking (i.e., the correct verse is in the top-1,
top-5, top-10 nearest neighbors) and average first over verses and then languages.

\subsection{Verse classification}

We evaluate our multilingual embeddings on 
Taxi1500 \citep{ma2023taxi1500}. It provides 860/106/111 verses for
train/valid/test sets in more than 1,500 languages. Each
verse is annotated with one of six classes:
`recommendation', `faith', `description', `sin', `grace',
and `violence'. We use a subset
of 1,245 languages, those
covered by both Taxi1500 and \embname.
We perform
zero-shot transfer by training a logistic classifier
on English train and evaluating on the test
set of the other 1,244 languages. Similar to verse
retrieval, we
represent a verse as
the average of its embeddings. We report macro $F_1$,
first averaged over verses (per language) and then
averaged over languages.

\subsection{Results \& Discussion}\seclabel{results_discussion}
Table \ref{tab:results} shows that
\embname BOW, S-ID, CLIQUE, $N(t)$, Glot500-m and
Eflomal-aligned on all three tasks:
roundtrip translation, verse retrieval, and verse
classification.
\embname shows a large improvement over the baselines,
especially for
roundtrip translation and verse retrieval.
The bad performance of BOW is expected, as previously mentioned, 
because the English vocabulary does not necessarily transfer to other languages. 
\embname's improvement over S-ID
is probably due to the fact that there is only verse ID
information provided to serve as verse-level context
information in S-ID. Token-level
alignment information, however,
is not available to S-ID. In other words,
using abstract context identifiers alone cannot provide
enough information to learn good multilingual embeddings
for crosslingual transfer.
When comparing \embname with XLM-R and Glot500-m, we see a clear improvement in either verse retrieval or verse classification. The major reason is that both XLM-R and Glot500-m are not trained on all languages that are supported by \embname. Due to a lack of data in some low-resource languages, it is difficult to train a good language model in those languages. In contrast, \embname demonstrates the possibility of multilingual embeddings: with a small multilingual corpus where we can extract colexifications, it is already enough to support large-scale zero-shot transfer for the low-resource languages by training embeddings.

CLIQUE, $N(t)$, and Eflomal-aligned achieve similar
performance on roundtrip translation (top-1). However, when
$k$ becomes larger ($k=$ 5 or 10), we see that CLIQUE
performs better than $N(t)$ and Eflomal-aligned. This is not
surprising, since CLIQUE specifically creates cliques of
tokens that are translations of each other in
different languages. Therefore the representations of
translations should be similar. Eflomal-aligned also
achieves good performance on roundtrip translation when $k$
is large ($k=$ 5 or 10) and very close performance
to \embname in verse retrieval / classification. There are
a few possible explanations.
First, the word alignments are noisy in Eflomal-aligned
because it operates on the
token level and any information hidden inside each token (i.e., ngrams inside each token) cannot be extracted and utilized (see
the discussion also
in \citet{liu-etal-2023-crosslingual}).
Therefore, by increasing
$k$ in roundtrip translation, the influence of such alignment
noise is offset, resulting in better results. Second, as we
use the average of embeddings of tokens in a verse as the verse
representation in verse retrieval / classification; this 
can mitigate the impact of unimportant tokens. 

For verse classification, we find that different
embeddings achieve similar performance except for S-ID. On
the one hand, this phenomenon indicates that S-ID, though
it learns from abstract context information, cannot align
words from different languages that refer to the same
concepts well, thus preventing transfer from English to
low-resource languages. On the other hand, it might indicate
that classification is a less difficult task: it does not require the
model to have equally good alignment for all
concepts as the model can achieve good results
just by aligning important concepts.
Nevertheless, 
\embname still achieves better results than other baselines,
suggesting it has better zero-shot transferability. 
See \secref{english_transfer_learning} for complete results.

\begin{figure}
    \setlength{\belowcaptionskip}{-0.5cm}
  \centering
  \includegraphics[width=0.48\textwidth]{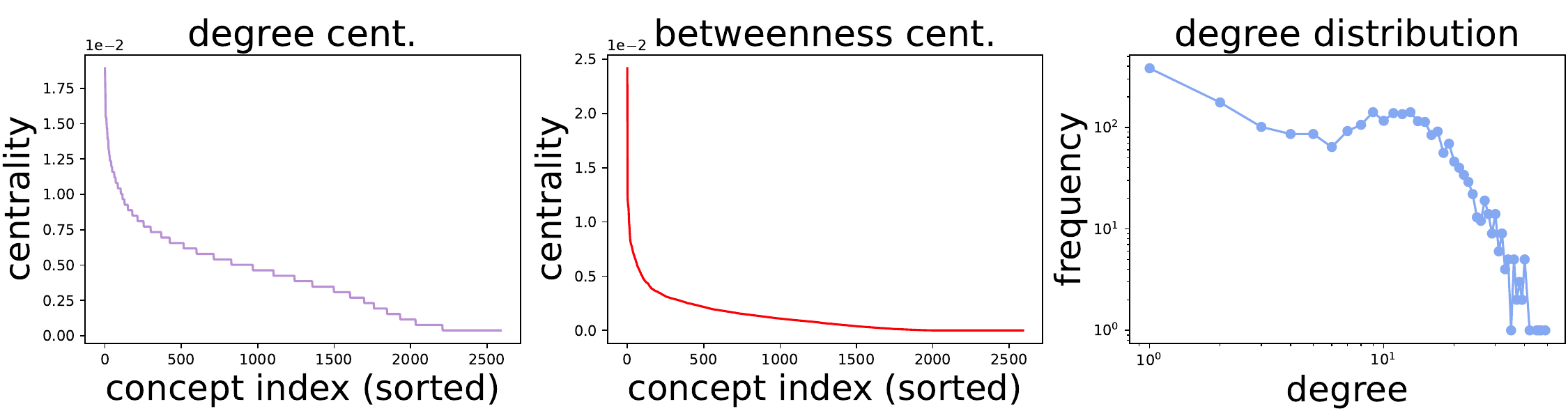}
  \caption{Degree centrality (left), betweenness centrality (middle) and degree distribution (right) of \networkone.}
  \label{fig:small_world}
\end{figure}

\section{Analysis}\seclabel{analysis}
\subsection{Analyses on \networkone}

% We find \networkone forms a graph with a very large connected component, with a small number of isolated connected components when $\lambda \geq 20$. 
% For (2), as we cannot verify it, we will be conservative
% This shows we can almost reach any concept starting from any concept. 

\paragraph{Basic statistics.} 
We find that \networkone has one
very large connected component along with a few small
connected components. See a visualization of the largest
community in \networkone ($\lambda=50$) in
Figure \ref{fig:largest_community} in the Appendix.
Therefore,
in the largest community, 
there is always a path
in the colexification graph
between  two concepts
even if they are less
related.
Figure \ref{fig:small_world} shows
degree/ betweenness
centrality and
degree distribution of \networkone.
From the figure, we can infer that the connectivity
can be attributed to (1) a small group of concepts that are
involved in many colexification patterns and (2) a small
group of edges serving as ``bridges'' to connect concepts
that are rarely colexified in some languages. 
Therefore, \networkone, a graph built by
the identified colexification patterns 
across many languages, approximately
forms a small-world or scale-free \citep{barabasi2003scale}
network. See \secref{construction} for graph-related statistics
of \networkone under different $\lambda$.

\begin{figure}
    \setlength{\abovecaptionskip}{0cm}
    \setlength{\belowcaptionskip}{-0.5cm}
  \centering
  \includegraphics[width=0.45\textwidth]{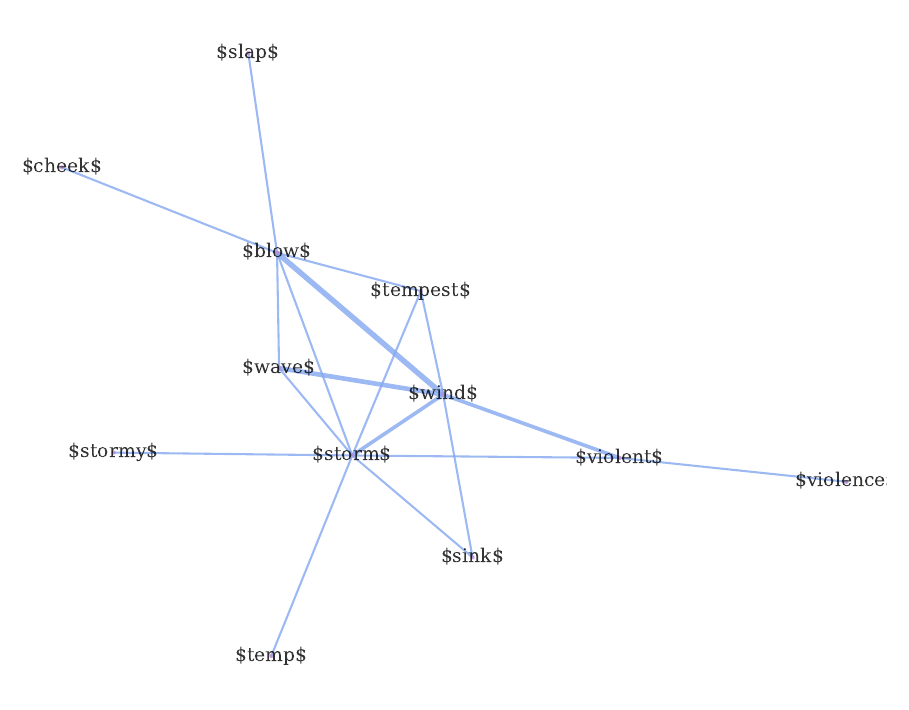}
  \caption{Community \#29.
Line thickness indicates the number of languages that
  instantiate a colexification.}
  \label{fig:community29}
\end{figure}

\paragraph{Communities.} We use the Louvain algorithm 
 \citep{blondel2008fast} to find
 communities in \networkone. We identify 288
 communities. Each community forms a cluster of
 semantically related concepts. 
Figure \ref{fig:community29}
gives the example of community\#29:
it contains several concepts related to <wind>, <storm> and
 <wave>. We see that <wind> is often colexified with
 <blow> (wind blows), with <wave>
 (waves are caused by wind)
 and with <violent> (winds can be fierce).
At the center of a community, we often
find a densely connected clique,
 indicating their connections are strong in many
 languages. Some concepts, 
located at the fringe of the community and connected with
 one of the densely-connected concepts in the center,
 are less related to the semantic field of the
 community and serve as ``bridges'' to connect with
 other communities. See \secref{communities} for further details of the identified communities.
 
\subsection{Transfer learning beyond English}\seclabel{main_transfer_beyond_eng}

NLP research in general and even typological studies are
frequently conducted from an English-centric perspective. To
reduce this bias and further verify 
our multilingual embeddings' transfer
capability, we additionally use \emph{all} available languages (1,245 languages) as the query / train languages for retrieval and
classification tasks. To this end, we conduct large-scale experiments that contain 1,245 $\times$ 1,245 transfer directions. 
The setup is the same as in \secref{experiments}, where each language takes the role of English as the query / train language.
We again represent each verse as the average
of the embeddings of its units.
For each language, we calculate the average top-$k$
($k=$ 1, 5, or 10) accuracy for verse retrieval and macro
$F_1$ for verse classification over all other languages except the language itself. 

In Table \ref{tab:results_beyongd_en}, we list the transfer
performance of three major languages that are typologically
different from English:
Arabic (arb), Russian (rus), and Chinese (zho); and three
languages that achieve \textbf{the worst} overall performance:
Apinayé (apn), Mündü (muh), and Salt-Yui
(sll). See \secref{beyond_transfer} for 
complete results for all languages.  
For high-resource
languages, the performance is close to that achieved for
English (see Table \ref{tab:results}), indicating that
the ngrams are well-aligned and \embname has good transfer
ability.  Chinese performs better than Arabic and
Russian. The possible reasons are as follows: (1) Both
Arabic and Russian are morphologically rich whereas Chinese
is not. Morphological variation makes finding aligned ngrams
in the forward pass harder, with a negative impact on
performance; (2) To prevent bad tokenization for Chinese,
we allow all ngrams (unlimited-length combination of
continuous characters) in a verse to be candidates in the
forward pass. This setting gives ngrams more freedom and
thus better results are expected.  For the three
low-resource languages, we find that they diverge morphologically and
typologically from most high-resource
languages. Apinayé and Mündü seem to frequently use
several consecutive whitespace-tokenized syllables to
express a single concept, which makes finding the correct
alignments much harder.
Salt-Yui, on the other hand, seems to be highly ambiguous because the writing does not reflect its contrastive tones \citep{irwin1974salt}. We hypothesize such ambiguity can negatively influence performance. See \secref{beyond_transfer} for an analysis of the factors that can influence transfer performance. 

\begin{table}
	\footnotesize
        \centering
 \begin{tabular}{lcccc}
		\toprule
		& \multicolumn{3}{c}{verse retrieval} & verse classification \\
		\cmidrule(lr){2-4} \cmidrule(lr){5-5}
		& top-1 & top-5 & top-10 & \\
		\midrule
		arb & 0.56 & 0.72 & 0.78 & 0.47 (0.07) \\
		rus & 0.55 & 0.72 & 0.78 & 0.48 (0.07) \\
		zho & 0.60 & 0.77 & 0.82 & 0.49 (0.05) \\
            \midrule
            apn & 0.21 & 0.38 & 0.46 & 0.38 (0.07) \\
            muh & 0.22 & 0.39 & 0.48 & 0.38 (0.07) \\
            sll & 0.24 & 0.42 & 0.51 & 0.39 (0.08) \\
            \midrule 
            avg. & 0.46 & 0.64 & 0.71 & 0.44 (0.08)\\
            % med. & 0.46 & 0.65 & 0.72 & 0.44 \\
		\bottomrule
	\end{tabular}
	
	\caption{Verse retrieval/classification for three
        high-resource languages, the three worst
        performing languages, and average over all languages (avg.). 
        We also report BOW results for verse classification (in parentheses), which serves as the random baseline.
        In contrast to the good
        performance for Arabic (arb), Russian (rus) and Chinese (zho),
        Apinayé (apn), Mündü (muh) and Salt-Yui
        (sll) each pose specific difficulties for inducing reliable
        colexification patterns.
 }\label{tab:results_beyongd_en}
\vspace{-0.3cm}
\end{table}

% by leveraging Conceptualaizer \citep{liu-etal-2023-crosslingual}. 

\section{Conclusion}
In this work, we present the multilingual
graphs \networkone and \networktwo, based on 
% the identification
colexifications extracted from a highly parallel corpus.  
Comparing with CLICS, we
show that we identify many gold-standard patterns
in \networkone. In addition, we analyze the structure
of \networkone and show it nearly forms a scale-free graph,
with many communities of semantically related concepts. Most
importantly, we contribute to crosslingual transfer learning
by inducing multilingual embeddings
\embname that are learned on
\networktwo. Our experiments indicate that
\embname largely represents concepts across languages in the
same semantic space. We show that 
\embname
outperforms several approaches, including multilingual embeddings and pretrained models, on three downstream
tasks. This indicates that embeddings learned from
colexification graphs improve  crosslingual transfer,
especially for low-resource languages for which it is often
infeasible to pretrain good models. Finally, our embeddings exhibit robust transfer
performance across many different source languages.
\newpage
\section*{Limitations}

Theoretically, one could identify, explore, and analyze colexification patterns from any parallel corpora and construct graphs of colexifications using the methods proposed in this paper. We use the PBC, a genre-specific parallel corpus in this work, which can limit some of the concepts to religions. Nevertheless, the goal of this work is to explore colexification patterns in as many languages as possible, including a lot of low-resource languages, without relying on any external resources. Therefore, the PBC corpus is a good fit for us. 

 We conduct extensive experiments to verify the crosslingual transfer capability of the multilingual embeddings learned on \networktwo. However, some experiments are in-domain (the evaluation tasks are still related to the Bible), e.g., verse retrieval and verse classification. The major reason is that we want to test the embedding's performance on all our supported languages. Unfortunately, as far as we know, evaluation datasets that cover such a wide range of languages, including low-resource languages, are missing in the community. Some datasets, for example, Tatoeba\footnote{\url{https://tatoeba.org}}, support hundreds of languages but contain many concepts, e.g., pizza, that do not occur in the Bible. Therefore, we do not evaluate our embeddings on those datasets.

\section*{Acknowledgement}
We would like to thank Verena Blaschke for constructive discussions and Yuxuan Zahed for the implementation of the online demo of \networkone and \networktwo. This work was funded by the European Research Council (grant \#740516).

\bibliography{anthology,custom}
\bibliographystyle{acl_natbib}

\appendix

\section{Choice of hyperparameters and discussion}\seclabel{hyperparam}
\subsection{Forward/backward pass}

Two hyperparameters in the forward pass and backward pass when searching for the colexification patterns can influence the results, i.e., (1) the maximum number of iterations $M$ for a given concept in each language and (2) the threshold $\alpha$ for the minimum cumulative coverage of the set of identified ngrams. We set $M=3$ and $\alpha = .9$ as default values for all involved computations. We are different with Conceptualizer \citep{liu-etal-2023-crosslingual} in the setting of $M$. Conceptualizer sets $M$ to 5 whereas we set it to 3. The major reasons are as follows. We are searching for colexification patterns with high accuracy. This requires us to identify the target-language ngrams that instantiate the colexifications with high certainty. Based on empirical explorations, we find that when $M$ is large (e.g., $>3$), we can include less accurate or even unrelated ngrams (because those ngrams are rare and occur in the same verse where the concept occurs, which is also discussed by \citet{liu-etal-2023-crosslingual}). By setting $M=3$ in the forward pass, we will be more confident that the identified target-language ngrams are highly correlated with the concept and this setting achieves the best performance for a few examples in our manual inspection. As for the minimum cumulative coverage threshold $\alpha$, we directly follow the setting in Conceptualizer, i.e., 0.9, to ensure that the forward pass and backward pass find enough ngrams/concepts while guaranteeing the quality of the associations.

\begin{table}
\centering
\footnotesize
\setlength\tabcolsep{3pt}
\begin{tabular}{lrrrr}
\toprule
$\lambda$ & {\#nodes} & {\#edges} & {degree} & {\#components} \\
\midrule
1 & 5870 & 1000937 & 170.61 & 1 \\
5 & 4028 & 122798 & 30.48 & 1 \\
10 & 3562 & 58031 & 16.29 & 1 \\
20 & 3133 & 30175 & 9.63 & 2 \\
50 & 2591 & 13607 & 5.25 & 9 \\
100 & 2221 & 7634 & 3.44 & 60 \\
\bottomrule
\end{tabular}
\caption{\label{tab:basic_stats}
Basic statistics of \networkone under different thresholds $\lambda$. We report the number of nodes (\#nodes), the number of edges (\#edges), the average degree per node (degree), and the number of connected components (\#components).
}
\end{table}

\begin{table}
\centering
\footnotesize
\setlength\tabcolsep{10pt}
\begin{tabular}{lrrr}
\toprule
$\lambda$ & {\#nodes} & {\#edges} & {degree} \\ 
\midrule
1 & 3,613,546 & 15,251,571 & 4.22\\
5 & 3,611,704 & 12,197,241 & 3.37\\
10 & 3,611,238 & 11,227,402 & 3.11\\
20 & 3,610,809 & 10,369,418 & 2.87\\
50 & 3,610,267 & 9,235,395 & 2.56\\
100 & 3,609,897 & 8,314,760 & 2.30\\
\bottomrule
\end{tabular}
\caption{\label{tab:basic_stats_expandednet}
Basic statistics of \networktwo under different thresholds $\lambda$. We report the number of nodes (\#nodes), the number of edges (\#edges), and the average degree per node (degree).
}
\end{table}

\begin{table*}
    \footnotesize
    \centering
    \begin{tabular}{lccccccc}
        \toprule
        $\lambda$ & \multicolumn{3}{c}{Roundtrip translation} & \multicolumn{3}{c}{verse retrieval} & verse classification \\
        \cmidrule(lr){2-4} \cmidrule(lr){5-7} \cmidrule(lr){8-8}
        & top-1 & top-5 & top-10 & top-1 & top-5 & top-10 & \\
        \midrule
        1 & 0.29 & 0.66 & 0.78 & 0.51 & 0.68 & 0.74 & \textbf{0.51} \\
        5 & 0.34 & 0.71 & 0.82 & 0.58 & 0.74 & 0.79 & \underline{0.49} \\
        10 & 0.36 & 0.74 & 0.84 & 0.59 & 0.75 & 0.81 & 0.48 \\
        20 & 0.38 & 0.77 & 0.87 & 0.63 & 0.78 & 0.83 & 0.48 \\
        50 & \underline{0.40} & \underline{0.81} & \underline{0.91} & \underline{0.65} & \underline{0.80} & \underline{0.84} & \underline{0.49} \\
        100 & \textbf{0.42} & \textbf{0.84} &  \textbf{0.93} & \textbf{0.66} & \textbf{0.81} & \textbf{0.85} & 0.45 \\
        \bottomrule
    \end{tabular}
    
    \caption{Results of multilingual embeddings trained on \networktwo under different $\lambda$. Each number for roundtrip translation (top-$k$ accuracy, $k = [1, 5, 10]$) is the average of 10 runs with 3 randomly selected intermediate languages. Each number in verse retrieval (top-$k$ accuracy, $k = [1, 5, 10]$) and verse classification (macro $F_1$) is the average over \textbf{all} available languages. \textbf{Bold} (\underline{underlined}): best (second-best) result per column.}
    \label{tab:results_different_langs}
\end{table*}

\subsection{\networkone/\networktwo construction}\seclabel{construction}
In the construction of \networkone and \networktwo, we have an important hyperparameter $\lambda$: the minimum number of languages for a colexification edge to be included.
As shown in Table \ref{tab:basic_stats}, different $\lambda$ can influence the number of nodes and edges in \networkone as well as the number of connected components. It is clear that
both \#edges and degree decrease dramatically from
$\lambda=$ 1 to 5, which might indicate: (1)
increasing $\lambda$ decreases the number of incorrectly 
identified colexification patterns 
(e.g., due to verse-level misalignment); (2) some colexification
patterns might be specific to very few languages. Because of many plausible incorrect edges between concepts, when $\lambda=$ 1, 5 or 10, \networkone forms a large connected graph. When $\lambda$ is larger (e.g., 50 or 100), the graph is no longer connected because many less reliable edges are removed from it.

The influence of $\lambda$ also apply to \networktwo,
since edges being removed in \networkone also impact \networktwo: pairs of edges that are expanded from removed edges from \networkone are then not included in \networktwo. We show the number of nodes and edges as well as the average degree in \networktwo under different $\lambda$ in Table \ref{tab:basic_stats_expandednet}. The changes in degree with the increase of $\lambda$ are not as prominent as in \networkone (shown in Table \ref{tab:basic_stats}). This is mainly because the number of nodes in \networktwo is far more than that in \networkone. Most of the nodes are only associated with around 3 other nodes in \networktwo, which indicates that many ngrams from target languages colexify about three concepts, because most of the nodes in \networktwo belongs to the ngram nodes. Each concept, however, can be frequently associated with more than 3 concepts in \networkone, as we noticed the average degree of \networkone ($\lambda=$50) is around 5.

The number of nodes and edges also influences the random walks which we used for sampling, thus influencing the quality of multilingual embeddings trained on \networktwo using Node2Vec \citep{grover2016node2vec}. Therefore, we conduct experiments using embeddings trained on \networktwo under different $\lambda$. Same as \secref{experiments}, we conduct experiments on roundtrip translation, verse retrieval, and verse classification tasks. For roundtrip translation, we again set $l_0=$ English and use 2,221 words that occur in all embeddings as the start points. For verse retrieval (resp. classification), we also use English as the query (resp. train) language, and report top-$k$ accuracy (resp. macro $F_1$ score), averaged over all languages. Results are shown in Table \ref{tab:results_different_langs}.

\begin{figure}
  \centering
  \includegraphics[width=0.48\textwidth]{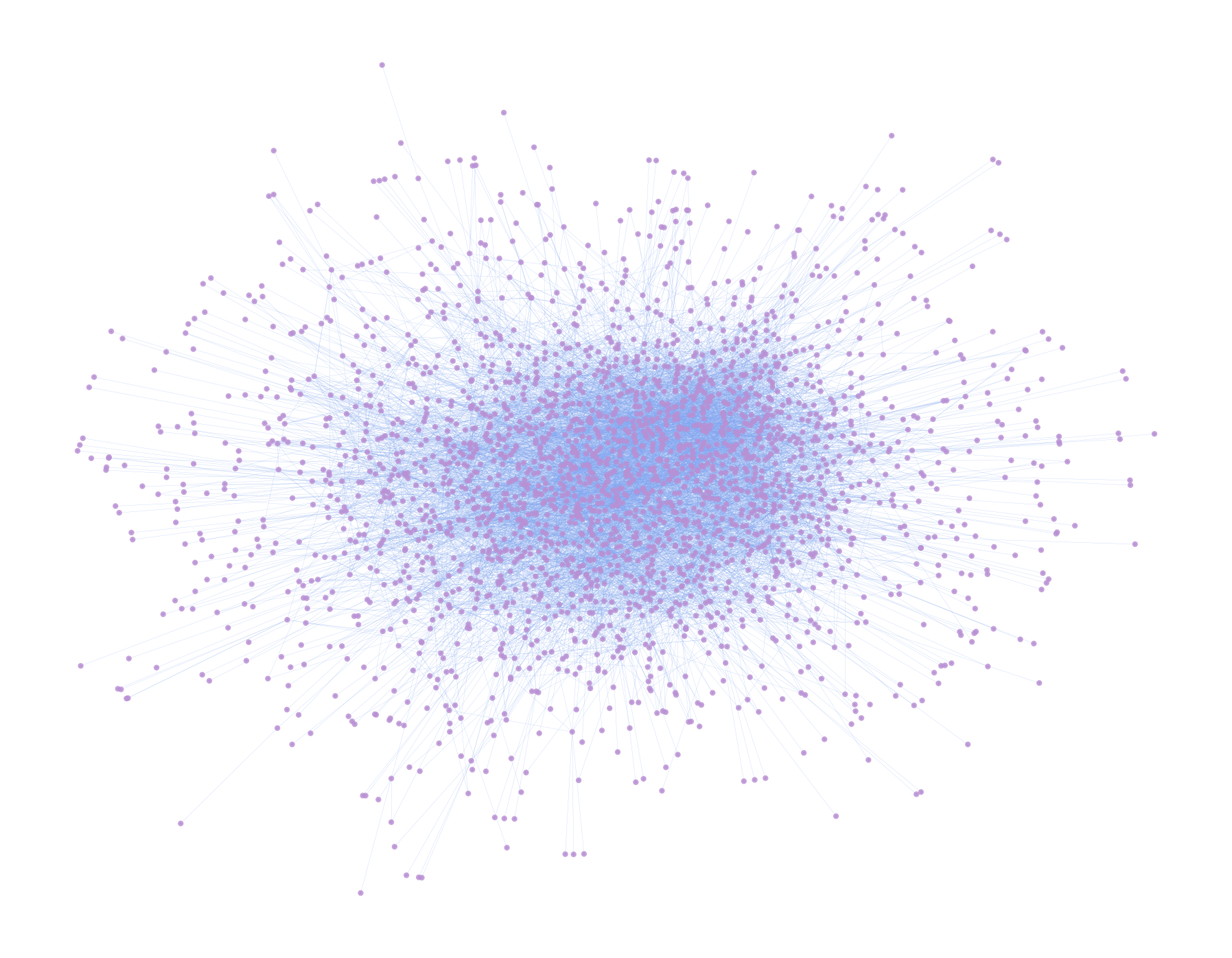}
  \caption{Visualization of the largest community which
  contains 2,581 nodes out of 2,591 nodes
  in \networkone. Each node is a concept and each edge
  indicates that two concepts are colexified in
at least 50 languages.}
  \label{fig:largest_community}
\end{figure}

We see there are different trends between the changes in $\lambda$ and changes in performance for different tasks: (1) the performance of roundtrip translation is positively correlated with $\lambda$ and the best result is achieved when $\lambda=$100; (2) the performance of verse retrieval is also positively correlated with $\lambda$ ; (3) the performance of verse classification is generally negatively correlated with $\lambda$ where the best result is achieved when $\lambda=$1. Those trends can be explained as follows. Roundtrip translation and verse retrieval, compared with verse classification, require better alignment quality among concepts and ngrams. When $\lambda$ is small, some incorrect edges are included in the graph. These edges induce noises for sampling, therefore slightly noisy embeddings are obtained, negatively influencing the performance. As for verse classification, the results suggest that when we have fewer out-of-vocabulary ngrams in the embeddings (higher $\lambda$ induces fewer ngrams in \networktwo), slightly better performance is achieved. Moreover, $\lambda$ seems to have a more obvious impact on roundtrip translation and verse retrieval than on verse classification. In summary, the results verify our choice of $\lambda=$50, a relatively large number, in the main content of this paper, as it offers very competitive results compared to other choices while not losing many interesting patterns.

\section{Investigation of identified colexifications}\seclabel{investigation}

In \secref{colex_identification}, we show that we identify many ground-truth colexification patterns compared with CLICS. However, there are quite a few colexification patterns that are not present in CLICS. Therefore, we conduct a qualitative investigation on those colexification patterns. We classify each pattern (an colexification edge in \networkone between two concepts) as one of the following categories: (1) full colexification (2) partial colexification and (3) incorrect colexification.

\paragraph{Full colexification.} Full colexification indicates that a word in a language directly colexifies two concepts. We list 4 examples of colexifications not included in CLICS but verified by us. 
An obvious example is that <ground> and <land> are colexified in many languages, e.g., through \begin{CJK*}{UTF8}{bsmi}土地\end{CJK*} in Japanese (jpn) and \begin{CJK*}{UTF8}{bsmi}大地\end{CJK*} in Chinese (zho). 
<early> and <tomorrow> are frequently colexified in Turkic languages, e.g.,
\begin{itemize}
    \item Southern Altai (alt): \texttt{\$\foreignlanguage{russian}{эртен}}
    \item Bashkir (bak): \texttt{\$\foreignlanguage{russian}{иртән}\$}
    \item Kyrgyz (kir): \texttt{\$\foreignlanguage{russian}{эртен}}
    \item Nogai (nog): \texttt{\$\foreignlanguage{russian}{эртен}}
\end{itemize}
Another interesting example is that <love> and <wish> are frequently colexified, e.g., in Min Nan Chinese (nan) through the word \texttt{\$ài} (character: \begin{CJK*}{UTF8}{bsmi}愛\end{CJK*}), as the character means both <love> and <wish>. Lastly, in Western Frisian (fry), <dragon> and <snake> are colexified through the word \texttt{\$draek\$}, for which we manually verify in PBC. It is worth noting that there is another word \texttt{slang} which denotes <snake> in Western Frisian.

\begin{table*}
\centering
\small
\setlength\tabcolsep{3pt}
\begin{tabular}{lllll}
        \toprule
        types & incorrect colex. & languages & ngrams & context \\
        \midrule
        & <four> <twenty> & cat & \texttt{\$quatre\$}, \texttt{\$vint} & \#66004004:	... \textbf{vint}-i-\textbf{quatre} setials més ... \\
        \multirow{1}{*}{\textbf{co-occurrence}} & <left> <right> & nog & \texttt{\$\foreignlanguage{russian}{онъ}\$}, \texttt{\$\foreignlanguage{russian}{сол}\$} & \#01048013: \foreignlanguage{russian}{... \textbf{онъ} колы ... \textbf{сол} ...}\\ 
        & <want> <know> & nds & \texttt{\$weete\$}, \texttt{\$well\$} & \#46011003:	Oba etj \textbf{well} , daut jie \textbf{weete} ...\\
        \midrule
        \multirow{2}{*}{\textbf{free translation}} & <man> <answer>  & cat & \texttt{\$contest} & \#40012048: Però ell va \textbf{contest}ar ... \\
        & <hundred> <thousand> & cmn & \texttt{\begin{CJK*}{UTF8}{bsmi}十万\end{CJK*}} & \#13005021: ... \begin{CJK*}{UTF8}{bsmi}以及人口\textbf{十万}\end{CJK*}...\\
        \bottomrule
    \end{tabular}
\caption{\label{tab:incorrect_colexification}
Examples of incorrectly identified colexifications in \networkone.
}
\end{table*}

\paragraph{Partial colexification. } Partial colexification denotes the pattern that does not involve an entire word, but rather part of it. The part can be a shared root or a shared element in a compound. Since our algorithm works on the character ngrams within a word, we find many partial colexification patterns. For example, <stand> and <build> are colexified in Kazakh (kaz) through \texttt{\$\foreignlanguage{russian}{тұр}}. In Kazakh, <stand> is expressed by the word \foreignlanguage{russian}{тұру} while <build> by \foreignlanguage{russian}{тұрғызу}. \foreignlanguage{russian}{тұрғызу} is actually derived from the root \foreignlanguage{russian}{тұр-} plus a causative suffix \foreignlanguage{russian}{-ғыз}, so that \foreignlanguage{russian}{тұрғыз-} means ‘to make something stand’, thus meaning `build, erect'. Such partial colexifications through a root can even include more than two concepts, e.g., <morning>, <early>, and <next> in many Turkic languages:
\begin{itemize}
    \item Turkmen (tuk): \\ 
    \texttt{\$ertesi} $\rightarrow$ <next>\\
    \texttt{\$erte} $\rightarrow$ <morning>
    \item Turkish (tur): \\
    \texttt{\$ertesi} $\rightarrow$ <next>\\
    \texttt{\$erte} $\rightarrow$ <morning> <early>
    \item Uyghur (uig): \\
    \texttt{\$\foreignlanguage{russian}{әтиси}\$} $\rightarrow$ <next>\\
    \texttt{\$\foreignlanguage{russian}{әтигән}} $\rightarrow$ <morning> \\
    \texttt{\$\foreignlanguage{russian}{әти}} $\rightarrow$ <early>
\end{itemize}
Some concepts may be expressed using multiple lexemes, forming a compound, and a part of the compound may also occur in the expression of a different concept.
Note that, in some languages (such as German), a compound is written together without any space in between, whereas in some other languages (such as English), there is a space between each part. In either case, these are considered as compounds, since all the separate elements together constitute one concept. This cannot be confused with co-occurrence, where the two concepts themselves co-occur. For example, in Tatar (tat), two colors: <purple> and <scarlet> are partially colexified through \foreignlanguage{russian}{\$кызыл\$}, because <purple> is \foreignlanguage{russian}{куе кызыл} (literally `thick red'), which contains a part \foreignlanguage{russian}{кызыл} meaning `red, scarlet'. Such partial relation also frequently exists in numbers. For example, \texttt{empat belas} (resp. \begin{CJK*}{UTF8}{bsmi}十四\end{CJK*}), which means 14, and \texttt{empat puluh} (resp. \begin{CJK*}{UTF8}{bsmi}四十\end{CJK*}), which means 40, are partially colexified in Indonesian (ind) (resp. Chinese (zho)), as \texttt{empat} (resp. \begin{CJK*}{UTF8}{bsmi}四\end{CJK*}) means 4. Some languages, e.g., Chinese and German, construct compounds without inserting blanks between each lexeme, so we also observe many partial colexifications in Chinese and Germanic languages, e.g., :
\begin{itemize}
    \item Chinese (cmn): \\ 
    \begin{CJK*}{UTF8}{bsmi}震\end{CJK*} $\rightarrow$ <tremble>\\
    \begin{CJK*}{UTF8}{bsmi}地震\end{CJK*} $\rightarrow$ <earthquake>
    \item Bavarian (bar): \\
    \texttt{bibn} $\rightarrow$ <tremble>\\
    \texttt{\$erdbibn\$} $\rightarrow$ <earthquake>
    \item German (deu): \\
    \texttt{\$beben} $\rightarrow$ <tremble>\\
    \texttt{\$erdbeben\$} $\rightarrow$ <earthquake>
\end{itemize}
In summary, many identified colexification patterns in \networkone belong to this category, which is the reason why we found many patterns that do not exist in CLICS, since CLICS only includes full colexification patterns.

\paragraph{Incorrect colexification.} As an automatic statistical method, the results are not immune to errors. Typically, we find the incorrectly identified colexifications are mainly due to two reasons: (1) \textbf{co-occurrence} and (2) \textbf{free translation}. We list some incorrectly identified colexifications in Table \ref{tab:incorrect_colexification}. Co-occurrence denotes that two particular concepts tend to co-occur very often so that the algorithm wrongly establishes connections between the concept. For example, we found <four> and <twenty> are associated in Catalan (cat) because the ngrams \texttt{\$quatre\$}, \texttt{\$vint} which refer to the two concepts respectively co-occur very frequently in PBC. Similarly, <left> and <right> for Nogai (nog), and <want> and <know> for Low German (nds) also belong to this type of error. Free translation means that the translation is not done word by word so that the corresponding word for a specific concept does not occur in the same sentence. In this case, the algorithm has no chance of finding the corresponding ngram, which ideally would align with the intended concept. Free translation is very common in the Bible because of its religious textbook nature. For example, in Catalan (cat), the English verse \#40012048 starts with ``But to the \textbf{man} who told him'' but the Catalan translation starts with ``Però ell va \textbf{contest}ar al qui deia això'', which means ``\textit{But he \textbf{answer}ed the one who said this}'', where the concept <man> does not occur in Catalan and the concept <answer> does not occur in English verse. Similarly, Chinese word \begin{CJK*}{UTF8}{bsmi}十万\end{CJK*} means one hundred thousand, i.e., 100,000 (with \begin{CJK*}{UTF8}{bsmi}十\end{CJK*} being 10 and \begin{CJK*}{UTF8}{bsmi}万\end{CJK*} being 10,000). As the formation of the number expression in Chinese is different from its English counterpart, the algorithm wrongly associates <hundred> and <thousand>.

\section{Communities of \networkone}
\begin{figure*}[htbp]
\centering

\subfigure[community \#6]{
\begin{minipage}[t]{0.33\linewidth}
\centering
\includegraphics[width=1.8in]{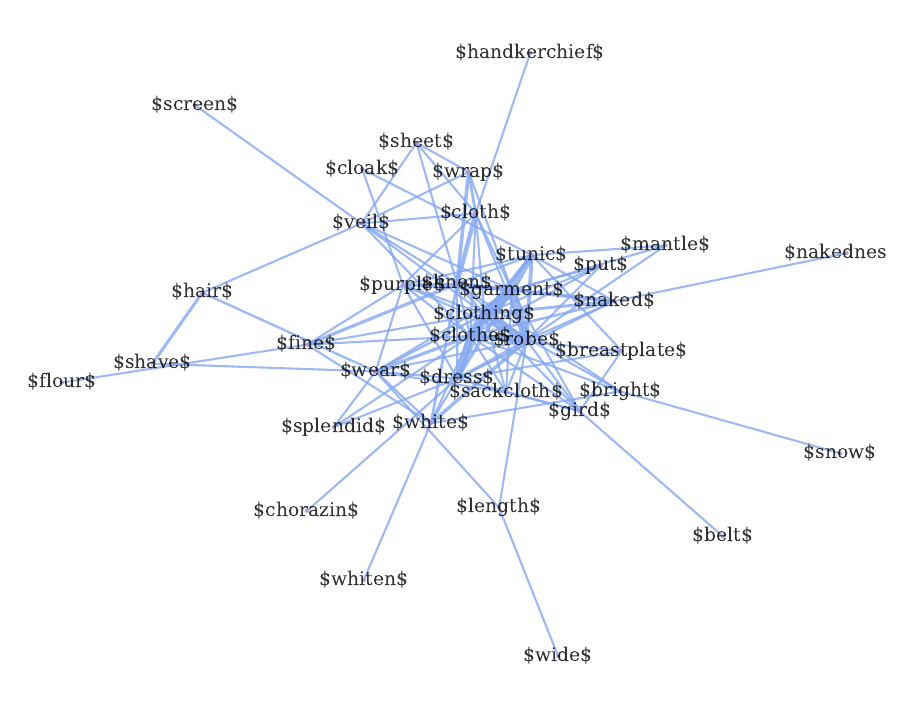}
\end{minipage}%
}%
\subfigure[community \#26]{
\begin{minipage}[t]{0.33\linewidth}
\centering
\includegraphics[width=1.8in]{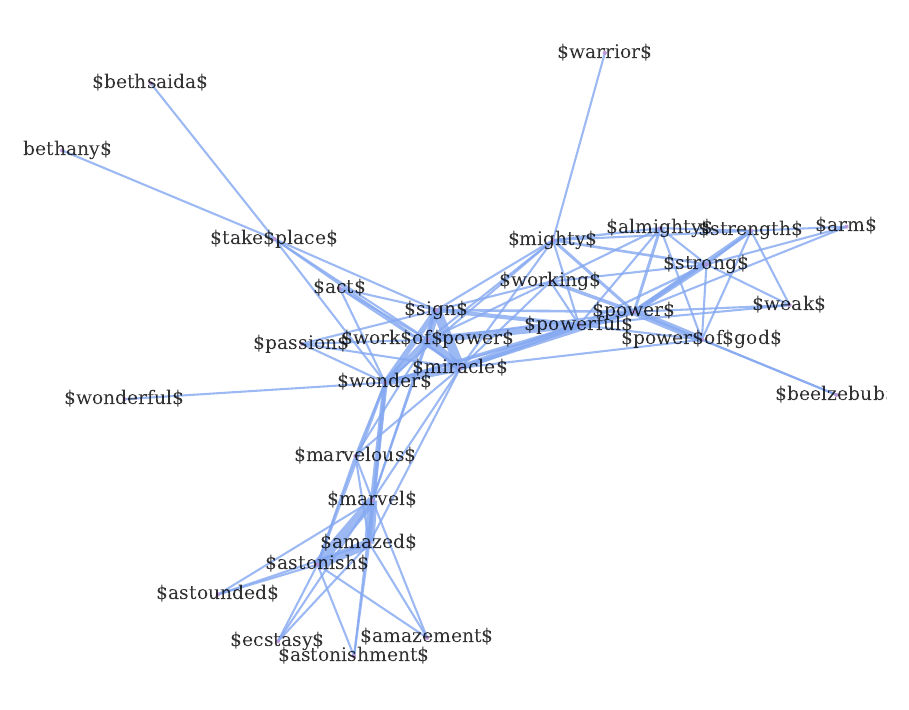}
\end{minipage}%
}%
\subfigure[community \#29]{
\begin{minipage}[t]{0.33\linewidth}
\centering
\includegraphics[width=1.8in]{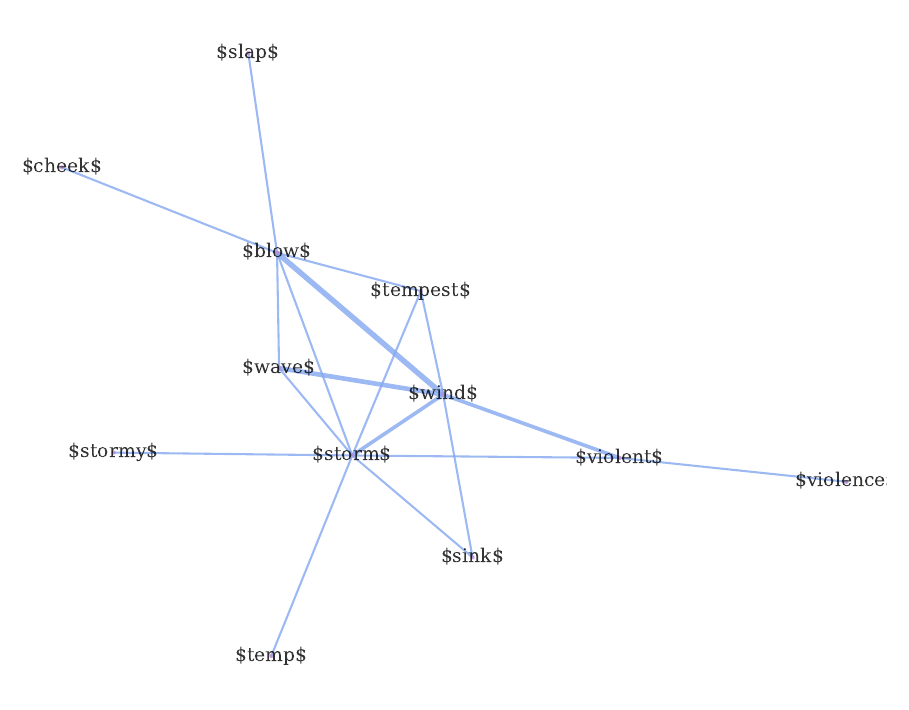}
\end{minipage}%
}%

\subfigure[community \#38]{
\begin{minipage}[t]{0.33\linewidth}
\centering
\includegraphics[width=1.8in]{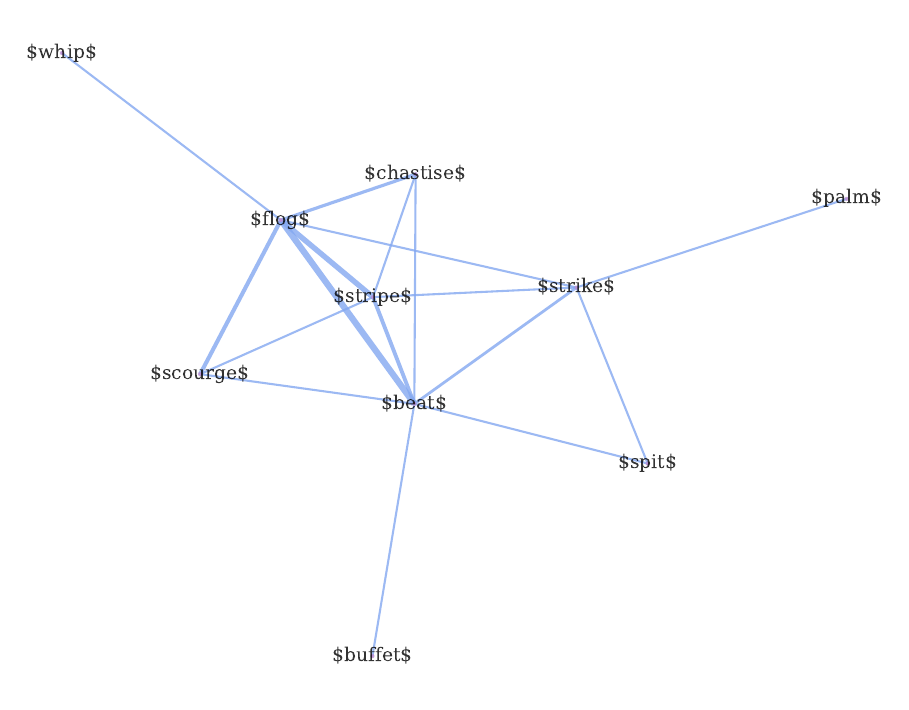}
\end{minipage}%
}%
\subfigure[community \#40]{
\begin{minipage}[t]{0.33\linewidth}
\centering
\includegraphics[width=1.8in]{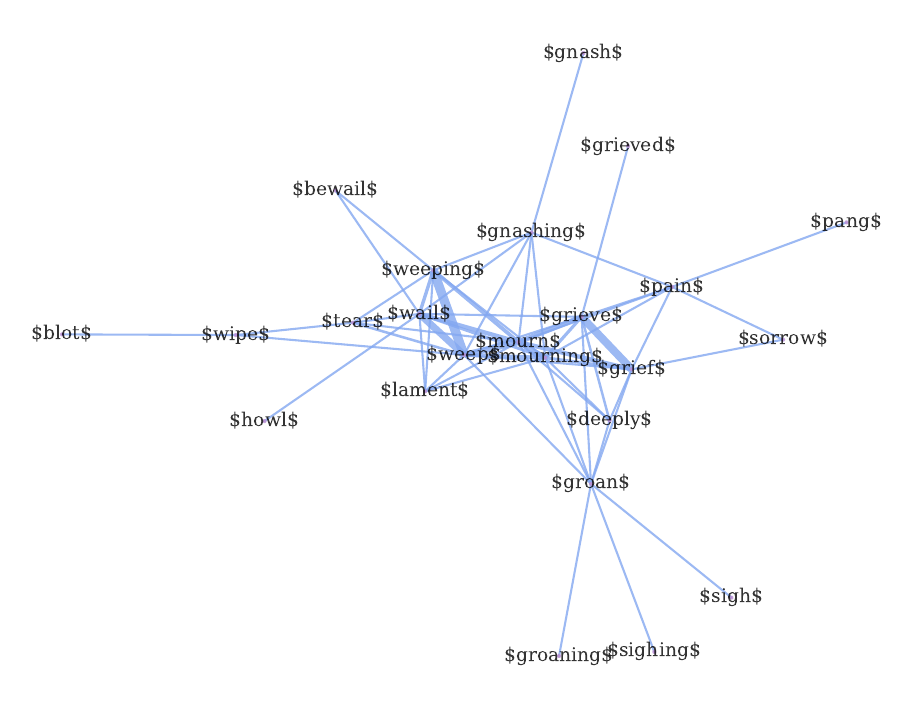}
\end{minipage}%
}%
\subfigure[community \#46]{
\begin{minipage}[t]{0.33\linewidth}
\centering
\includegraphics[width=1.8in]{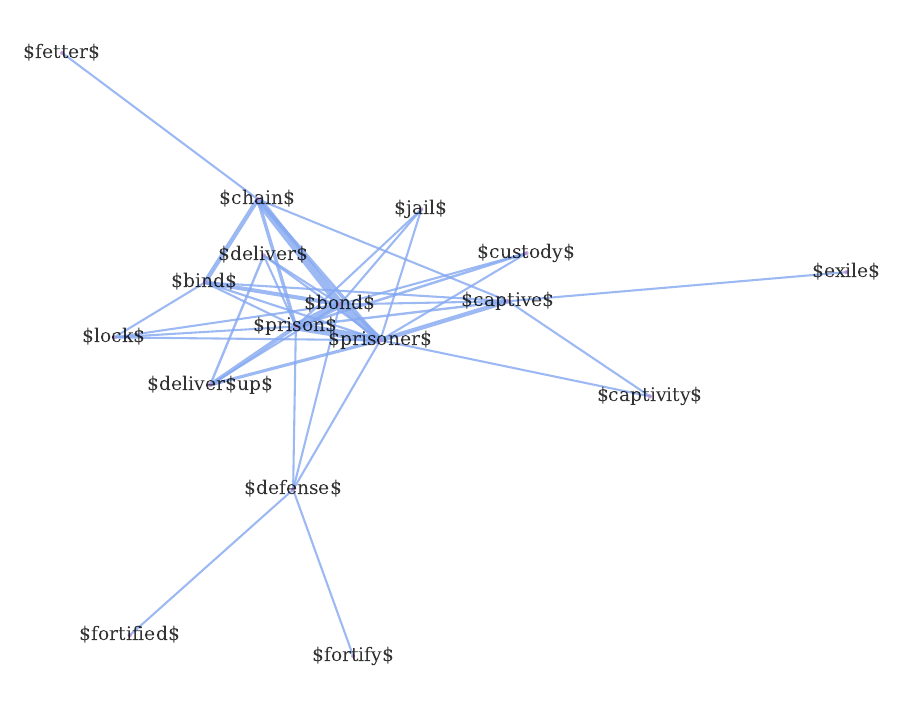}
\end{minipage}%
}%

\subfigure[community \#60]{
\begin{minipage}[t]{0.33\linewidth}
\centering
\includegraphics[width=1.8in]{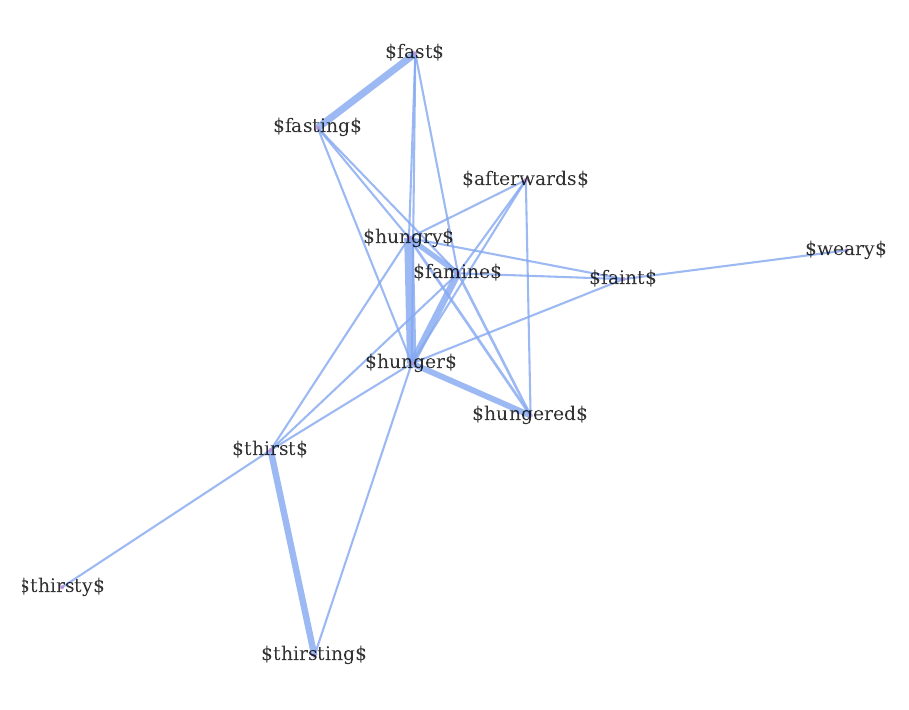}
\end{minipage}%
}%
\subfigure[community \#73]{
\begin{minipage}[t]{0.33\linewidth}
\centering
\includegraphics[width=1.8in]{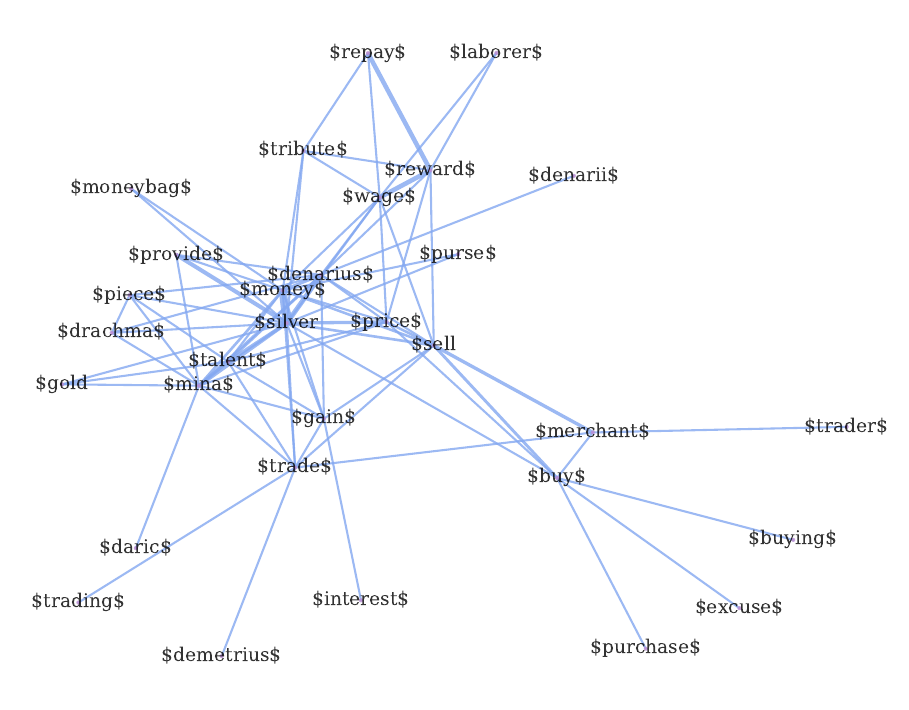}
\end{minipage}%
}%
\subfigure[community \#79]{
\begin{minipage}[t]{0.33\linewidth}
\centering
\includegraphics[width=1.8in]{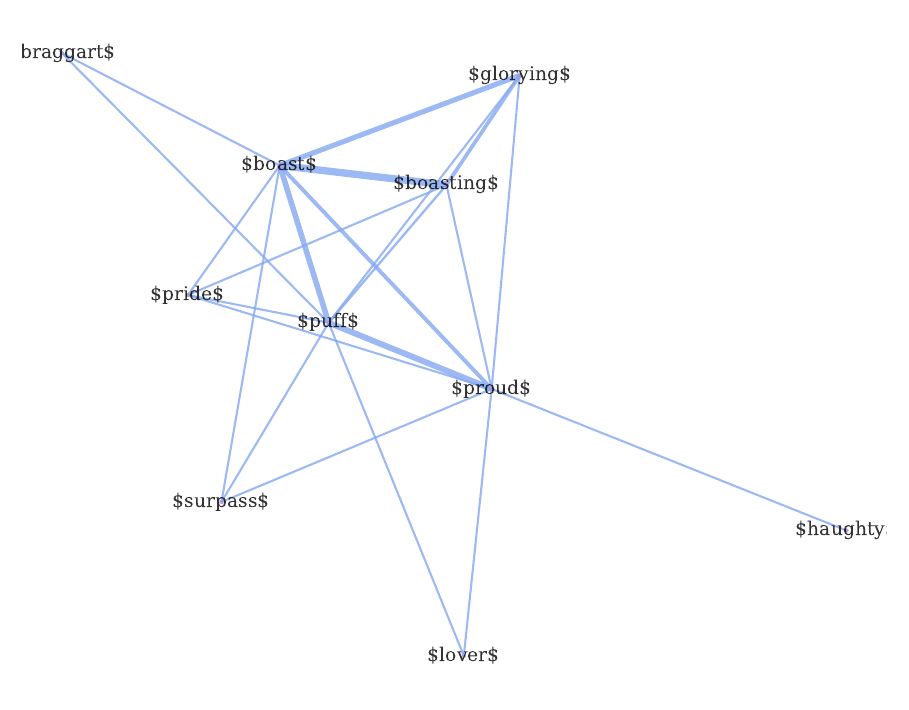}
\end{minipage}%
}%

\subfigure[community \#125]{
\begin{minipage}[t]{0.33\linewidth}
\centering
\includegraphics[width=1.8in]{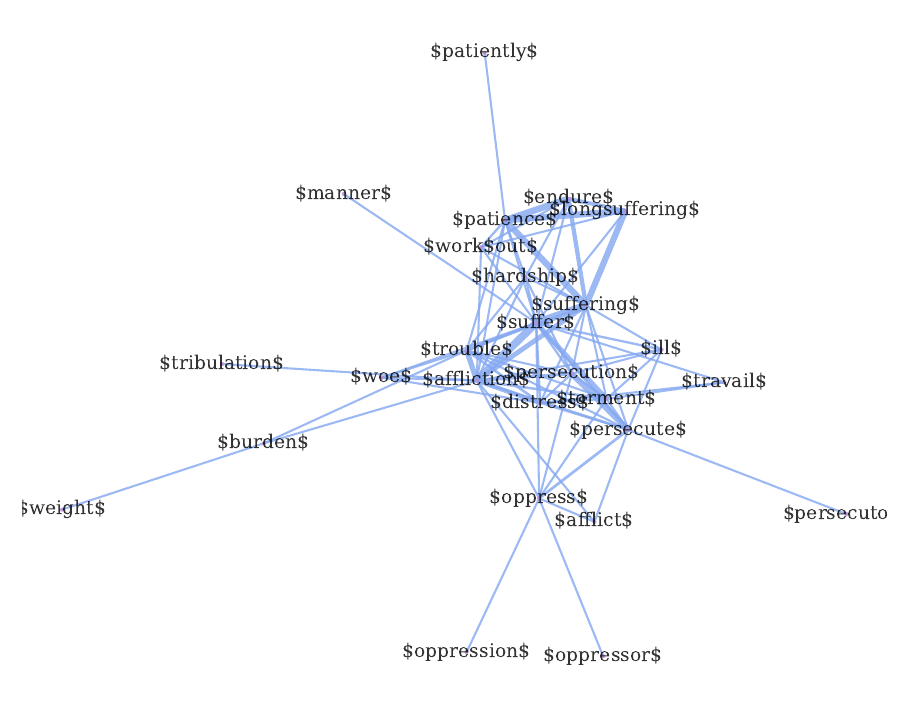}
\end{minipage}%
}%
\subfigure[community \#143]{
\begin{minipage}[t]{0.33\linewidth}
\centering
\includegraphics[width=1.8in]{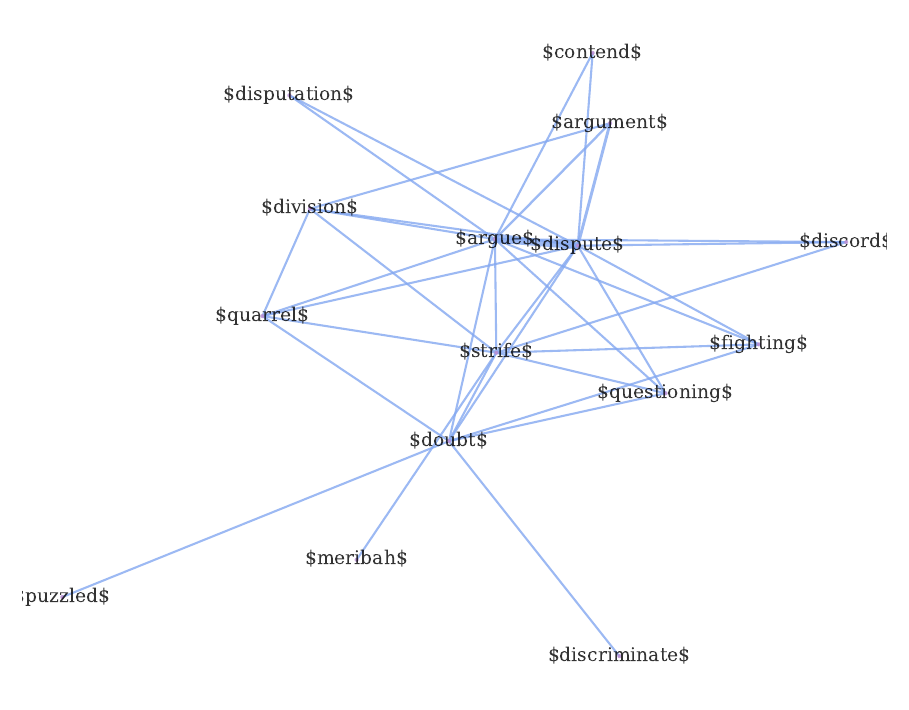}
\end{minipage}%
}%
\subfigure[community \#194]{
\begin{minipage}[t]{0.33\linewidth}
\centering
\includegraphics[width=1.8in]{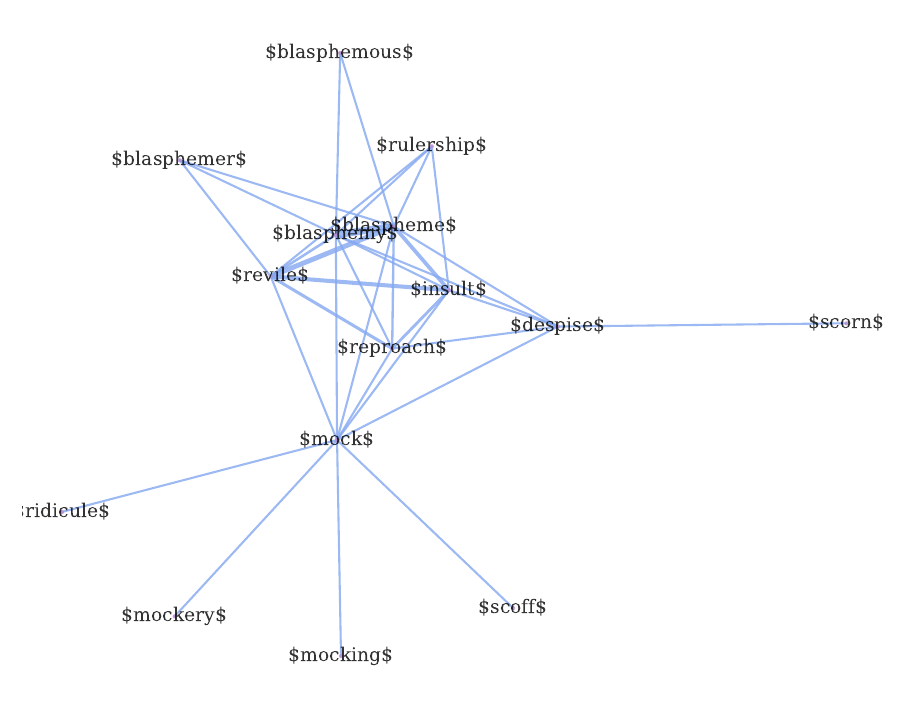}
\end{minipage}%
}%

\subfigure[community \#214]{
\begin{minipage}[t]{0.33\linewidth}
\centering
\includegraphics[width=1.8in]{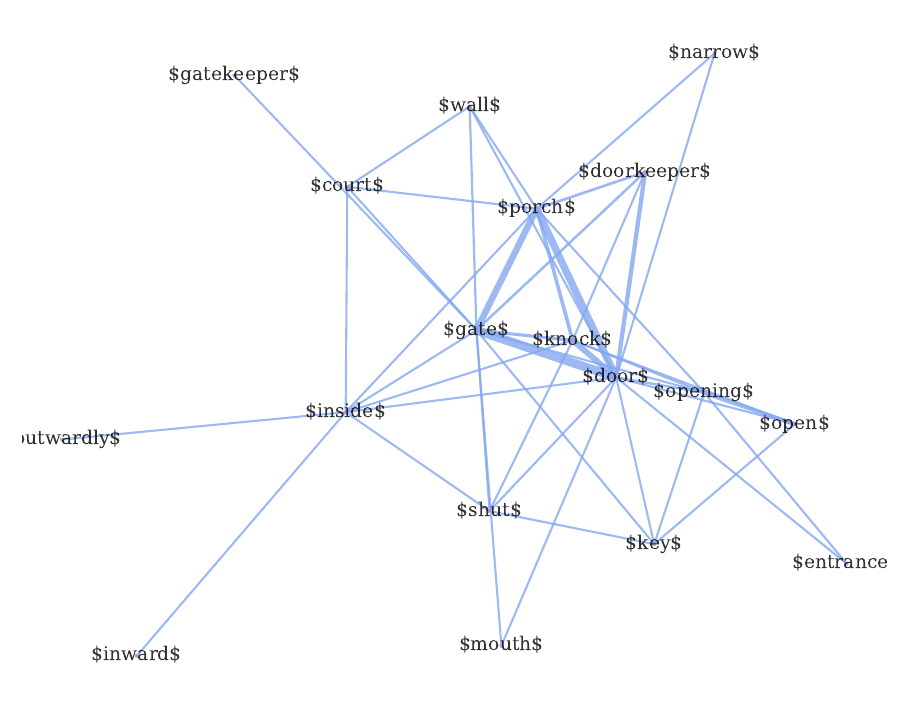}
\end{minipage}%
}%
\subfigure[community \#221]{
\begin{minipage}[t]{0.33\linewidth}
\centering
\includegraphics[width=1.8in]{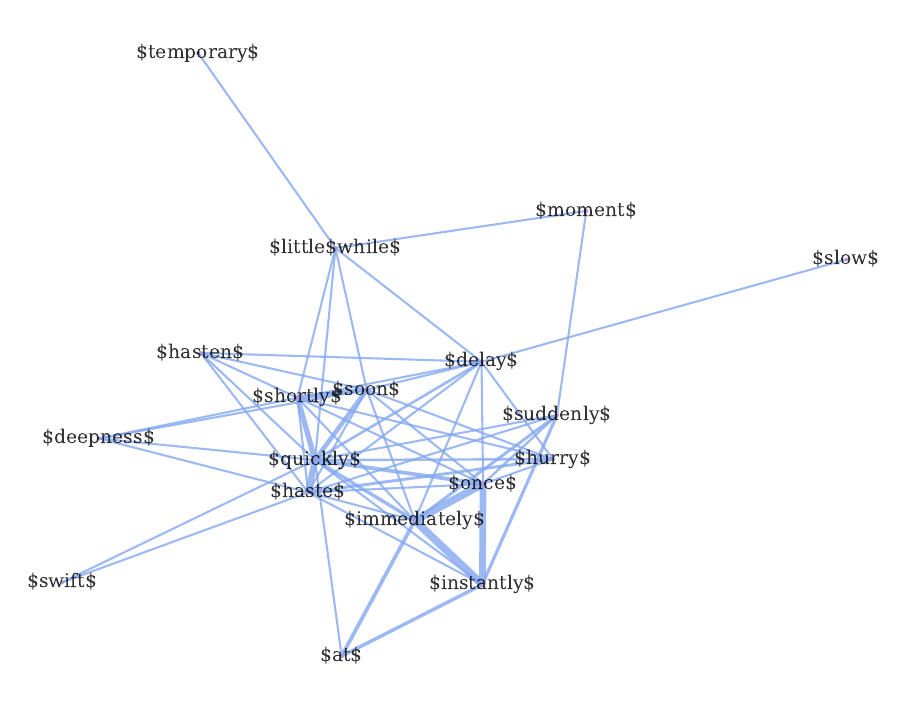}
\end{minipage}%
}%
\subfigure[community \#263]{
\begin{minipage}[t]{0.33\linewidth}
\centering
\includegraphics[width=1.8in]{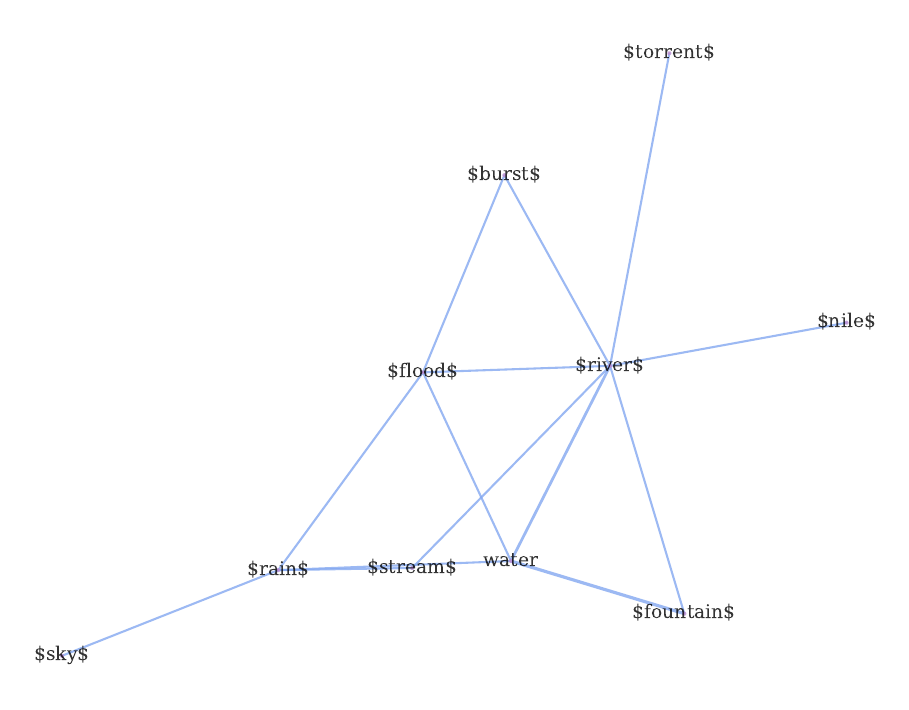}
\end{minipage}%
}%
\centering
\caption{Visualizations of 15 randomly selected communities that have more than 10 nodes from 288 communities detected in \networkone. Each community forms a cluster of concepts that are semantically related to each other. E.g., community \#60 is related to the concept <hunger>; and community \#73 is related to the concept <money>.}
\label{fig:communities}
\end{figure*}

\seclabel{communities}
There are 288 communities in total detected in \networkone ($\lambda=$ 50) by Louvain community detection algorithm \citep{blondel2008fast}. Two important hyperparameters, i.e., resolution and random seed are set to 0.1 and 114514 respectively. As mentioned in \secref{analysis}, each community is a cluster of concepts that are semantically related to each other. We create a demonstration website to show the subnetworks of each concept and the community figures.\footnote{\url{https://conceptexplorer.cis.lmu.de/}} For illustration purposes, we randomly select 15 communities that have more than 10 nodes for illustration in this paper. See Visualizations of those communities in Figure \ref{fig:communities}.

\section{Influence of language families \& areas}
\begin{figure*}
% \fbox{
\begin{minipage}{0.74\textwidth}
    % \hspace{-0.1cm}
    \includegraphics[width=0.32\textwidth]{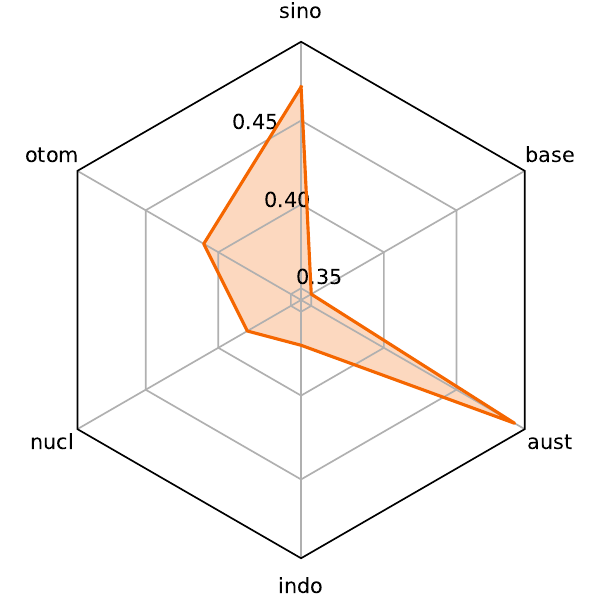}
    % \hspace{0.2cm}
    \includegraphics[width=0.32\textwidth]{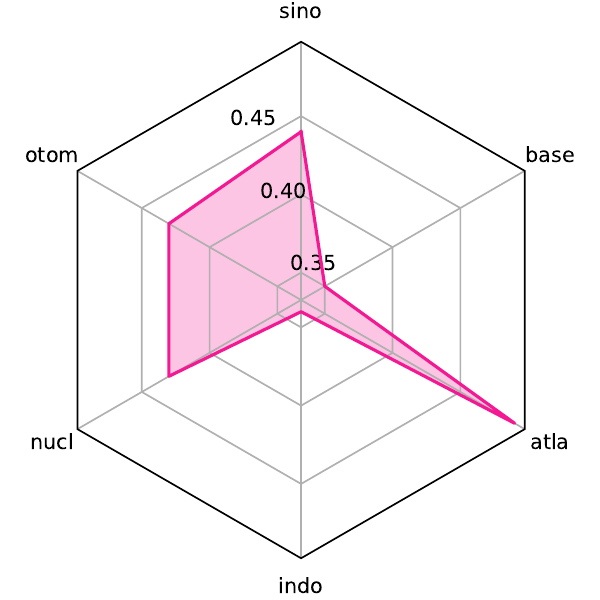}
    \includegraphics[width=0.32\textwidth]{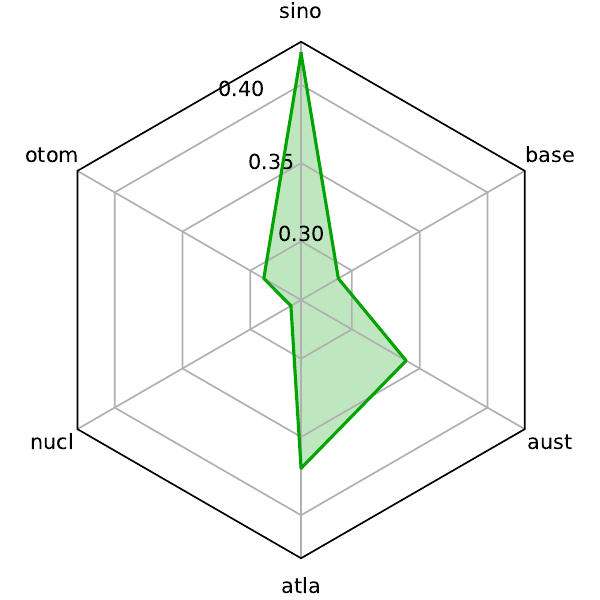}
    
    \includegraphics[width=0.32\textwidth]{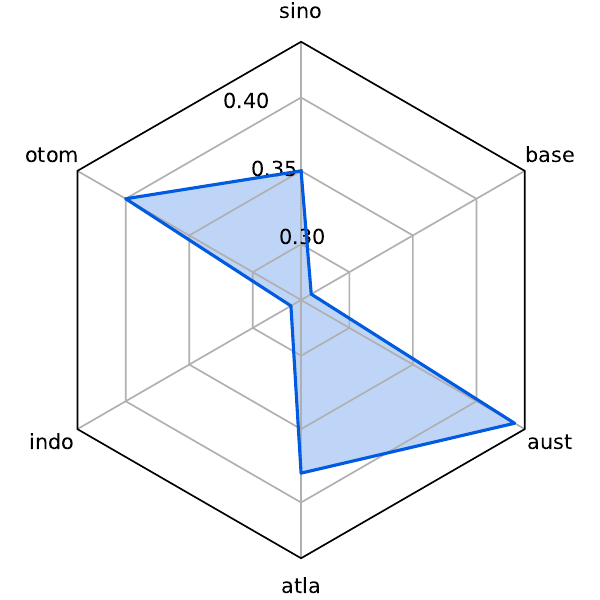}
    \includegraphics[width=0.32\textwidth]{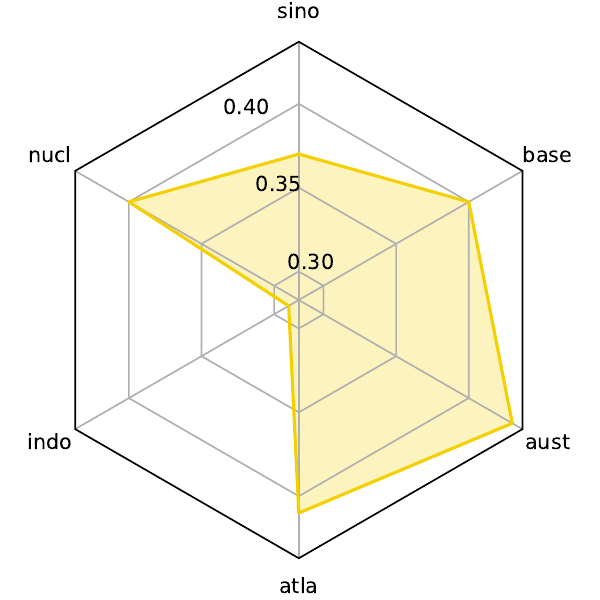}
    \includegraphics[width=0.32\textwidth]{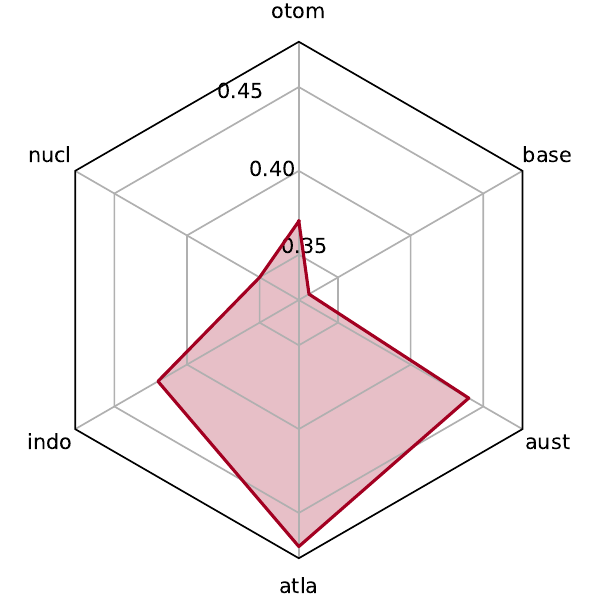}
\end{minipage}
\begin{minipage}{0.24\textwidth}
    \includegraphics[width=\textwidth]{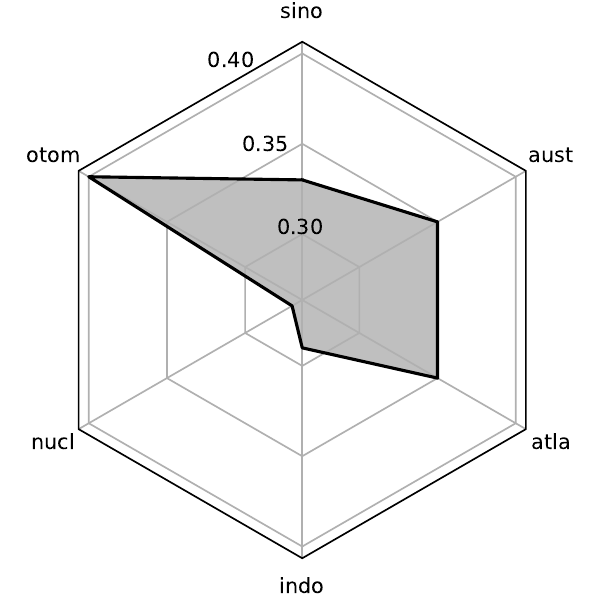}
\end{minipage}
% }
\caption{
%(Alternative to Figure \ref{fig:pairwise_ari})
Pairwise ARIs between language family-specific subnetworks. Each subfigure contains pairwise ARIs between one family (indicated by the color: \textcolor{orange}{\textbf{atla}}, \textcolor{pink}{\textbf{aust}}, \textcolor{green}{\textbf{indo}}, \textcolor{blue}{\textbf{nucl}}, \textcolor{yellow}{\textbf{otom}}, \textcolor{red}{\textbf{sino}}, \textbf{base}) and all other families (indicated on the edges). The ARIs are computed by averaging the results of 50 runs using the Louvain algorithm with different random states. Pairs of the same family, e.g., \textcolor{green}{\textbf{indo}}-\textcolor{green}{\textbf{indo}}, are not shown because the ARI will always be 1 in such cases. \textbf{base} is the graph including all edges, i.e., \networkone. Note that the scale is adjusted for each family individually.}
\label{fig:pairwise_ari_family}
\end{figure*}

\begin{figure*}
% \fbox{
\begin{minipage}{0.74\textwidth}
    % \hspace{-0.1cm}
    \includegraphics[width=0.32\textwidth]{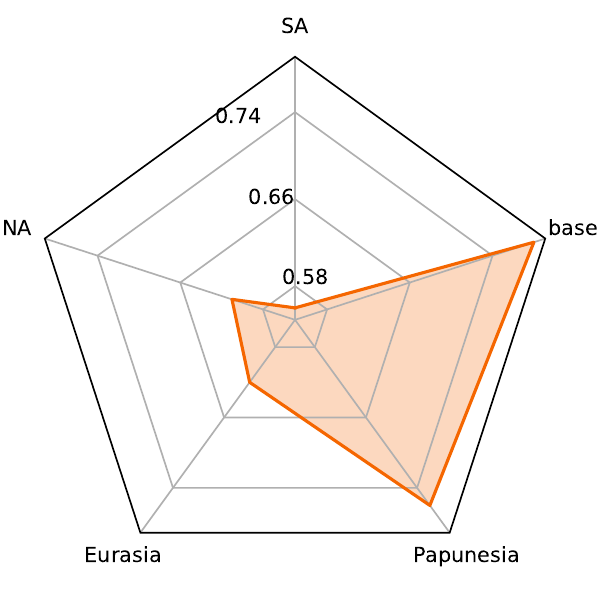}
    % \hspace{0.2cm}
    \includegraphics[width=0.32\textwidth]{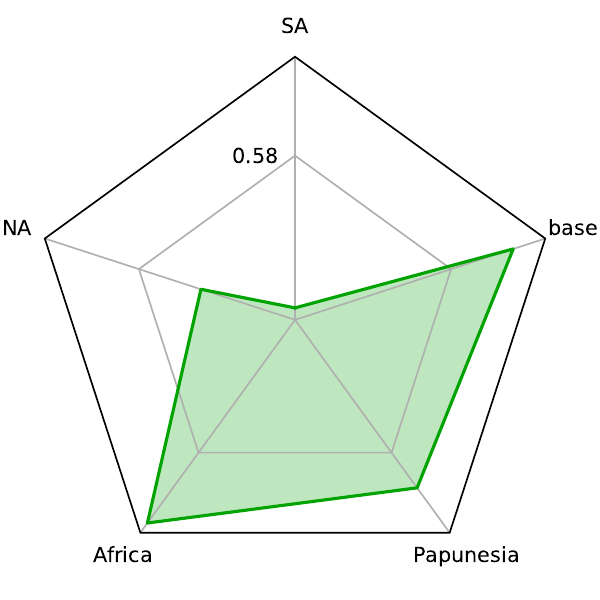}
    \includegraphics[width=0.32\textwidth]{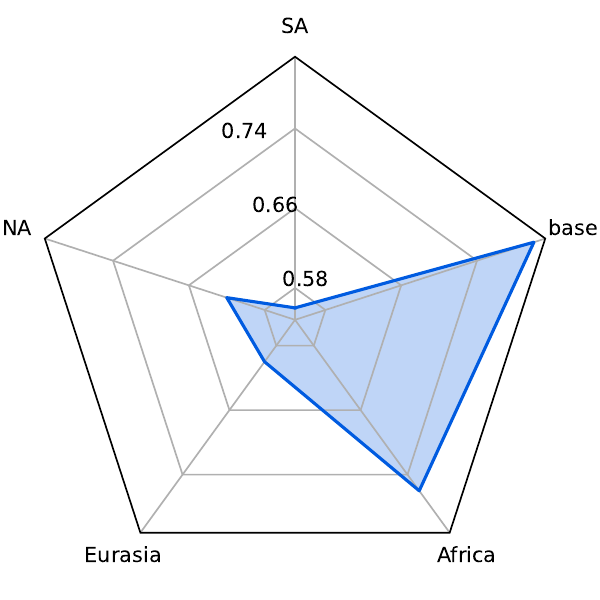}
    
    \includegraphics[width=0.32\textwidth]{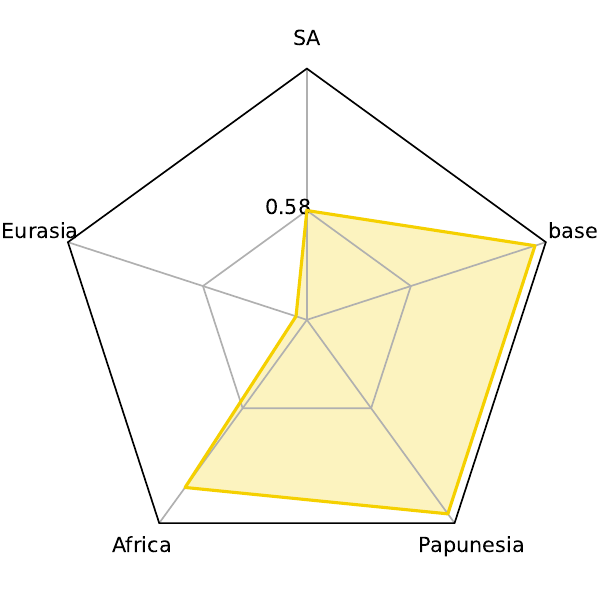}
    \includegraphics[width=0.32\textwidth]{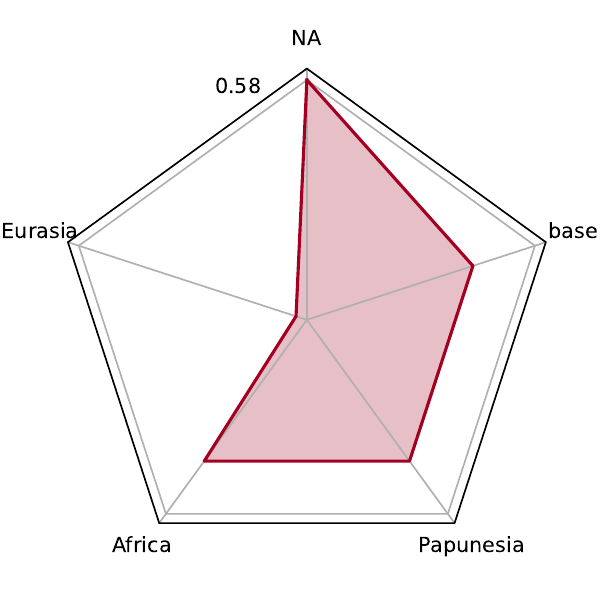}
\end{minipage}
\begin{minipage}{0.24\textwidth}
    \includegraphics[width=\textwidth]{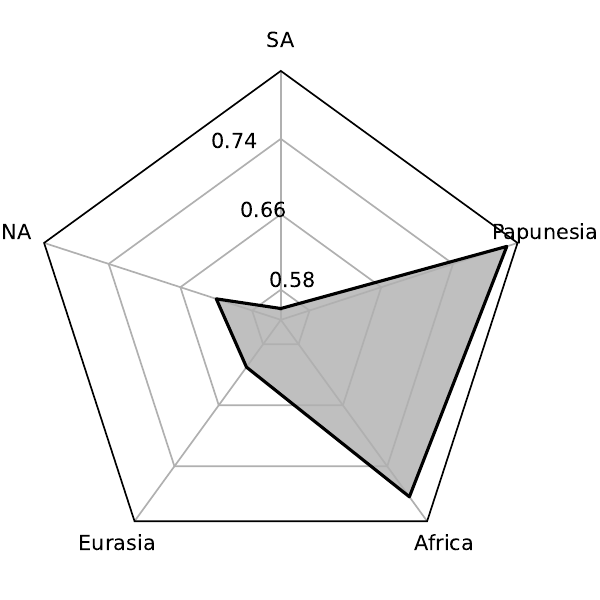}
\end{minipage}

% }
\caption{
%(Alternative to Figure \ref{fig:pairwise_ari})
Pairwise ARIs between area-specific subnetworks. Each subfigure contains pairwise ARIs between one area (indicated by the color: \textcolor{orange}{\textbf{Africa}}, \textcolor{green}{\textbf{Eurasia}}, \textcolor{blue}{\textbf{Papunesia}}, \textcolor{yellow}{\textbf{NA}}, \textcolor{red}{\textbf{SA}}, \textbf{base}) and all other areas (indicated on the edges). The ARIs are computed by averaging the results of 50 runs using the Louvain algorithm with different random states. Pairs of the same area, e.g., \textcolor{orange}{\textbf{Africa}}-\textcolor{orange}{\textbf{Africa}}, are not shown because the ARI will always be 1 in such cases. \textbf{base} is the graph including all edges, i.e., \networkone. Note that the scale is adjusted for each area individually.}
\label{fig:pairwise_ari_area}
\end{figure*}

\seclabel{families_areas}
We create
subnetworks specific to each language family and to each area.
We consider six language families that have more than 50
languages in PBC:
Austronesian (\textcolor{pink}{\textbf{aust}}),
Atlantic-Congo (\textcolor{orange}{\textbf{atla}}),
Indo-European (\textcolor{green}{\textbf{indo}}), Nuclear
Trans New Guinea (\textcolor{blue}{\textbf{nucl}}),
Otomanguean (\textcolor{yellow}{\textbf{otom}}) and
Sino-Tibetan (\textcolor{red}{\textbf{sino}}).
We consider five areas:
South America
(\textcolor{red}{\textbf{SA}}), North America
(\textcolor{yellow}{\textbf{NA}}), \textcolor{green}{\textbf{Eurasia}}, \textcolor{orange}{\textbf{Africa}}
and \textcolor{blue}{\textbf{Papunesia}}. We
only keep the edges in \networkone that occur in each
language family (resp.\ area) for the subnetwork of each
language family (resp.\ area). To quantify  agreement of
community structure, we use adjusted rand index
(ARI) \citep{hubert1985comparing,steinley2004properties}, similar
to \citep{jackson2019emotion}.
We
also compute ARI between \networkone and each
subnetwork.
Figures \ref{fig:pairwise_ari_family}
and \ref{fig:pairwise_ari_area}
show
pairwise ARI for language families
and areas.
It is clear
that any language family subnetwork cannot represent the
global colexification patterns encoded in \networkone, since
no family's ARI with \networkone is high. In
addition, no two language families have a similar
community structure according to ARI:
for the pair with
the highest ARI,
\textcolor{orange}{\textbf{atla}}-\textcolor{pink}{\textbf{aust}},
$\mbox{ARI}=0.5$. In comparison,
area-specific subnetworks generally have
larger pairwise ARIs.
The two areas
\textcolor{orange}{\textbf{Africa}}
and \textcolor{blue}{\textbf{Papunesia}} have a very high
ARI of 0.76 and also high ARIs with \networkone
(0.78 and 0.80). This can be explained by (1)
there are many languages in those two areas so there are
more possible colexifications included in the subnetworks
and (2) the diversity (in terms of colexification) of
languages spoken in these two areas is high. In  summary,
relatively low ARIs between families and areas also suggest
many colexification patterns are only specific to a small
group of languages (either in a specific language family or
in an area).

\section{English-centric transfer learning}\seclabel{english_transfer_learning}

\begin{sidewaysfigure*}
  \centering
  \includegraphics[width=0.8\columnwidth]{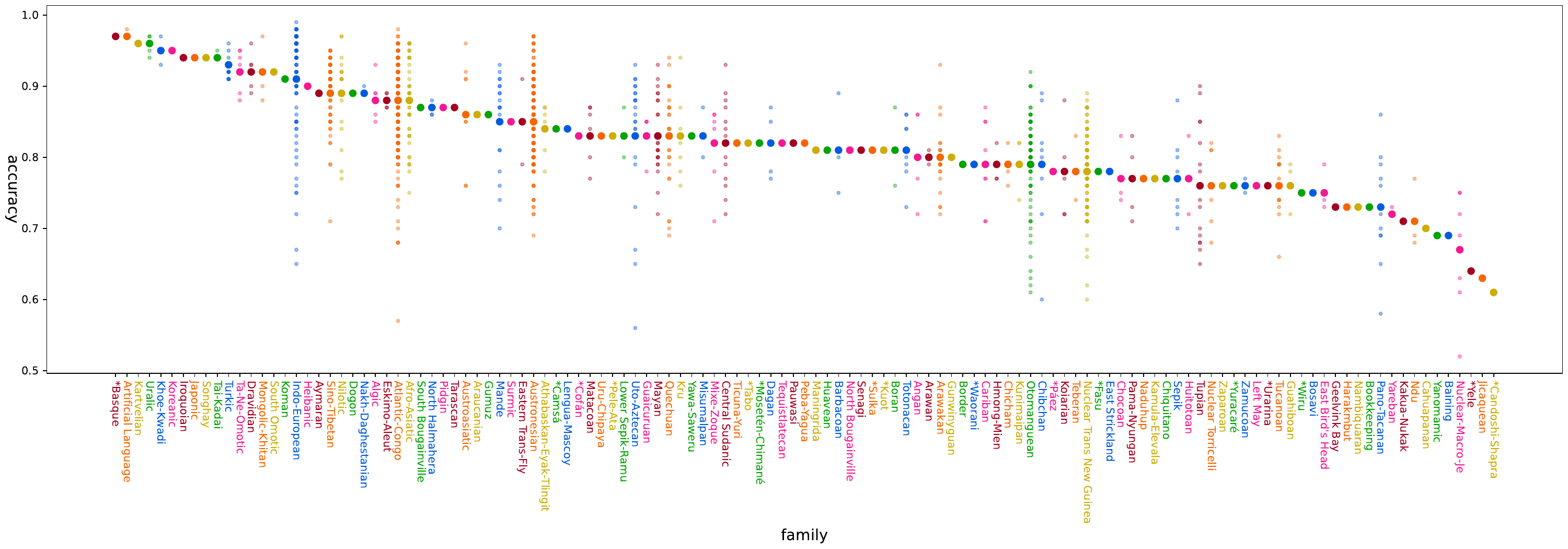}
  \includegraphics[width=0.8\columnwidth]{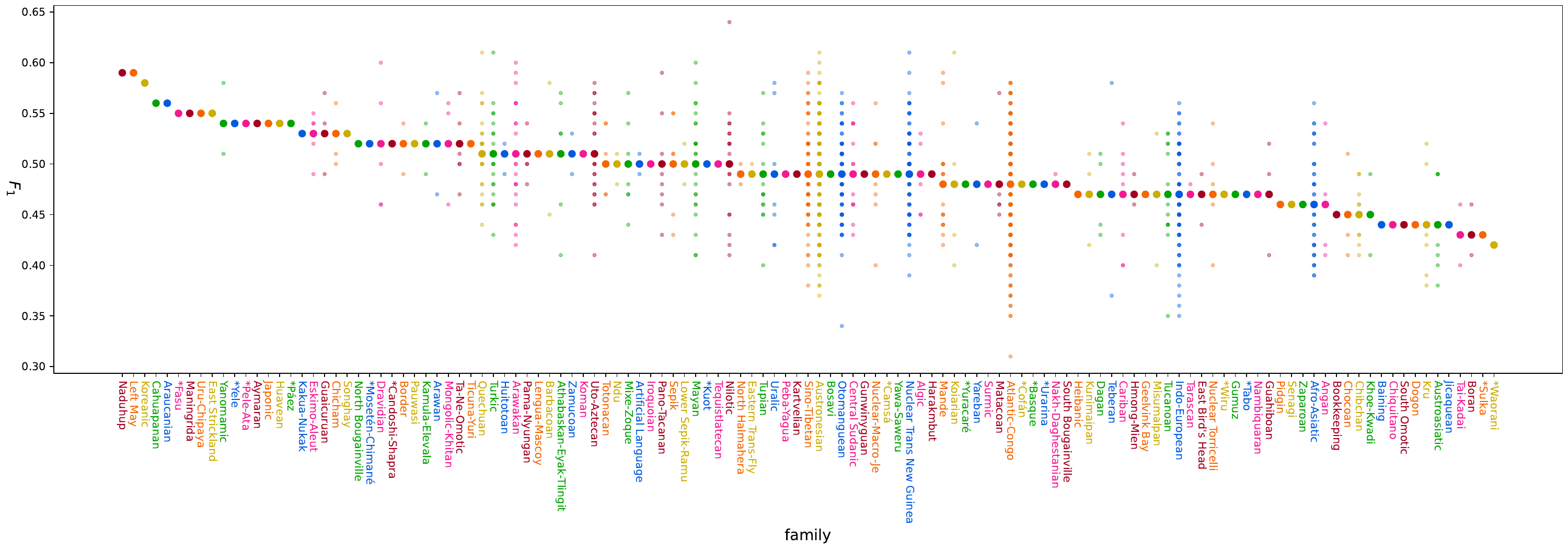}
  \caption{Top-10 accuracy of verse retrieval (top) and
  $F_1$ for verse classification (bottom) by language
  family for \embname. Each small dot represents a
  language, each large dot an average per family. Families
  are color-coded for better readability and sorted by
  accuracy/$F_1$. Asterisk (*) before
  name: language isolate. Most results are good 
  (average accuracy higher than 0.8 and average $F_1$ higher
  than 0.5 for half of the language families), but transfer performance varies 
  across language families.}
  \label{fig:sentence_retrieval}
\end{sidewaysfigure*}

We have shown the English-centric transfer performance of verse retrieval and verse classification averaged over languages in Table \ref{tab:results}. We believe it is also important to have a fine-grained view of the results for individual languages, to better understand the crosslingual transfer capability of \embname. Therefore, we show the transfer performance (sentence retrieval and sentence classification) of each individual language clustered by its corresponding language family in Figure \ref{fig:sentence_retrieval}. Globally, we see that results not only vary across language families but also vary within each language family. English

We find in Figure \ref{fig:sentence_retrieval} (top) that, though
a top-10 accuracy higher than 0.5 is achieved for all
languages, the average retrieval accuracy in the
Indo-European language family is slightly better than in
other families which have many languages 
(e.g., Sino-Tibetan or Otomanguean language family). 
We speculate this is probably because
other Indo-European languages can learn more accurate
alignments as our source language is English which also
belongs to the same family. Better alignments influence 
the quality of embeddings in that language,
therefore having an impact on the transfer performance.

The trend in classification as
shown in Figure \ref{fig:sentence_retrieval} (bottom),
however, is slightly different: average $F_1$  remains
stable at around 0.5 for almost all language families, with
less variance in each language family. This is evidence for
our conjecture that classification is a less difficult
task: apparently, good performance can be obtained if only
words referring to the most important concepts
that are highly associated with specific classes
are aligned well.

In summary, good performance indicates
that \embname assigns similar representations to ngrams that
refer to the same concept, thus improving crosslingual transfer.
\clearpage
\section{Beyond English-centric transfer}\seclabel{beyond_transfer}
We show the complete transfer performance by using any language as train/query language (1,245 languages in total, as we filtered some languages which have a very small size of train or test set). The results are shown in Table \ref{tab:table1}, \ref{tab:table2}, \ref{tab:table3}, \ref{tab:table4}, \ref{tab:table5}.

We hypothesize that the quality of identified colexification can influence the transfer learning performance. Some languages, because their morphology, typology, or conceptualization are very different from other languages, might pose difficulties in finding reliable colexification patterns, thus being detrimental to crosslingual transfer.  To this end, we compute the average colexification patterns per ngram (avg\_colex) for each language. That is, for language $l$, we compute the average number of neighbors of an ngram in \networktwo. The neighbors of an ngram node are concept nodes, which indicates the concepts that this ngram can refer to. The higher the avg\_colex is for a language, the more polysemous or ambiguous the ngrams tend to be. Of course, the extracted colexification patterns are not always correct due to verse-level misalignment, free translation, or some language-specific properties like morphology. Therefore the metric avg\_colex can, to some degree, indicate the level of difficulty to find correct alignments. 

We list the number of target-language ngrams in \networktwo (\#ngrams) as well as avg\_colex for the languages we show in \secref{main_transfer_beyond_eng}: Arabic (arb), Russian (rus), Chinese (zho), Apinayé (apn), Mündü (muh), Salt-Yui (sll) as well as the average over all languages in Table \ref{tab:avg_colex_per_ngram}. Three high-resource languages, which are typologically and morphologically different from each other, show similar trends in their statistics: more ngrams are included in \networktwo while avg\_colex is less than the average. This might indicate that the languages are less ambiguous and the colexifications extracted are mostly reliable, which explains good crosslingual performance when they are used as the train/query languages. On the contrary, the three worst-performing languages have exactly the inverse trend, which indicates it is harder to identify reliable colexifications, thus the performance is bad when they are served as the source languages.

To further test our hypothesis, we compute the Pearson correlation between the performance (classification $F_1$ score and retrieval accuracy) and avg\_colex. The results are shown in Table \ref{tab:pearson_test}. It is evident that \#ngrams is weakly positively correlated with the performance while avg\_colex is negatively correlated. However, it is important to note that the correlation is not high: there are quite a few languages that have small \#ngrams but large avg\_colex perform quite well when they are used as the source languages for large-scale transfer. For example, Bislama (bis), whose \#ngrams is only 1,202 but avg\_colex is 4.81, achieves good performance: 0.41, 0.46, 0.66, 0.73 for classification, retrieval top-1, top-5, and top-10 respectively. We speculate this is because Bislama is highly influenced by English \citep{tryon1987bislama}, therefore the patterns extracted are reliable since the concepts are represented in English lemmata. We leave the further exploration of finding reliable colexifications from a parallel corpus for future research.

To sum up, the quality of the colexification patterns extracted for a language is closely related to the transfer performance when it is served as the train/query language. Due to various language-specific properties, the model can have difficulties in inducing reliable colexification patterns.

\begin{table}
\centering
\setlength\tabcolsep{6pt}
\begin{tabular}{lrr}
\toprule
language & \#ngrams & avg\_colex \\
\midrule
arb & 4,107 & 1.84 \\
rus & 3,574 & 2.21 \\
zho & 3,659 & 2.07 \\
\midrule
apn & 2,119 & 2.91 \\
muh & 1,408 & 4.00 \\
sll & 2,118 & 2.93 \\
\midrule
avg. & 2,702 & 2.64 \\
\bottomrule
\end{tabular}
\caption{\label{tab:avg_colex_per_ngram} Number of target-language ngrams in \networktwo (\#ngrams) and the average number of colexified concepts per ngram (avg\_colex) for Arabic (arb), Russian (rus), Chinese (zho), Apinayé (apn), Mündü (muh), Salt-Yui (sll) as well as the average over all languages. We see that the lower three worst performing languages have fewer \#ngrams but larger avg\_colex than the average statistics over all languages.
}
\end{table}
\begin{table}
\centering
\setlength\tabcolsep{6pt}
\begin{tabular}{lrrrr}
\toprule
& c & r1 & r5 & r10 \\
\midrule
\#ngrams & 0.20 & 0.28 & 0.25 & 0.24 \\
avg\_colex & -0.18 & -0.25 & -0.21 & -0.19 \\
\bottomrule
\end{tabular}
\caption{\label{tab:pearson_test} Pearson correlations between \#ngrams/ avg\_colex and the transfer performance (c: classification $F_1$ score, r1: retrieval top-1 accuracy, r5: retrieval top-5 accuracy, r10: retrieval top-10 accuracy). All values are statistically significant under $p=0.01$.}
\end{table}

\begin{table*}
\centering
\small
    \resizebox{\textwidth}{!}{
    \begin{tabular}{lcccc|lcccc|lcccc}
        \toprule
    language & classification & \multicolumn{3}{c}{retrieval} & 
    language & classification & \multicolumn{3}{c}{retrieval} & 
    language & classification & \multicolumn{3}{c}{retrieval} \\
    \cmidrule(lr){2-2} \cmidrule(lr){3-5} \cmidrule(lr){7-7} \cmidrule(lr){8-10} \cmidrule(lr){12-12} \cmidrule(lr){13-15}
    & & top-1 & top-5 & top-10 & & & top-1 & top-5 & top-10 & & & top-1 & top-5 & top-10\\
    \midrule
aai & 0.46 & 0.47 & 0.65 & 0.72 & aak & 0.42 & 0.39 & 0.57 & 0.64 & aau & 0.45 & 0.47 & 0.67 & 0.74 \\
aaz & 0.40 & 0.39 & 0.58 & 0.65 & abt & 0.42 & 0.32 & 0.50 & 0.57 & abx & 0.44 & 0.52 & 0.70 & 0.76 \\
aby & 0.40 & 0.30 & 0.49 & 0.57 & acd & 0.44 & 0.39 & 0.58 & 0.66 & ace & 0.44 & 0.51 & 0.69 & 0.75 \\
acf & 0.43 & 0.52 & 0.71 & 0.77 & ach & 0.48 & 0.53 & 0.71 & 0.77 & acn & 0.44 & 0.51 & 0.70 & 0.76 \\
acr & 0.42 & 0.47 & 0.65 & 0.72 & acu & 0.43 & 0.39 & 0.58 & 0.66 & ade & 0.46 & 0.45 & 0.64 & 0.71 \\
adh & 0.47 & 0.53 & 0.71 & 0.77 & adi & 0.45 & 0.48 & 0.66 & 0.72 & adj & 0.43 & 0.42 & 0.61 & 0.68 \\
adl & 0.44 & 0.44 & 0.63 & 0.70 & aeb & 0.46 & 0.54 & 0.72 & 0.78 & aeu & 0.44 & 0.44 & 0.63 & 0.71 \\
aey & 0.46 & 0.47 & 0.65 & 0.72 & afr & 0.47 & 0.58 & 0.74 & 0.79 & agd & 0.43 & 0.38 & 0.58 & 0.66 \\
agg & 0.44 & 0.43 & 0.62 & 0.69 & agm & 0.44 & 0.39 & 0.57 & 0.64 & agn & 0.44 & 0.50 & 0.68 & 0.75 \\
agr & 0.41 & 0.43 & 0.62 & 0.69 & agt & 0.43 & 0.38 & 0.56 & 0.63 & agu & 0.42 & 0.40 & 0.59 & 0.66 \\
agw & 0.46 & 0.41 & 0.60 & 0.68 & ahk & 0.44 & 0.38 & 0.57 & 0.65 & aia & 0.42 & 0.44 & 0.63 & 0.71 \\
aii & 0.43 & 0.55 & 0.71 & 0.77 & aim & 0.42 & 0.41 & 0.61 & 0.68 & aji & 0.44 & 0.48 & 0.67 & 0.73 \\
ajz & 0.46 & 0.52 & 0.70 & 0.77 & akb & 0.43 & 0.51 & 0.70 & 0.76 & ake & 0.47 & 0.48 & 0.66 & 0.73 \\
akh & 0.46 & 0.31 & 0.49 & 0.58 & ald & 0.41 & 0.49 & 0.68 & 0.75 & alj & 0.45 & 0.46 & 0.65 & 0.73 \\
aln & 0.47 & 0.56 & 0.72 & 0.78 & alp & 0.41 & 0.41 & 0.60 & 0.68 & alq & 0.44 & 0.51 & 0.69 & 0.76 \\
alt & 0.47 & 0.58 & 0.75 & 0.81 & alz & 0.45 & 0.52 & 0.69 & 0.75 & ame & 0.40 & 0.40 & 0.58 & 0.66 \\
amf & 0.47 & 0.55 & 0.72 & 0.78 & amh & 0.44 & 0.55 & 0.73 & 0.78 & amk & 0.42 & 0.48 & 0.66 & 0.73 \\
amm & 0.43 & 0.39 & 0.59 & 0.66 & amn & 0.45 & 0.40 & 0.59 & 0.67 & amp & 0.40 & 0.34 & 0.52 & 0.60 \\
amr & 0.41 & 0.33 & 0.52 & 0.59 & amu & 0.42 & 0.46 & 0.64 & 0.71 & ann & 0.44 & 0.44 & 0.63 & 0.70 \\
anv & 0.44 & 0.42 & 0.61 & 0.68 & aoj & 0.45 & 0.41 & 0.60 & 0.67 & aom & 0.41 & 0.38 & 0.57 & 0.64 \\
aon & 0.42 & 0.39 & 0.58 & 0.66 & aoz & 0.51 & 0.52 & 0.70 & 0.76 & apb & 0.43 & 0.47 & 0.65 & 0.72 \\
ape & 0.39 & 0.40 & 0.60 & 0.67 & apn & 0.38 & 0.21 & 0.38 & 0.46 & apr & 0.44 & 0.40 & 0.60 & 0.68 \\
apt & 0.45 & 0.52 & 0.69 & 0.76 & apu & 0.45 & 0.44 & 0.62 & 0.69 & apw & 0.45 & 0.47 & 0.65 & 0.72 \\
apy & 0.42 & 0.35 & 0.53 & 0.61 & apz & 0.45 & 0.35 & 0.54 & 0.61 & arb & 0.47 & 0.56 & 0.72 & 0.78 \\
are & 0.41 & 0.41 & 0.59 & 0.67 & arl & 0.44 & 0.36 & 0.55 & 0.63 & arn & 0.41 & 0.48 & 0.67 & 0.73 \\
arz & 0.46 & 0.53 & 0.70 & 0.76 & asg & 0.42 & 0.41 & 0.60 & 0.67 & aso & 0.45 & 0.38 & 0.57 & 0.64 \\
ata & 0.46 & 0.44 & 0.64 & 0.71 & atb & 0.44 & 0.53 & 0.71 & 0.77 & atd & 0.45 & 0.44 & 0.64 & 0.71 \\
atg & 0.46 & 0.45 & 0.64 & 0.71 & att & 0.43 & 0.47 & 0.66 & 0.73 & auc & 0.45 & 0.38 & 0.57 & 0.64 \\
auy & 0.42 & 0.37 & 0.56 & 0.64 & ava & 0.46 & 0.52 & 0.70 & 0.76 & avt & 0.44 & 0.27 & 0.45 & 0.53 \\
avu & 0.42 & 0.29 & 0.48 & 0.56 & awa & 0.43 & 0.50 & 0.68 & 0.75 & awb & 0.41 & 0.42 & 0.60 & 0.67 \\
awi & 0.44 & 0.34 & 0.53 & 0.61 & ayo & 0.42 & 0.36 & 0.56 & 0.63 & ayr & 0.42 & 0.47 & 0.65 & 0.72 \\
aze & 0.47 & 0.53 & 0.70 & 0.76 & azg & 0.42 & 0.41 & 0.59 & 0.67 & azz & 0.45 & 0.45 & 0.64 & 0.71 \\
bak & 0.45 & 0.55 & 0.72 & 0.78 & bam & 0.47 & 0.57 & 0.74 & 0.80 & ban & 0.46 & 0.52 & 0.70 & 0.76 \\
bao & 0.47 & 0.39 & 0.58 & 0.66 & bar & 0.48 & 0.47 & 0.64 & 0.71 & bav & 0.47 & 0.42 & 0.62 & 0.69 \\
bba & 0.45 & 0.51 & 0.69 & 0.76 & bbb & 0.42 & 0.34 & 0.53 & 0.60 & bbj & 0.47 & 0.45 & 0.64 & 0.71 \\
bbr & 0.43 & 0.36 & 0.55 & 0.62 & bch & 0.44 & 0.44 & 0.64 & 0.71 & bci & 0.48 & 0.42 & 0.61 & 0.68 \\
bcl & 0.49 & 0.58 & 0.75 & 0.81 & bcw & 0.43 & 0.42 & 0.60 & 0.68 & bdd & 0.43 & 0.44 & 0.63 & 0.70 \\
bdh & 0.36 & 0.34 & 0.53 & 0.61 & bef & 0.46 & 0.34 & 0.53 & 0.60 & bel & 0.43 & 0.61 & 0.77 & 0.82 \\
bem & 0.45 & 0.52 & 0.70 & 0.76 & ben & 0.49 & 0.52 & 0.69 & 0.76 & beq & 0.47 & 0.56 & 0.73 & 0.79 \\
bex & 0.41 & 0.38 & 0.58 & 0.65 & bfd & 0.47 & 0.47 & 0.65 & 0.72 & bfo & 0.40 & 0.49 & 0.67 & 0.74 \\
bgr & 0.42 & 0.52 & 0.70 & 0.77 & bgs & 0.44 & 0.49 & 0.68 & 0.75 & bgz & 0.45 & 0.51 & 0.69 & 0.75 \\
bhl & 0.45 & 0.30 & 0.49 & 0.57 & bhp & 0.45 & 0.49 & 0.67 & 0.74 & bib & 0.41 & 0.46 & 0.65 & 0.72 \\
big & 0.48 & 0.38 & 0.56 & 0.64 & bim & 0.46 & 0.49 & 0.68 & 0.74 & bis & 0.41 & 0.46 & 0.66 & 0.73 \\
biu & 0.47 & 0.56 & 0.73 & 0.79 & biv & 0.42 & 0.46 & 0.64 & 0.71 & bjr & 0.45 & 0.33 & 0.52 & 0.60 \\
bjv & 0.40 & 0.37 & 0.57 & 0.65 & bkd & 0.47 & 0.49 & 0.68 & 0.75 & bkq & 0.44 & 0.33 & 0.51 & 0.59 \\
bku & 0.44 & 0.45 & 0.64 & 0.71 & bkv & 0.40 & 0.47 & 0.66 & 0.73 & blh & 0.40 & 0.43 & 0.63 & 0.70 \\
blw & 0.43 & 0.45 & 0.64 & 0.71 & blz & 0.45 & 0.54 & 0.72 & 0.78 & bmb & 0.46 & 0.56 & 0.73 & 0.79 \\
bmh & 0.41 & 0.32 & 0.51 & 0.59 & bmq & 0.43 & 0.45 & 0.65 & 0.71 & bmr & 0.46 & 0.46 & 0.65 & 0.72 \\
bmu & 0.46 & 0.44 & 0.63 & 0.70 & bnj & 0.41 & 0.47 & 0.67 & 0.74 & bnp & 0.47 & 0.41 & 0.60 & 0.67 \\
boa & 0.42 & 0.35 & 0.53 & 0.61 & boj & 0.41 & 0.41 & 0.60 & 0.68 & bom & 0.43 & 0.52 & 0.70 & 0.77 \\
bon & 0.45 & 0.38 & 0.58 & 0.66 & box & 0.43 & 0.46 & 0.66 & 0.73 & bpr & 0.42 & 0.50 & 0.69 & 0.75 \\
bps & 0.43 & 0.48 & 0.66 & 0.73 & bqc & 0.42 & 0.47 & 0.65 & 0.72 & bqj & 0.47 & 0.53 & 0.71 & 0.77 \\
bqp & 0.46 & 0.50 & 0.68 & 0.74 & bru & 0.43 & 0.40 & 0.60 & 0.68 & bsc & 0.44 & 0.54 & 0.72 & 0.78 \\
bsn & 0.41 & 0.31 & 0.47 & 0.54 & bss & 0.44 & 0.50 & 0.68 & 0.74 & btd & 0.47 & 0.50 & 0.68 & 0.75 \\
bth & 0.50 & 0.57 & 0.74 & 0.80 & bto & 0.50 & 0.57 & 0.74 & 0.80 & btt & 0.45 & 0.40 & 0.60 & 0.67 \\
btx & 0.47 & 0.57 & 0.74 & 0.80 & bud & 0.44 & 0.52 & 0.70 & 0.76 & bug & 0.47 & 0.52 & 0.70 & 0.76 \\
buk & 0.45 & 0.39 & 0.59 & 0.66 & bul & 0.46 & 0.57 & 0.74 & 0.79 & bum & 0.46 & 0.44 & 0.62 & 0.69 \\
bus & 0.45 & 0.51 & 0.69 & 0.75 & bvr & 0.41 & 0.38 & 0.57 & 0.65 & bvz & 0.41 & 0.32 & 0.51 & 0.59 \\
bwq & 0.46 & 0.48 & 0.67 & 0.74 & bwu & 0.46 & 0.42 & 0.61 & 0.69 & bxr & 0.45 & 0.54 & 0.71 & 0.77 \\
byr & 0.45 & 0.45 & 0.64 & 0.71 & byx & 0.43 & 0.29 & 0.48 & 0.56 & bzd & 0.44 & 0.45 & 0.64 & 0.71 \\
bzh & 0.39 & 0.39 & 0.59 & 0.66 & bzi & 0.42 & 0.46 & 0.65 & 0.72 & bzj & 0.46 & 0.48 & 0.66 & 0.73 \\
caa & 0.45 & 0.40 & 0.60 & 0.68 & cab & 0.42 & 0.49 & 0.66 & 0.73 & cac & 0.44 & 0.37 & 0.56 & 0.64 \\
caf & 0.44 & 0.42 & 0.61 & 0.68 & cag & 0.45 & 0.45 & 0.64 & 0.71 & cak & 0.43 & 0.41 & 0.60 & 0.68 \\
cao & 0.47 & 0.40 & 0.59 & 0.66 & cap & 0.42 & 0.44 & 0.63 & 0.70 & caq & 0.45 & 0.51 & 0.69 & 0.75 \\
car & 0.43 & 0.50 & 0.68 & 0.75 & cas & 0.45 & 0.44 & 0.64 & 0.71 & cat & 0.47 & 0.54 & 0.71 & 0.77 \\
cav & 0.37 & 0.33 & 0.52 & 0.60 & cax & 0.40 & 0.42 & 0.62 & 0.69 & cbc & 0.42 & 0.39 & 0.58 & 0.65 \\
cbi & 0.44 & 0.45 & 0.63 & 0.70 & cbk & 0.42 & 0.48 & 0.66 & 0.73 & cbr & 0.40 & 0.34 & 0.53 & 0.60 \\
cbs & 0.42 & 0.34 & 0.52 & 0.59 & cbt & 0.46 & 0.31 & 0.49 & 0.57 & cbu & 0.44 & 0.27 & 0.44 & 0.52 \\
cbv & 0.43 & 0.31 & 0.50 & 0.58 & cce & 0.48 & 0.56 & 0.73 & 0.79 & cco & 0.42 & 0.37 & 0.56 & 0.64 \\
ceb & 0.47 & 0.57 & 0.74 & 0.79 & ceg & 0.41 & 0.40 & 0.60 & 0.67 & ces & 0.47 & 0.57 & 0.73 & 0.79 \\
cfm & 0.48 & 0.44 & 0.63 & 0.70 & cgc & 0.47 & 0.46 & 0.65 & 0.72 & cha & 0.46 & 0.58 & 0.75 & 0.80 \\
chd & 0.42 & 0.42 & 0.60 & 0.67 & che & 0.45 & 0.44 & 0.63 & 0.70 & chf & 0.44 & 0.43 & 0.62 & 0.70 \\
chk & 0.46 & 0.50 & 0.69 & 0.76 & chq & 0.43 & 0.38 & 0.57 & 0.65 & chr & 0.46 & 0.51 & 0.69 & 0.75 \\
chu & 0.46 & 0.61 & 0.77 & 0.82 & chv & 0.48 & 0.53 & 0.71 & 0.77 & chz & 0.42 & 0.43 & 0.62 & 0.70 \\
cjo & 0.43 & 0.35 & 0.54 & 0.62 & cjp & 0.45 & 0.50 & 0.68 & 0.75 & cjv & 0.44 & 0.29 & 0.47 & 0.55 \\
ckb & 0.51 & 0.59 & 0.75 & 0.80 & cko & 0.44 & 0.46 & 0.65 & 0.72 & cle & 0.43 & 0.44 & 0.63 & 0.70 \\
clu & 0.44 & 0.51 & 0.69 & 0.75 & cly & 0.42 & 0.33 & 0.52 & 0.61 & cme & 0.41 & 0.42 & 0.61 & 0.68 \\
cmn & 0.49 & 0.60 & 0.76 & 0.82 & cmo & 0.41 & 0.46 & 0.65 & 0.73 & cnh & 0.47 & 0.45 & 0.63 & 0.70 \\
cni & 0.40 & 0.34 & 0.53 & 0.61 & cnl & 0.46 & 0.41 & 0.59 & 0.67 & cnt & 0.41 & 0.44 & 0.62 & 0.69 \\
cnw & 0.45 & 0.45 & 0.63 & 0.70 & coe & 0.43 & 0.34 & 0.53 & 0.61 & cof & 0.42 & 0.38 & 0.57 & 0.65 \\
cok & 0.45 & 0.39 & 0.58 & 0.66 & con & 0.40 & 0.43 & 0.62 & 0.69 & cop & 0.46 & 0.56 & 0.73 & 0.79 \\
cor & 0.50 & 0.60 & 0.76 & 0.81 & cot & 0.44 & 0.43 & 0.61 & 0.68 & cpa & 0.38 & 0.38 & 0.57 & 0.65 \\
cpb & 0.42 & 0.44 & 0.63 & 0.70 & cpc & 0.44 & 0.44 & 0.64 & 0.71 & cpu & 0.44 & 0.45 & 0.64 & 0.71 \\
cpy & 0.45 & 0.43 & 0.62 & 0.70 & crm & 0.44 & 0.54 & 0.71 & 0.77 & crn & 0.47 & 0.39 & 0.58 & 0.66 \\
crq & 0.42 & 0.40 & 0.59 & 0.66 & crs & 0.48 & 0.52 & 0.70 & 0.76 & crt & 0.43 & 0.41 & 0.60 & 0.67 \\
\bottomrule
\end{tabular}
    }
    \caption{Transfer performance using other languages as the train/query language (Part I).}\label{tab:table1}
\end{table*}

\begin{table*}
\centering
\small
    \resizebox{\textwidth}{!}{
    \begin{tabular}{lcccc|lcccc|lcccc}
        \toprule
    language & classification & \multicolumn{3}{c}{retrieval} & 
    language & classification & \multicolumn{3}{c}{retrieval} & 
    language & classification & \multicolumn{3}{c}{retrieval} \\
    \cmidrule(lr){2-2} \cmidrule(lr){3-5} \cmidrule(lr){7-7} \cmidrule(lr){8-10} \cmidrule(lr){12-12} \cmidrule(lr){13-15}
    & & top-1 & top-5 & top-10 & & & top-1 & top-5 & top-10 & & & top-1 & top-5 & top-10\\
    \midrule
    crx & 0.44 & 0.41 & 0.59 & 0.67 & csk & 0.45 & 0.54 & 0.72 & 0.78 & cso & 0.41 & 0.41 & 0.60 & 0.68 \\
csy & 0.47 & 0.47 & 0.65 & 0.72 & cta & 0.43 & 0.29 & 0.48 & 0.56 & ctd & 0.47 & 0.46 & 0.64 & 0.71 \\
ctp & 0.42 & 0.25 & 0.43 & 0.51 & ctu & 0.41 & 0.41 & 0.60 & 0.67 & cub & 0.44 & 0.40 & 0.59 & 0.66 \\
cuc & 0.47 & 0.47 & 0.65 & 0.72 & cui & 0.43 & 0.41 & 0.60 & 0.68 & cuk & 0.43 & 0.47 & 0.66 & 0.72 \\
cul & 0.45 & 0.40 & 0.59 & 0.67 & cut & 0.37 & 0.32 & 0.51 & 0.59 & cux & 0.44 & 0.39 & 0.58 & 0.66 \\
cwe & 0.49 & 0.54 & 0.71 & 0.77 & cwt & 0.45 & 0.54 & 0.72 & 0.77 & cya & 0.40 & 0.34 & 0.54 & 0.62 \\
cym & 0.49 & 0.52 & 0.70 & 0.76 & czt & 0.43 & 0.49 & 0.68 & 0.75 & daa & 0.43 & 0.45 & 0.64 & 0.72 \\
dad & 0.44 & 0.48 & 0.67 & 0.74 & dah & 0.43 & 0.38 & 0.57 & 0.65 & dan & 0.48 & 0.57 & 0.74 & 0.80 \\
ded & 0.47 & 0.47 & 0.66 & 0.73 & des & 0.45 & 0.40 & 0.57 & 0.65 & deu & 0.49 & 0.56 & 0.73 & 0.78 \\
dgc & 0.45 & 0.44 & 0.63 & 0.71 & dgi & 0.47 & 0.49 & 0.67 & 0.74 & dgr & 0.41 & 0.44 & 0.63 & 0.70 \\
dgz & 0.42 & 0.38 & 0.57 & 0.65 & dhm & 0.46 & 0.56 & 0.73 & 0.79 & dig & 0.45 & 0.50 & 0.69 & 0.75 \\
dik & 0.46 & 0.38 & 0.57 & 0.65 & dip & 0.44 & 0.47 & 0.66 & 0.72 & dis & 0.46 & 0.52 & 0.70 & 0.76 \\
dje & 0.47 & 0.56 & 0.74 & 0.80 & djk & 0.42 & 0.34 & 0.53 & 0.61 & djr & 0.44 & 0.42 & 0.61 & 0.68 \\
dnj & 0.45 & 0.35 & 0.53 & 0.61 & dob & 0.45 & 0.47 & 0.66 & 0.73 & dop & 0.43 & 0.42 & 0.61 & 0.68 \\
dow & 0.45 & 0.50 & 0.69 & 0.75 & dtp & 0.44 & 0.51 & 0.70 & 0.76 & dts & 0.46 & 0.52 & 0.70 & 0.77 \\
due & 0.45 & 0.48 & 0.67 & 0.74 & dug & 0.47 & 0.50 & 0.69 & 0.75 & duo & 0.43 & 0.43 & 0.63 & 0.70 \\
dur & 0.45 & 0.41 & 0.61 & 0.69 & dwr & 0.48 & 0.57 & 0.73 & 0.79 & dww & 0.43 & 0.49 & 0.68 & 0.75 \\
dyi & 0.44 & 0.50 & 0.68 & 0.75 & dyo & 0.49 & 0.52 & 0.71 & 0.77 & dyu & 0.46 & 0.49 & 0.68 & 0.74 \\
ebk & 0.47 & 0.54 & 0.72 & 0.78 & efi & 0.48 & 0.48 & 0.65 & 0.72 & eka & 0.48 & 0.48 & 0.67 & 0.74 \\
ell & 0.48 & 0.59 & 0.75 & 0.81 & emp & 0.41 & 0.47 & 0.66 & 0.73 & enb & 0.47 & 0.47 & 0.65 & 0.72 \\
eng & 0.49 & 0.65 & 0.80 & 0.84 & enl & 0.44 & 0.47 & 0.66 & 0.73 & enm & 0.46 & 0.59 & 0.75 & 0.81 \\
epo & 0.47 & 0.61 & 0.77 & 0.82 & eri & 0.42 & 0.42 & 0.61 & 0.68 & ese & 0.39 & 0.29 & 0.47 & 0.55 \\
esi & 0.44 & 0.52 & 0.70 & 0.77 & esk & 0.50 & 0.52 & 0.70 & 0.76 & est & 0.47 & 0.57 & 0.73 & 0.79 \\
esu & 0.46 & 0.54 & 0.72 & 0.78 & etu & 0.43 & 0.45 & 0.64 & 0.72 & eus & 0.50 & 0.59 & 0.75 & 0.81 \\
ewe & 0.50 & 0.50 & 0.68 & 0.74 & eza & 0.46 & 0.44 & 0.63 & 0.70 & faa & 0.44 & 0.36 & 0.54 & 0.61 \\
fai & 0.46 & 0.37 & 0.55 & 0.63 & fal & 0.43 & 0.52 & 0.70 & 0.77 & fao & 0.45 & 0.55 & 0.72 & 0.78 \\
ffm & 0.47 & 0.55 & 0.73 & 0.79 & fij & 0.44 & 0.50 & 0.70 & 0.76 & fil & 0.44 & 0.60 & 0.76 & 0.81 \\
fin & 0.46 & 0.59 & 0.76 & 0.81 & fon & 0.43 & 0.43 & 0.62 & 0.70 & for & 0.42 & 0.43 & 0.62 & 0.69 \\
fra & 0.47 & 0.56 & 0.73 & 0.78 & fry & 0.47 & 0.56 & 0.72 & 0.78 & fub & 0.47 & 0.54 & 0.71 & 0.78 \\
fuf & 0.46 & 0.56 & 0.73 & 0.79 & fuh & 0.50 & 0.55 & 0.73 & 0.79 & fuq & 0.47 & 0.56 & 0.73 & 0.79 \\
fuv & 0.46 & 0.54 & 0.71 & 0.78 & gaa & 0.46 & 0.53 & 0.70 & 0.76 & gag & 0.45 & 0.60 & 0.76 & 0.81 \\
gah & 0.46 & 0.40 & 0.58 & 0.66 & gam & 0.40 & 0.33 & 0.52 & 0.60 & gaw & 0.44 & 0.38 & 0.57 & 0.64 \\
gbi & 0.43 & 0.45 & 0.63 & 0.70 & gbo & 0.47 & 0.43 & 0.62 & 0.69 & gbr & 0.41 & 0.32 & 0.50 & 0.58 \\
gde & 0.43 & 0.50 & 0.68 & 0.75 & gdg & 0.45 & 0.45 & 0.64 & 0.71 & gdn & 0.44 & 0.40 & 0.59 & 0.67 \\
gdr & 0.45 & 0.48 & 0.67 & 0.74 & geb & 0.43 & 0.40 & 0.59 & 0.67 & gej & 0.46 & 0.51 & 0.68 & 0.75 \\
gfk & 0.45 & 0.46 & 0.65 & 0.72 & ghe & 0.49 & 0.54 & 0.72 & 0.78 & ghs & 0.43 & 0.33 & 0.51 & 0.59 \\
gid & 0.44 & 0.46 & 0.65 & 0.73 & gil & 0.48 & 0.51 & 0.69 & 0.75 & giz & 0.47 & 0.45 & 0.64 & 0.71 \\
gjn & 0.42 & 0.52 & 0.70 & 0.76 & gkn & 0.42 & 0.45 & 0.64 & 0.72 & gkp & 0.49 & 0.44 & 0.63 & 0.70 \\
gle & 0.47 & 0.48 & 0.66 & 0.72 & gmv & 0.46 & 0.55 & 0.72 & 0.78 & gnb & 0.46 & 0.49 & 0.67 & 0.74 \\
gnd & 0.37 & 0.37 & 0.57 & 0.65 & gng & 0.41 & 0.52 & 0.70 & 0.77 & gnn & 0.46 & 0.35 & 0.54 & 0.62 \\
gnw & 0.46 & 0.47 & 0.65 & 0.72 & gof & 0.46 & 0.57 & 0.73 & 0.79 & gog & 0.48 & 0.56 & 0.73 & 0.78 \\
gor & 0.47 & 0.52 & 0.69 & 0.75 & gqr & 0.41 & 0.38 & 0.58 & 0.66 & grt & 0.46 & 0.55 & 0.72 & 0.78 \\
gso & 0.47 & 0.48 & 0.67 & 0.73 & gub & 0.37 & 0.29 & 0.47 & 0.55 & guc & 0.40 & 0.38 & 0.56 & 0.64 \\
gud & 0.48 & 0.52 & 0.70 & 0.77 & guh & 0.43 & 0.40 & 0.58 & 0.66 & gui & 0.44 & 0.46 & 0.66 & 0.73 \\
guj & 0.44 & 0.46 & 0.64 & 0.71 & guk & 0.44 & 0.52 & 0.70 & 0.76 & gul & 0.45 & 0.48 & 0.66 & 0.73 \\
gum & 0.45 & 0.48 & 0.66 & 0.72 & gun & 0.41 & 0.49 & 0.67 & 0.74 & guo & 0.40 & 0.37 & 0.56 & 0.63 \\
guq & 0.43 & 0.40 & 0.60 & 0.67 & gur & 0.46 & 0.41 & 0.60 & 0.68 & guw & 0.49 & 0.55 & 0.71 & 0.77 \\
gux & 0.47 & 0.52 & 0.70 & 0.76 & guz & 0.46 & 0.53 & 0.70 & 0.76 & gvc & 0.42 & 0.41 & 0.60 & 0.67 \\
gvf & 0.43 & 0.30 & 0.49 & 0.57 & gvl & 0.49 & 0.44 & 0.63 & 0.71 & gvn & 0.43 & 0.37 & 0.55 & 0.63 \\
gwi & 0.42 & 0.39 & 0.58 & 0.65 & gya & 0.48 & 0.42 & 0.61 & 0.68 & gym & 0.40 & 0.39 & 0.58 & 0.66 \\
gyr & 0.46 & 0.44 & 0.63 & 0.70 & hae & 0.47 & 0.55 & 0.73 & 0.79 & hag & 0.43 & 0.42 & 0.62 & 0.70 \\
hak & 0.44 & 0.58 & 0.75 & 0.81 & hat & 0.47 & 0.41 & 0.61 & 0.68 & hau & 0.49 & 0.56 & 0.72 & 0.78 \\
haw & 0.42 & 0.50 & 0.69 & 0.75 & hay & 0.46 & 0.52 & 0.70 & 0.76 & hch & 0.44 & 0.51 & 0.69 & 0.75 \\
heb & 0.46 & 0.55 & 0.72 & 0.78 & heg & 0.45 & 0.36 & 0.55 & 0.63 & heh & 0.49 & 0.54 & 0.71 & 0.77 \\
hif & 0.46 & 0.45 & 0.63 & 0.70 & hig & 0.43 & 0.47 & 0.67 & 0.74 & hil & 0.47 & 0.60 & 0.77 & 0.82 \\
hin & 0.47 & 0.56 & 0.73 & 0.79 & hix & 0.41 & 0.40 & 0.59 & 0.66 & hla & 0.44 & 0.40 & 0.59 & 0.67 \\
hlt & 0.44 & 0.54 & 0.73 & 0.79 & hmo & 0.46 & 0.53 & 0.71 & 0.77 & hne & 0.48 & 0.55 & 0.72 & 0.78 \\
hnj & 0.50 & 0.45 & 0.64 & 0.72 & hnn & 0.42 & 0.50 & 0.68 & 0.75 & hns & 0.46 & 0.43 & 0.62 & 0.70 \\
hop & 0.45 & 0.49 & 0.67 & 0.74 & hot & 0.45 & 0.37 & 0.56 & 0.64 & hra & 0.44 & 0.49 & 0.67 & 0.74 \\
hrv & 0.46 & 0.58 & 0.75 & 0.80 & hto & 0.42 & 0.43 & 0.62 & 0.69 & hub & 0.44 & 0.37 & 0.56 & 0.64 \\
hui & 0.46 & 0.39 & 0.58 & 0.66 & hun & 0.49 & 0.57 & 0.73 & 0.79 & hus & 0.41 & 0.50 & 0.68 & 0.75 \\
huu & 0.41 & 0.35 & 0.54 & 0.62 & huv & 0.44 & 0.38 & 0.57 & 0.65 & hvn & 0.42 & 0.44 & 0.64 & 0.71 \\
hwc & 0.38 & 0.41 & 0.60 & 0.68 & hye & 0.46 & 0.56 & 0.73 & 0.79 & ian & 0.37 & 0.27 & 0.44 & 0.52 \\
iba & 0.46 & 0.58 & 0.75 & 0.81 & ibo & 0.47 & 0.46 & 0.65 & 0.72 & icr & 0.45 & 0.45 & 0.64 & 0.71 \\
ifa & 0.44 & 0.41 & 0.60 & 0.67 & ifb & 0.44 & 0.43 & 0.62 & 0.69 & ifk & 0.42 & 0.40 & 0.58 & 0.65 \\
ifu & 0.42 & 0.46 & 0.65 & 0.72 & ify & 0.48 & 0.32 & 0.50 & 0.58 & ign & 0.40 & 0.41 & 0.60 & 0.67 \\
ike & 0.47 & 0.55 & 0.72 & 0.78 & ikk & 0.44 & 0.54 & 0.72 & 0.78 & ikw & 0.44 & 0.53 & 0.71 & 0.77 \\
ilb & 0.48 & 0.57 & 0.74 & 0.80 & ilo & 0.47 & 0.56 & 0.73 & 0.78 & imo & 0.47 & 0.34 & 0.53 & 0.61 \\
inb & 0.47 & 0.45 & 0.64 & 0.71 & ind & 0.46 & 0.57 & 0.73 & 0.79 & ino & 0.42 & 0.39 & 0.58 & 0.66 \\
iou & 0.43 & 0.35 & 0.55 & 0.63 & ipi & 0.41 & 0.38 & 0.57 & 0.65 & iqw & 0.45 & 0.45 & 0.64 & 0.71 \\
iri & 0.40 & 0.46 & 0.65 & 0.72 & irk & 0.43 & 0.54 & 0.72 & 0.78 & iry & 0.45 & 0.53 & 0.71 & 0.78 \\
isd & 0.45 & 0.47 & 0.66 & 0.73 & isl & 0.46 & 0.58 & 0.75 & 0.80 & ita & 0.45 & 0.59 & 0.75 & 0.81 \\
itv & 0.46 & 0.56 & 0.74 & 0.79 & ium & 0.44 & 0.42 & 0.61 & 0.69 & ivb & 0.47 & 0.47 & 0.65 & 0.72 \\
ivv & 0.42 & 0.43 & 0.62 & 0.70 & iws & 0.39 & 0.35 & 0.54 & 0.62 & ixl & 0.41 & 0.41 & 0.59 & 0.67 \\
izr & 0.42 & 0.46 & 0.66 & 0.73 & izz & 0.45 & 0.47 & 0.66 & 0.73 & jac & 0.44 & 0.44 & 0.64 & 0.71 \\
jae & 0.45 & 0.49 & 0.68 & 0.75 & jam & 0.43 & 0.41 & 0.60 & 0.68 & jav & 0.47 & 0.48 & 0.67 & 0.73 \\
jbu & 0.45 & 0.37 & 0.56 & 0.64 & jic & 0.36 & 0.28 & 0.46 & 0.54 & jiv & 0.40 & 0.40 & 0.60 & 0.68 \\
jmc & 0.49 & 0.49 & 0.67 & 0.73 & jpn & 0.45 & 0.61 & 0.77 & 0.82 & jra & 0.43 & 0.53 & 0.72 & 0.78 \\
jvn & 0.46 & 0.45 & 0.63 & 0.70 & kaa & 0.45 & 0.54 & 0.71 & 0.77 & kab & 0.45 & 0.51 & 0.69 & 0.75 \\
kac & 0.41 & 0.41 & 0.60 & 0.68 & kal & 0.46 & 0.58 & 0.75 & 0.80 & kan & 0.44 & 0.60 & 0.77 & 0.82 \\
kao & 0.44 & 0.54 & 0.72 & 0.78 & kaq & 0.44 & 0.37 & 0.56 & 0.64 & kat & 0.46 & 0.59 & 0.76 & 0.81 \\
kaz & 0.45 & 0.55 & 0.72 & 0.78 & kbc & 0.40 & 0.47 & 0.66 & 0.73 & kbh & 0.46 & 0.47 & 0.65 & 0.72 \\
kbm & 0.42 & 0.33 & 0.52 & 0.60 & kbp & 0.43 & 0.51 & 0.69 & 0.76 & kbq & 0.43 & 0.44 & 0.62 & 0.70 \\
kbr & 0.43 & 0.53 & 0.71 & 0.76 & kck & 0.52 & 0.56 & 0.73 & 0.79 & kdc & 0.46 & 0.56 & 0.73 & 0.79 \\
kde & 0.48 & 0.59 & 0.75 & 0.81 & kdi & 0.45 & 0.54 & 0.72 & 0.78 & kdj & 0.45 & 0.57 & 0.74 & 0.80 \\
\bottomrule
\end{tabular}
    }
    \caption{Transfer performance using other languages as the train/query language (Part II).}\label{tab:table2}
\end{table*}

\begin{table*}
\centering
\small
    \resizebox{\textwidth}{!}{
    \begin{tabular}{lcccc|lcccc|lcccc}
        \toprule
    language & classification & \multicolumn{3}{c}{retrieval} & 
    language & classification & \multicolumn{3}{c}{retrieval} & 
    language & classification & \multicolumn{3}{c}{retrieval} \\
    \cmidrule(lr){2-2} \cmidrule(lr){3-5} \cmidrule(lr){7-7} \cmidrule(lr){8-10} \cmidrule(lr){12-12} \cmidrule(lr){13-15}
    & & top-1 & top-5 & top-10 & & & top-1 & top-5 & top-10 & & & top-1 & top-5 & top-10\\
    \midrule
    kdl & 0.41 & 0.42 & 0.61 & 0.69 & kek & 0.47 & 0.39 & 0.58 & 0.66 & ken & 0.47 & 0.46 & 0.65 & 0.72 \\
kew & 0.43 & 0.35 & 0.54 & 0.62 & kez & 0.43 & 0.46 & 0.65 & 0.72 & kff & 0.47 & 0.48 & 0.66 & 0.72 \\
kgf & 0.45 & 0.46 & 0.65 & 0.72 & kgk & 0.46 & 0.37 & 0.56 & 0.64 & kgp & 0.41 & 0.30 & 0.49 & 0.58 \\
khk & 0.47 & 0.57 & 0.74 & 0.80 & khm & 0.45 & 0.56 & 0.73 & 0.79 & khs & 0.44 & 0.32 & 0.51 & 0.59 \\
khy & 0.45 & 0.54 & 0.71 & 0.77 & khz & 0.46 & 0.50 & 0.69 & 0.75 & kia & 0.39 & 0.42 & 0.61 & 0.68 \\
kik & 0.48 & 0.54 & 0.71 & 0.77 & kin & 0.46 & 0.55 & 0.72 & 0.78 & kir & 0.46 & 0.53 & 0.70 & 0.76 \\
kix & 0.43 & 0.53 & 0.71 & 0.77 & kjb & 0.40 & 0.39 & 0.59 & 0.66 & kje & 0.42 & 0.38 & 0.58 & 0.66 \\
kjh & 0.46 & 0.52 & 0.70 & 0.76 & kjs & 0.43 & 0.37 & 0.56 & 0.64 & kki & 0.45 & 0.54 & 0.71 & 0.77 \\
kkj & 0.45 & 0.46 & 0.65 & 0.72 & klv & 0.43 & 0.49 & 0.67 & 0.74 & kma & 0.44 & 0.48 & 0.67 & 0.73 \\
kmg & 0.42 & 0.44 & 0.63 & 0.70 & kmh & 0.40 & 0.27 & 0.45 & 0.53 & kmk & 0.45 & 0.50 & 0.68 & 0.75 \\
kmm & 0.47 & 0.48 & 0.67 & 0.73 & kmo & 0.43 & 0.35 & 0.54 & 0.62 & kmr & 0.47 & 0.59 & 0.75 & 0.80 \\
kms & 0.41 & 0.32 & 0.51 & 0.59 & kmu & 0.40 & 0.39 & 0.59 & 0.66 & kne & 0.44 & 0.50 & 0.68 & 0.75 \\
knf & 0.44 & 0.52 & 0.71 & 0.77 & kng & 0.45 & 0.52 & 0.70 & 0.76 & knj & 0.40 & 0.44 & 0.63 & 0.71 \\
knk & 0.45 & 0.50 & 0.68 & 0.75 & kno & 0.37 & 0.34 & 0.53 & 0.61 & knv & 0.42 & 0.39 & 0.58 & 0.65 \\
kog & 0.40 & 0.41 & 0.60 & 0.67 & kor & 0.46 & 0.57 & 0.74 & 0.79 & kpf & 0.43 & 0.39 & 0.58 & 0.65 \\
kpg & 0.41 & 0.39 & 0.58 & 0.66 & kpj & 0.45 & 0.36 & 0.54 & 0.62 & kpr & 0.45 & 0.35 & 0.53 & 0.61 \\
kpv & 0.46 & 0.55 & 0.73 & 0.78 & kpw & 0.41 & 0.27 & 0.45 & 0.53 & kpx & 0.42 & 0.35 & 0.54 & 0.62 \\
kpz & 0.44 & 0.53 & 0.71 & 0.77 & kqe & 0.48 & 0.52 & 0.70 & 0.77 & kqo & 0.44 & 0.45 & 0.64 & 0.71 \\
kqp & 0.41 & 0.38 & 0.57 & 0.65 & kqs & 0.43 & 0.53 & 0.71 & 0.77 & kqy & 0.42 & 0.53 & 0.70 & 0.77 \\
krc & 0.47 & 0.57 & 0.74 & 0.79 & kri & 0.42 & 0.43 & 0.62 & 0.70 & krj & 0.43 & 0.57 & 0.74 & 0.80 \\
ksc & 0.45 & 0.45 & 0.64 & 0.71 & ksd & 0.46 & 0.50 & 0.68 & 0.75 & ksf & 0.48 & 0.51 & 0.69 & 0.76 \\
ksr & 0.43 & 0.44 & 0.63 & 0.70 & kss & 0.43 & 0.45 & 0.64 & 0.71 & ksw & 0.48 & 0.51 & 0.69 & 0.75 \\
ktb & 0.45 & 0.56 & 0.73 & 0.79 & ktj & 0.43 & 0.40 & 0.60 & 0.68 & kto & 0.42 & 0.41 & 0.61 & 0.68 \\
ktu & 0.47 & 0.55 & 0.72 & 0.78 & kua & 0.46 & 0.54 & 0.71 & 0.77 & kub & 0.44 & 0.40 & 0.60 & 0.67 \\
kud & 0.46 & 0.45 & 0.63 & 0.70 & kue & 0.47 & 0.42 & 0.62 & 0.69 & kum & 0.42 & 0.52 & 0.70 & 0.76 \\
kup & 0.41 & 0.31 & 0.49 & 0.57 & kus & 0.45 & 0.52 & 0.70 & 0.77 & kvj & 0.44 & 0.50 & 0.69 & 0.76 \\
kvn & 0.44 & 0.42 & 0.60 & 0.68 & kwd & 0.42 & 0.47 & 0.66 & 0.73 & kwf & 0.41 & 0.48 & 0.66 & 0.73 \\
kwi & 0.41 & 0.37 & 0.56 & 0.64 & kwj & 0.42 & 0.36 & 0.55 & 0.63 & kxc & 0.49 & 0.53 & 0.71 & 0.77 \\
kxm & 0.46 & 0.46 & 0.66 & 0.73 & kxw & 0.43 & 0.37 & 0.57 & 0.65 & kyc & 0.43 & 0.36 & 0.55 & 0.63 \\
kyf & 0.45 & 0.49 & 0.67 & 0.73 & kyg & 0.42 & 0.43 & 0.61 & 0.69 & kyq & 0.43 & 0.43 & 0.63 & 0.70 \\
kyu & 0.41 & 0.53 & 0.71 & 0.77 & kyz & 0.40 & 0.33 & 0.52 & 0.60 & kze & 0.47 & 0.39 & 0.58 & 0.65 \\
lac & 0.38 & 0.34 & 0.53 & 0.62 & lai & 0.46 & 0.57 & 0.75 & 0.80 & laj & 0.46 & 0.53 & 0.71 & 0.77 \\
lam & 0.43 & 0.56 & 0.73 & 0.78 & lao & 0.47 & 0.56 & 0.73 & 0.79 & las & 0.43 & 0.45 & 0.64 & 0.71 \\
lat & 0.46 & 0.57 & 0.74 & 0.80 & lav & 0.45 & 0.58 & 0.75 & 0.80 & lbk & 0.46 & 0.53 & 0.71 & 0.77 \\
lcm & 0.47 & 0.39 & 0.58 & 0.66 & ldi & 0.47 & 0.49 & 0.68 & 0.74 & lee & 0.48 & 0.44 & 0.63 & 0.70 \\
lef & 0.44 & 0.40 & 0.60 & 0.67 & leh & 0.46 & 0.51 & 0.69 & 0.75 & lem & 0.45 & 0.46 & 0.64 & 0.71 \\
leu & 0.47 & 0.45 & 0.65 & 0.72 & lew & 0.46 & 0.50 & 0.69 & 0.76 & lex & 0.46 & 0.41 & 0.60 & 0.68 \\
lgm & 0.43 & 0.55 & 0.72 & 0.78 & lhi & 0.44 & 0.39 & 0.59 & 0.67 & lhm & 0.43 & 0.49 & 0.68 & 0.74 \\
lhu & 0.48 & 0.42 & 0.62 & 0.70 & lia & 0.47 & 0.51 & 0.69 & 0.76 & lid & 0.43 & 0.35 & 0.54 & 0.62 \\
lif & 0.49 & 0.48 & 0.67 & 0.73 & lin & 0.47 & 0.53 & 0.71 & 0.77 & lip & 0.39 & 0.42 & 0.60 & 0.67 \\
lit & 0.49 & 0.58 & 0.75 & 0.81 & ljp & 0.48 & 0.54 & 0.71 & 0.78 & lmk & 0.45 & 0.52 & 0.69 & 0.76 \\
lmp & 0.46 & 0.47 & 0.66 & 0.74 & lob & 0.47 & 0.47 & 0.66 & 0.72 & lol & 0.48 & 0.50 & 0.68 & 0.74 \\
lom & 0.46 & 0.44 & 0.63 & 0.71 & loz & 0.48 & 0.52 & 0.70 & 0.76 & lsi & 0.42 & 0.40 & 0.59 & 0.67 \\
lsm & 0.47 & 0.52 & 0.70 & 0.76 & lug & 0.45 & 0.57 & 0.74 & 0.79 & luo & 0.46 & 0.50 & 0.68 & 0.74 \\
lus & 0.41 & 0.54 & 0.71 & 0.77 & lwo & 0.41 & 0.37 & 0.56 & 0.64 & lww & 0.45 & 0.31 & 0.50 & 0.58 \\
lzh & 0.47 & 0.62 & 0.77 & 0.82 & maa & 0.44 & 0.47 & 0.66 & 0.73 & mad & 0.46 & 0.52 & 0.70 & 0.76 \\
maf & 0.47 & 0.50 & 0.69 & 0.75 & mah & 0.46 & 0.50 & 0.69 & 0.75 & mai & 0.44 & 0.57 & 0.74 & 0.80 \\
maj & 0.45 & 0.50 & 0.69 & 0.75 & mak & 0.44 & 0.52 & 0.70 & 0.77 & mal & 0.45 & 0.56 & 0.72 & 0.78 \\
mam & 0.44 & 0.48 & 0.67 & 0.74 & maq & 0.40 & 0.41 & 0.59 & 0.67 & mar & 0.44 & 0.50 & 0.68 & 0.75 \\
mau & 0.45 & 0.40 & 0.59 & 0.67 & mav & 0.40 & 0.28 & 0.46 & 0.55 & maw & 0.45 & 0.47 & 0.66 & 0.73 \\
maz & 0.43 & 0.38 & 0.58 & 0.65 & mbb & 0.44 & 0.48 & 0.67 & 0.74 & mbc & 0.45 & 0.37 & 0.55 & 0.62 \\
mbd & 0.45 & 0.42 & 0.61 & 0.69 & mbf & 0.47 & 0.59 & 0.75 & 0.80 & mbh & 0.46 & 0.45 & 0.64 & 0.71 \\
mbi & 0.43 & 0.45 & 0.64 & 0.71 & mbj & 0.41 & 0.41 & 0.61 & 0.68 & mbl & 0.45 & 0.29 & 0.48 & 0.57 \\
mbs & 0.43 & 0.45 & 0.64 & 0.71 & mbt & 0.45 & 0.53 & 0.71 & 0.77 & mca & 0.45 & 0.48 & 0.67 & 0.74 \\
mcb & 0.39 & 0.39 & 0.58 & 0.65 & mcd & 0.42 & 0.33 & 0.51 & 0.59 & mcf & 0.41 & 0.22 & 0.39 & 0.47 \\
mck & 0.47 & 0.51 & 0.69 & 0.76 & mcn & 0.45 & 0.50 & 0.69 & 0.75 & mco & 0.40 & 0.42 & 0.60 & 0.68 \\
mcp & 0.45 & 0.50 & 0.68 & 0.75 & mcq & 0.45 & 0.42 & 0.62 & 0.69 & mcu & 0.44 & 0.39 & 0.58 & 0.66 \\
mda & 0.43 & 0.34 & 0.53 & 0.61 & mdy & 0.46 & 0.51 & 0.69 & 0.75 & med & 0.44 & 0.28 & 0.46 & 0.54 \\
mee & 0.43 & 0.48 & 0.67 & 0.73 & mej & 0.42 & 0.34 & 0.54 & 0.62 & mek & 0.42 & 0.43 & 0.63 & 0.70 \\
men & 0.45 & 0.48 & 0.67 & 0.73 & meq & 0.43 & 0.40 & 0.61 & 0.68 & meu & 0.46 & 0.48 & 0.67 & 0.74 \\
mfe & 0.47 & 0.54 & 0.71 & 0.77 & mfh & 0.46 & 0.40 & 0.59 & 0.66 & mfi & 0.45 & 0.43 & 0.62 & 0.69 \\
mfk & 0.44 & 0.44 & 0.63 & 0.70 & mfq & 0.45 & 0.48 & 0.66 & 0.73 & mfy & 0.45 & 0.44 & 0.62 & 0.69 \\
mfz & 0.43 & 0.43 & 0.62 & 0.69 & mhi & 0.44 & 0.40 & 0.59 & 0.66 & mhl & 0.44 & 0.38 & 0.57 & 0.64 \\
mhr & 0.45 & 0.57 & 0.74 & 0.79 & mhx & 0.45 & 0.39 & 0.59 & 0.66 & mhy & 0.43 & 0.52 & 0.70 & 0.76 \\
mib & 0.41 & 0.44 & 0.63 & 0.70 & mic & 0.46 & 0.44 & 0.63 & 0.70 & mie & 0.48 & 0.45 & 0.64 & 0.71 \\
mif & 0.44 & 0.45 & 0.64 & 0.71 & mig & 0.44 & 0.49 & 0.67 & 0.74 & mih & 0.43 & 0.35 & 0.55 & 0.63 \\
mil & 0.40 & 0.33 & 0.51 & 0.59 & min & 0.45 & 0.52 & 0.70 & 0.76 & mio & 0.39 & 0.32 & 0.52 & 0.60 \\
miq & 0.45 & 0.42 & 0.62 & 0.69 & mir & 0.38 & 0.34 & 0.51 & 0.59 & mit & 0.43 & 0.40 & 0.58 & 0.66 \\
miy & 0.42 & 0.42 & 0.62 & 0.70 & miz & 0.40 & 0.35 & 0.55 & 0.63 & mjc & 0.42 & 0.36 & 0.55 & 0.63 \\
mjw & 0.49 & 0.53 & 0.71 & 0.78 & mkd & 0.46 & 0.59 & 0.75 & 0.81 & mkl & 0.44 & 0.48 & 0.67 & 0.73 \\
mkn & 0.43 & 0.38 & 0.57 & 0.65 & mks & 0.44 & 0.38 & 0.58 & 0.66 & mlh & 0.48 & 0.43 & 0.63 & 0.70 \\
mlp & 0.45 & 0.40 & 0.59 & 0.67 & mlt & 0.49 & 0.57 & 0.74 & 0.79 & mmn & 0.44 & 0.43 & 0.62 & 0.70 \\
mmo & 0.46 & 0.41 & 0.59 & 0.67 & mmx & 0.41 & 0.44 & 0.63 & 0.70 & mna & 0.40 & 0.36 & 0.56 & 0.64 \\
mnb & 0.44 & 0.56 & 0.74 & 0.79 & mnf & 0.45 & 0.42 & 0.61 & 0.69 & mnh & 0.44 & 0.44 & 0.63 & 0.70 \\
mnk & 0.46 & 0.57 & 0.74 & 0.80 & mnx & 0.41 & 0.35 & 0.54 & 0.62 & moc & 0.44 & 0.48 & 0.65 & 0.72 \\
mog & 0.41 & 0.48 & 0.66 & 0.72 & mop & 0.41 & 0.40 & 0.59 & 0.67 & mor & 0.46 & 0.50 & 0.69 & 0.75 \\
mos & 0.46 & 0.44 & 0.63 & 0.70 & mox & 0.46 & 0.47 & 0.66 & 0.72 & mpg & 0.45 & 0.49 & 0.68 & 0.74 \\
mpm & 0.44 & 0.31 & 0.50 & 0.58 & mps & 0.41 & 0.26 & 0.43 & 0.52 & mpt & 0.41 & 0.35 & 0.53 & 0.61 \\
mqb & 0.47 & 0.38 & 0.58 & 0.66 & mqj & 0.45 & 0.53 & 0.71 & 0.77 & mqy & 0.46 & 0.49 & 0.68 & 0.74 \\
mri & 0.45 & 0.45 & 0.64 & 0.71 & mrw & 0.47 & 0.55 & 0.72 & 0.78 & msa & 0.47 & 0.58 & 0.75 & 0.81 \\
msb & 0.45 & 0.55 & 0.72 & 0.78 & mse & 0.48 & 0.44 & 0.63 & 0.70 & msk & 0.45 & 0.41 & 0.61 & 0.68 \\
msm & 0.43 & 0.49 & 0.67 & 0.74 & msy & 0.44 & 0.44 & 0.63 & 0.70 & mta & 0.44 & 0.44 & 0.63 & 0.70 \\
mtg & 0.44 & 0.36 & 0.55 & 0.63 & mti & 0.44 & 0.46 & 0.65 & 0.72 & mtj & 0.43 & 0.34 & 0.53 & 0.62 \\
mto & 0.40 & 0.47 & 0.66 & 0.73 & mtp & 0.47 & 0.50 & 0.68 & 0.74 & mua & 0.38 & 0.46 & 0.65 & 0.72 \\
muh & 0.38 & 0.22 & 0.39 & 0.48 & mur & 0.44 & 0.46 & 0.65 & 0.72 & mux & 0.42 & 0.37 & 0.55 & 0.63 \\
\bottomrule
\end{tabular}
    }
    \caption{Transfer performance using other languages as the train/query language (Part III).}\label{tab:table3}
\end{table*}

\begin{table*}
\centering
\small
    \resizebox{\textwidth}{!}{
    \begin{tabular}{lcccc|lcccc|lcccc}
        \toprule
    language & classification & \multicolumn{3}{c}{retrieval} & 
    language & classification & \multicolumn{3}{c}{retrieval} & 
    language & classification & \multicolumn{3}{c}{retrieval} \\
    \cmidrule(lr){2-2} \cmidrule(lr){3-5} \cmidrule(lr){7-7} \cmidrule(lr){8-10} \cmidrule(lr){12-12} \cmidrule(lr){13-15}
    & & top-1 & top-5 & top-10 & & & top-1 & top-5 & top-10 & & & top-1 & top-5 & top-10\\
    \midrule
muy & 0.43 & 0.43 & 0.62 & 0.69 & mva & 0.45 & 0.47 & 0.66 & 0.72 & mvn & 0.45 & 0.44 & 0.63 & 0.70 \\
mvp & 0.45 & 0.49 & 0.68 & 0.74 & mwm & 0.44 & 0.40 & 0.59 & 0.67 & mwq & 0.45 & 0.45 & 0.65 & 0.72 \\
mwv & 0.45 & 0.52 & 0.70 & 0.76 & mww & 0.40 & 0.45 & 0.65 & 0.72 & mxb & 0.39 & 0.42 & 0.61 & 0.69 \\
mxp & 0.44 & 0.37 & 0.56 & 0.63 & mxq & 0.43 & 0.45 & 0.64 & 0.71 & mxt & 0.43 & 0.37 & 0.56 & 0.64 \\
mya & 0.48 & 0.60 & 0.76 & 0.81 & myb & 0.43 & 0.48 & 0.67 & 0.73 & myk & 0.44 & 0.45 & 0.63 & 0.70 \\
myu & 0.41 & 0.38 & 0.58 & 0.65 & myv & 0.50 & 0.54 & 0.71 & 0.77 & myw & 0.40 & 0.41 & 0.60 & 0.68 \\
myx & 0.49 & 0.54 & 0.71 & 0.77 & myy & 0.44 & 0.33 & 0.51 & 0.59 & mza & 0.42 & 0.38 & 0.57 & 0.65 \\
mzh & 0.46 & 0.44 & 0.63 & 0.70 & mzk & 0.44 & 0.39 & 0.58 & 0.66 & mzl & 0.43 & 0.45 & 0.64 & 0.71 \\
mzm & 0.40 & 0.35 & 0.55 & 0.63 & mzw & 0.45 & 0.43 & 0.62 & 0.70 & nab & 0.38 & 0.34 & 0.53 & 0.61 \\
naf & 0.44 & 0.34 & 0.52 & 0.60 & nak & 0.42 & 0.32 & 0.51 & 0.59 & nan & 0.48 & 0.57 & 0.74 & 0.80 \\
naq & 0.49 & 0.57 & 0.74 & 0.80 & nas & 0.46 & 0.46 & 0.64 & 0.71 & nav & 0.46 & 0.50 & 0.69 & 0.76 \\
nbc & 0.45 & 0.54 & 0.72 & 0.78 & nbe & 0.46 & 0.55 & 0.72 & 0.78 & nbl & 0.50 & 0.57 & 0.73 & 0.79 \\
nca & 0.44 & 0.43 & 0.63 & 0.70 & nch & 0.45 & 0.47 & 0.65 & 0.72 & ncj & 0.42 & 0.45 & 0.63 & 0.70 \\
ncl & 0.44 & 0.40 & 0.58 & 0.66 & nct & 0.47 & 0.44 & 0.63 & 0.70 & ncu & 0.45 & 0.41 & 0.60 & 0.67 \\
ndc & 0.48 & 0.59 & 0.76 & 0.81 & nde & 0.49 & 0.56 & 0.73 & 0.79 & ndi & 0.43 & 0.46 & 0.65 & 0.72 \\
ndj & 0.45 & 0.55 & 0.72 & 0.78 & ndo & 0.46 & 0.55 & 0.72 & 0.78 & ndp & 0.51 & 0.52 & 0.70 & 0.76 \\
nds & 0.47 & 0.54 & 0.71 & 0.77 & ndz & 0.41 & 0.34 & 0.53 & 0.61 & neb & 0.42 & 0.43 & 0.62 & 0.69 \\
nep & 0.48 & 0.61 & 0.78 & 0.83 & nfa & 0.42 & 0.37 & 0.57 & 0.65 & nfr & 0.43 & 0.43 & 0.62 & 0.69 \\
ngc & 0.46 & 0.55 & 0.72 & 0.78 & ngp & 0.49 & 0.51 & 0.69 & 0.76 & ngu & 0.44 & 0.49 & 0.67 & 0.74 \\
nhd & 0.43 & 0.49 & 0.67 & 0.74 & nhe & 0.44 & 0.47 & 0.66 & 0.73 & nhg & 0.42 & 0.52 & 0.70 & 0.76 \\
nhi & 0.45 & 0.53 & 0.70 & 0.77 & nho & 0.47 & 0.39 & 0.59 & 0.66 & nhr & 0.47 & 0.54 & 0.72 & 0.78 \\
nhu & 0.45 & 0.46 & 0.65 & 0.72 & nhw & 0.42 & 0.47 & 0.66 & 0.72 & nhx & 0.41 & 0.47 & 0.66 & 0.73 \\
nhy & 0.46 & 0.51 & 0.69 & 0.75 & nii & 0.40 & 0.24 & 0.41 & 0.49 & nij & 0.48 & 0.50 & 0.68 & 0.75 \\
nim & 0.46 & 0.58 & 0.74 & 0.79 & nin & 0.44 & 0.36 & 0.55 & 0.63 & niq & 0.46 & 0.60 & 0.76 & 0.82 \\
niy & 0.43 & 0.51 & 0.69 & 0.75 & njb & 0.43 & 0.50 & 0.68 & 0.75 & njm & 0.43 & 0.50 & 0.67 & 0.74 \\
njn & 0.41 & 0.53 & 0.70 & 0.76 & njo & 0.45 & 0.56 & 0.73 & 0.78 & njz & 0.46 & 0.51 & 0.69 & 0.76 \\
nko & 0.45 & 0.50 & 0.69 & 0.75 & nlc & 0.43 & 0.38 & 0.57 & 0.64 & nld & 0.47 & 0.55 & 0.72 & 0.78 \\
nma & 0.44 & 0.53 & 0.70 & 0.76 & nmf & 0.44 & 0.51 & 0.69 & 0.75 & nmo & 0.48 & 0.50 & 0.68 & 0.75 \\
nmz & 0.40 & 0.48 & 0.67 & 0.73 & nnb & 0.46 & 0.59 & 0.76 & 0.81 & nng & 0.45 & 0.52 & 0.70 & 0.76 \\
nnh & 0.42 & 0.38 & 0.58 & 0.66 & nno & 0.46 & 0.55 & 0.72 & 0.78 & nnp & 0.48 & 0.48 & 0.66 & 0.73 \\
nnq & 0.46 & 0.53 & 0.71 & 0.77 & nnw & 0.42 & 0.43 & 0.63 & 0.70 & noa & 0.42 & 0.34 & 0.52 & 0.60 \\
nob & 0.48 & 0.58 & 0.74 & 0.80 & nog & 0.49 & 0.53 & 0.70 & 0.76 & nop & 0.42 & 0.41 & 0.60 & 0.68 \\
not & 0.42 & 0.37 & 0.56 & 0.64 & nou & 0.43 & 0.34 & 0.53 & 0.61 & nph & 0.46 & 0.56 & 0.74 & 0.79 \\
npi & 0.45 & 0.58 & 0.74 & 0.80 & npl & 0.46 & 0.49 & 0.67 & 0.73 & npo & 0.46 & 0.50 & 0.69 & 0.76 \\
npy & 0.44 & 0.49 & 0.67 & 0.74 & nsn & 0.46 & 0.46 & 0.65 & 0.72 & nso & 0.48 & 0.54 & 0.71 & 0.77 \\
ntp & 0.42 & 0.38 & 0.56 & 0.64 & ntr & 0.44 & 0.40 & 0.60 & 0.67 & nus & 0.47 & 0.50 & 0.68 & 0.74 \\
nuy & 0.48 & 0.44 & 0.63 & 0.70 & nvm & 0.45 & 0.34 & 0.52 & 0.59 & nwb & 0.42 & 0.39 & 0.58 & 0.65 \\
nwi & 0.44 & 0.39 & 0.58 & 0.66 & nya & 0.47 & 0.56 & 0.73 & 0.79 & nyf & 0.43 & 0.54 & 0.72 & 0.78 \\
nyn & 0.47 & 0.55 & 0.73 & 0.79 & nyo & 0.44 & 0.55 & 0.73 & 0.78 & nyy & 0.49 & 0.56 & 0.73 & 0.79 \\
obo & 0.47 & 0.50 & 0.68 & 0.75 & oji & 0.46 & 0.45 & 0.64 & 0.71 & ojs & 0.48 & 0.53 & 0.72 & 0.78 \\
okv & 0.43 & 0.32 & 0.51 & 0.59 & old & 0.45 & 0.50 & 0.68 & 0.74 & omw & 0.43 & 0.35 & 0.54 & 0.62 \\
ong & 0.42 & 0.31 & 0.50 & 0.58 & ons & 0.41 & 0.39 & 0.59 & 0.67 & ood & 0.44 & 0.36 & 0.55 & 0.63 \\
opm & 0.42 & 0.33 & 0.51 & 0.59 & ory & 0.44 & 0.52 & 0.69 & 0.76 & oss & 0.47 & 0.51 & 0.69 & 0.75 \\
ote & 0.47 & 0.43 & 0.62 & 0.69 & otm & 0.44 & 0.37 & 0.56 & 0.64 & otn & 0.40 & 0.42 & 0.61 & 0.68 \\
otq & 0.45 & 0.45 & 0.64 & 0.71 & ots & 0.43 & 0.31 & 0.49 & 0.57 & ozm & 0.47 & 0.46 & 0.65 & 0.72 \\
pab & 0.46 & 0.43 & 0.62 & 0.69 & pad & 0.42 & 0.41 & 0.61 & 0.68 & pag & 0.50 & 0.61 & 0.77 & 0.82 \\
pah & 0.41 & 0.39 & 0.59 & 0.66 & pam & 0.46 & 0.56 & 0.73 & 0.79 & pan & 0.47 & 0.54 & 0.72 & 0.78 \\
pao & 0.43 & 0.23 & 0.41 & 0.49 & pap & 0.47 & 0.58 & 0.75 & 0.80 & pbb & 0.42 & 0.39 & 0.58 & 0.65 \\
pbc & 0.46 & 0.47 & 0.66 & 0.73 & pbi & 0.41 & 0.44 & 0.63 & 0.70 & pbl & 0.48 & 0.52 & 0.70 & 0.76 \\
pcm & 0.48 & 0.48 & 0.67 & 0.73 & pdc & 0.47 & 0.53 & 0.71 & 0.77 & pdt & 0.48 & 0.52 & 0.70 & 0.77 \\
pes & 0.45 & 0.57 & 0.73 & 0.79 & pib & 0.45 & 0.57 & 0.75 & 0.80 & pio & 0.41 & 0.40 & 0.59 & 0.67 \\
pir & 0.46 & 0.39 & 0.58 & 0.65 & pis & 0.44 & 0.52 & 0.71 & 0.77 & pkb & 0.47 & 0.53 & 0.71 & 0.77 \\
plg & 0.42 & 0.47 & 0.66 & 0.73 & pls & 0.41 & 0.42 & 0.60 & 0.68 & plu & 0.44 & 0.42 & 0.61 & 0.68 \\
plw & 0.43 & 0.48 & 0.66 & 0.73 & pmf & 0.45 & 0.47 & 0.66 & 0.72 & pne & 0.45 & 0.45 & 0.65 & 0.72 \\
poe & 0.41 & 0.46 & 0.65 & 0.72 & poh & 0.47 & 0.39 & 0.59 & 0.66 & poi & 0.43 & 0.48 & 0.66 & 0.73 \\
pol & 0.47 & 0.57 & 0.73 & 0.79 & pon & 0.49 & 0.49 & 0.68 & 0.74 & por & 0.51 & 0.59 & 0.75 & 0.81 \\
poy & 0.45 & 0.56 & 0.73 & 0.79 & ppk & 0.48 & 0.50 & 0.69 & 0.75 & ppo & 0.46 & 0.38 & 0.57 & 0.65 \\
prf & 0.47 & 0.56 & 0.74 & 0.80 & pri & 0.46 & 0.45 & 0.65 & 0.72 & prk & 0.47 & 0.49 & 0.67 & 0.74 \\
prs & 0.44 & 0.55 & 0.73 & 0.78 & pse & 0.42 & 0.55 & 0.73 & 0.79 & ptp & 0.41 & 0.35 & 0.55 & 0.63 \\
ptu & 0.45 & 0.53 & 0.71 & 0.77 & pua & 0.45 & 0.45 & 0.63 & 0.71 & pwg & 0.44 & 0.46 & 0.66 & 0.73 \\
pww & 0.41 & 0.46 & 0.65 & 0.72 & qub & 0.43 & 0.37 & 0.56 & 0.64 & quc & 0.45 & 0.40 & 0.59 & 0.67 \\
quf & 0.40 & 0.41 & 0.59 & 0.66 & qug & 0.44 & 0.51 & 0.69 & 0.75 & quh & 0.47 & 0.53 & 0.70 & 0.77 \\
qul & 0.45 & 0.56 & 0.73 & 0.79 & qup & 0.48 & 0.38 & 0.57 & 0.65 & quw & 0.43 & 0.55 & 0.73 & 0.79 \\
quy & 0.45 & 0.50 & 0.69 & 0.75 & quz & 0.45 & 0.55 & 0.73 & 0.79 & qvc & 0.41 & 0.42 & 0.61 & 0.68 \\
qve & 0.45 & 0.49 & 0.67 & 0.74 & qvh & 0.41 & 0.34 & 0.52 & 0.60 & qvi & 0.48 & 0.51 & 0.70 & 0.76 \\
qvm & 0.40 & 0.34 & 0.52 & 0.60 & qvn & 0.43 & 0.41 & 0.59 & 0.66 & qvo & 0.44 & 0.46 & 0.64 & 0.70 \\
qvs & 0.40 & 0.41 & 0.59 & 0.66 & qvw & 0.42 & 0.45 & 0.64 & 0.71 & qvz & 0.43 & 0.42 & 0.60 & 0.68 \\
qwh & 0.47 & 0.48 & 0.66 & 0.72 & qxh & 0.40 & 0.36 & 0.55 & 0.62 & qxn & 0.44 & 0.47 & 0.67 & 0.73 \\
qxo & 0.42 & 0.32 & 0.50 & 0.58 & qxr & 0.42 & 0.50 & 0.69 & 0.76 & rai & 0.43 & 0.46 & 0.66 & 0.73 \\
rim & 0.47 & 0.51 & 0.69 & 0.76 & rkb & 0.40 & 0.30 & 0.48 & 0.56 & rmo & 0.42 & 0.40 & 0.59 & 0.67 \\
rmy & 0.46 & 0.52 & 0.69 & 0.75 & ron & 0.47 & 0.56 & 0.72 & 0.78 & roo & 0.46 & 0.42 & 0.61 & 0.68 \\
rop & 0.42 & 0.35 & 0.54 & 0.62 & rro & 0.43 & 0.49 & 0.67 & 0.74 & ruf & 0.44 & 0.55 & 0.73 & 0.79 \\
run & 0.50 & 0.54 & 0.72 & 0.78 & rus & 0.48 & 0.55 & 0.72 & 0.78 & rwo & 0.46 & 0.39 & 0.57 & 0.65 \\
sab & 0.37 & 0.28 & 0.47 & 0.55 & sag & 0.46 & 0.48 & 0.67 & 0.73 & sah & 0.46 & 0.53 & 0.71 & 0.77 \\
sas & 0.50 & 0.55 & 0.72 & 0.78 & sba & 0.45 & 0.40 & 0.59 & 0.66 & sbd & 0.44 & 0.41 & 0.60 & 0.68 \\
sbl & 0.43 & 0.51 & 0.69 & 0.76 & sda & 0.46 & 0.56 & 0.73 & 0.79 & sey & 0.43 & 0.43 & 0.63 & 0.70 \\
sgb & 0.42 & 0.43 & 0.61 & 0.69 & sgw & 0.45 & 0.51 & 0.70 & 0.76 & sgz & 0.41 & 0.38 & 0.57 & 0.65 \\
shi & 0.45 & 0.51 & 0.69 & 0.75 & shk & 0.44 & 0.48 & 0.67 & 0.73 & shp & 0.46 & 0.44 & 0.62 & 0.69 \\
shu & 0.45 & 0.50 & 0.69 & 0.75 & sig & 0.41 & 0.44 & 0.63 & 0.71 & sil & 0.43 & 0.44 & 0.64 & 0.71 \\
sim & 0.44 & 0.37 & 0.56 & 0.64 & sin & 0.48 & 0.49 & 0.66 & 0.73 & sja & 0.43 & 0.37 & 0.55 & 0.63 \\
sld & 0.46 & 0.47 & 0.66 & 0.73 & slk & 0.48 & 0.58 & 0.74 & 0.80 & sll & 0.39 & 0.24 & 0.42 & 0.51 \\
slv & 0.49 & 0.56 & 0.73 & 0.79 & sme & 0.47 & 0.62 & 0.78 & 0.83 & smk & 0.39 & 0.47 & 0.65 & 0.72 \\
sml & 0.45 & 0.50 & 0.69 & 0.76 & smo & 0.46 & 0.52 & 0.70 & 0.76 & sna & 0.49 & 0.56 & 0.73 & 0.79 \\
snc & 0.47 & 0.52 & 0.70 & 0.76 & snd & 0.44 & 0.55 & 0.73 & 0.78 & snn & 0.42 & 0.30 & 0.49 & 0.57 \\
snp & 0.43 & 0.42 & 0.61 & 0.68 & snw & 0.45 & 0.46 & 0.64 & 0.71 & sny & 0.38 & 0.30 & 0.49 & 0.57 \\
\bottomrule
\end{tabular}
    }
    \caption{Transfer performance using other languages as the train/query language (Part IV).}\label{tab:table4}
\end{table*}

\begin{table*}
\centering
\small
    \resizebox{\textwidth}{!}{
    \begin{tabular}{lcccc|lcccc|lcccc}
        \toprule
    language & classification & \multicolumn{3}{c}{retrieval} & 
    language & classification & \multicolumn{3}{c}{retrieval} & 
    language & classification & \multicolumn{3}{c}{retrieval} \\
    \cmidrule(lr){2-2} \cmidrule(lr){3-5} \cmidrule(lr){7-7} \cmidrule(lr){8-10} \cmidrule(lr){12-12} \cmidrule(lr){13-15}
    & & top-1 & top-5 & top-10 & & & top-1 & top-5 & top-10 & & & top-1 & top-5 & top-10\\
    \midrule
    som & 0.48 & 0.57 & 0.74 & 0.80 & soq & 0.47 & 0.43 & 0.62 & 0.69 & sot & 0.47 & 0.52 & 0.69 & 0.75 \\
soy & 0.45 & 0.48 & 0.67 & 0.74 & spa & 0.47 & 0.58 & 0.74 & 0.80 & spl & 0.38 & 0.25 & 0.42 & 0.50 \\
spp & 0.47 & 0.48 & 0.66 & 0.73 & sps & 0.44 & 0.42 & 0.60 & 0.67 & spy & 0.47 & 0.48 & 0.67 & 0.73 \\
sqi & 0.42 & 0.32 & 0.45 & 0.51 & sri & 0.42 & 0.43 & 0.62 & 0.69 & srm & 0.43 & 0.34 & 0.53 & 0.61 \\
srn & 0.48 & 0.50 & 0.68 & 0.74 & srp & 0.46 & 0.59 & 0.75 & 0.80 & srq & 0.41 & 0.31 & 0.50 & 0.59 \\
ssd & 0.47 & 0.38 & 0.58 & 0.65 & ssg & 0.41 & 0.37 & 0.56 & 0.64 & ssw & 0.47 & 0.57 & 0.74 & 0.80 \\
ssx & 0.45 & 0.39 & 0.58 & 0.65 & stn & 0.45 & 0.43 & 0.62 & 0.69 & stp & 0.39 & 0.27 & 0.45 & 0.53 \\
sua & 0.43 & 0.38 & 0.56 & 0.64 & sue & 0.42 & 0.36 & 0.56 & 0.64 & suk & 0.43 & 0.54 & 0.71 & 0.77 \\
sun & 0.43 & 0.48 & 0.66 & 0.73 & sur & 0.40 & 0.49 & 0.68 & 0.74 & sus & 0.39 & 0.51 & 0.70 & 0.77 \\
swe & 0.45 & 0.55 & 0.72 & 0.78 & swg & 0.48 & 0.57 & 0.73 & 0.79 & swh & 0.47 & 0.55 & 0.72 & 0.78 \\
swk & 0.45 & 0.57 & 0.75 & 0.80 & swp & 0.44 & 0.51 & 0.69 & 0.75 & sxb & 0.44 & 0.51 & 0.69 & 0.76 \\
sxn & 0.45 & 0.50 & 0.68 & 0.75 & syc & 0.49 & 0.53 & 0.70 & 0.76 & szb & 0.40 & 0.34 & 0.52 & 0.60 \\
tab & 0.45 & 0.51 & 0.69 & 0.75 & tac & 0.42 & 0.28 & 0.45 & 0.53 & taj & 0.47 & 0.46 & 0.65 & 0.72 \\
tam & 0.43 & 0.60 & 0.77 & 0.82 & taq & 0.44 & 0.48 & 0.66 & 0.73 & tar & 0.42 & 0.38 & 0.58 & 0.65 \\
tat & 0.46 & 0.56 & 0.74 & 0.79 & tav & 0.47 & 0.32 & 0.51 & 0.60 & taw & 0.39 & 0.33 & 0.52 & 0.60 \\
tbc & 0.47 & 0.38 & 0.58 & 0.65 & tbg & 0.41 & 0.35 & 0.54 & 0.62 & tbl & 0.43 & 0.45 & 0.64 & 0.72 \\
tbo & 0.45 & 0.46 & 0.66 & 0.73 & tby & 0.44 & 0.51 & 0.69 & 0.75 & tbz & 0.44 & 0.40 & 0.60 & 0.67 \\
tca & 0.44 & 0.41 & 0.60 & 0.67 & tcc & 0.49 & 0.52 & 0.69 & 0.76 & tcs & 0.40 & 0.43 & 0.63 & 0.70 \\
tcz & 0.45 & 0.52 & 0.69 & 0.76 & tdt & 0.45 & 0.53 & 0.71 & 0.77 & tee & 0.46 & 0.42 & 0.60 & 0.67 \\
tel & 0.46 & 0.54 & 0.72 & 0.78 & tem & 0.46 & 0.56 & 0.73 & 0.79 & teo & 0.49 & 0.57 & 0.74 & 0.79 \\
ter & 0.45 & 0.43 & 0.63 & 0.70 & tfr & 0.40 & 0.40 & 0.59 & 0.67 & tgk & 0.46 & 0.59 & 0.75 & 0.80 \\
tgl & 0.46 & 0.55 & 0.72 & 0.78 & tgp & 0.47 & 0.44 & 0.64 & 0.71 & tha & 0.49 & 0.54 & 0.72 & 0.78 \\
thk & 0.46 & 0.52 & 0.70 & 0.76 & tih & 0.47 & 0.51 & 0.69 & 0.76 & tik & 0.45 & 0.41 & 0.60 & 0.67 \\
tim & 0.44 & 0.30 & 0.49 & 0.57 & tir & 0.46 & 0.59 & 0.76 & 0.81 & tku & 0.42 & 0.44 & 0.63 & 0.70 \\
tlb & 0.47 & 0.51 & 0.69 & 0.75 & tlf & 0.44 & 0.30 & 0.48 & 0.56 & tlh & 0.50 & 0.58 & 0.75 & 0.80 \\
tna & 0.45 & 0.40 & 0.59 & 0.67 & tnn & 0.42 & 0.38 & 0.58 & 0.66 & tob & 0.43 & 0.39 & 0.59 & 0.67 \\
toc & 0.41 & 0.40 & 0.58 & 0.66 & toh & 0.49 & 0.54 & 0.71 & 0.77 & toj & 0.41 & 0.35 & 0.53 & 0.61 \\
too & 0.42 & 0.46 & 0.65 & 0.72 & top & 0.41 & 0.36 & 0.53 & 0.61 & tos & 0.40 & 0.39 & 0.58 & 0.65 \\
tpi & 0.42 & 0.48 & 0.68 & 0.75 & tpm & 0.41 & 0.31 & 0.51 & 0.59 & tpp & 0.45 & 0.50 & 0.68 & 0.74 \\
tpt & 0.44 & 0.45 & 0.63 & 0.70 & tpz & 0.46 & 0.38 & 0.58 & 0.66 & tqb & 0.39 & 0.29 & 0.47 & 0.55 \\
trc & 0.40 & 0.32 & 0.51 & 0.60 & trn & 0.43 & 0.47 & 0.66 & 0.73 & trq & 0.44 & 0.38 & 0.57 & 0.65 \\
tsn & 0.46 & 0.57 & 0.74 & 0.80 & tsz & 0.44 & 0.44 & 0.63 & 0.70 & ttc & 0.43 & 0.44 & 0.63 & 0.70 \\
tte & 0.45 & 0.46 & 0.65 & 0.72 & tuc & 0.42 & 0.39 & 0.58 & 0.66 & tue & 0.44 & 0.44 & 0.62 & 0.70 \\
tuf & 0.40 & 0.34 & 0.52 & 0.60 & tui & 0.43 & 0.38 & 0.57 & 0.65 & tuk & 0.46 & 0.52 & 0.70 & 0.76 \\
tum & 0.47 & 0.59 & 0.76 & 0.81 & tuo & 0.45 & 0.39 & 0.58 & 0.66 & tur & 0.45 & 0.55 & 0.72 & 0.78 \\
twi & 0.48 & 0.45 & 0.63 & 0.70 & twu & 0.44 & 0.47 & 0.66 & 0.73 & txu & 0.43 & 0.31 & 0.50 & 0.58 \\
tyv & 0.43 & 0.54 & 0.72 & 0.78 & tzh & 0.45 & 0.52 & 0.70 & 0.77 & tzj & 0.44 & 0.44 & 0.63 & 0.70 \\
tzo & 0.46 & 0.51 & 0.68 & 0.75 & ubr & 0.48 & 0.52 & 0.70 & 0.76 & ubu & 0.44 & 0.39 & 0.57 & 0.64 \\
udu & 0.46 & 0.44 & 0.62 & 0.70 & uig & 0.46 & 0.52 & 0.70 & 0.76 & ukr & 0.48 & 0.57 & 0.74 & 0.79 \\
upv & 0.44 & 0.42 & 0.61 & 0.68 & ura & 0.43 & 0.35 & 0.53 & 0.61 & urb & 0.39 & 0.33 & 0.53 & 0.61 \\
urd & 0.47 & 0.57 & 0.74 & 0.80 & urk & 0.45 & 0.49 & 0.69 & 0.76 & usa & 0.43 & 0.40 & 0.59 & 0.66 \\
usp & 0.38 & 0.46 & 0.65 & 0.72 & uvl & 0.42 & 0.39 & 0.59 & 0.67 & uzb & 0.44 & 0.57 & 0.74 & 0.80 \\
vag & 0.42 & 0.47 & 0.66 & 0.73 & ven & 0.47 & 0.55 & 0.72 & 0.78 & vie & 0.49 & 0.50 & 0.68 & 0.75 \\
viv & 0.44 & 0.47 & 0.66 & 0.72 & vmy & 0.44 & 0.49 & 0.68 & 0.74 & vun & 0.46 & 0.54 & 0.71 & 0.77 \\
vut & 0.45 & 0.47 & 0.66 & 0.73 & waj & 0.43 & 0.41 & 0.60 & 0.67 & wal & 0.48 & 0.53 & 0.71 & 0.77 \\
wap & 0.44 & 0.44 & 0.63 & 0.70 & war & 0.48 & 0.60 & 0.77 & 0.82 & way & 0.42 & 0.43 & 0.62 & 0.69 \\
wbm & 0.47 & 0.45 & 0.64 & 0.71 & wbp & 0.45 & 0.29 & 0.47 & 0.55 & wca & 0.41 & 0.33 & 0.52 & 0.60 \\
wer & 0.43 & 0.38 & 0.57 & 0.65 & whk & 0.46 & 0.49 & 0.67 & 0.74 & wiu & 0.44 & 0.36 & 0.55 & 0.63 \\
wmw & 0.49 & 0.55 & 0.72 & 0.78 & wnc & 0.40 & 0.34 & 0.53 & 0.60 & wnu & 0.41 & 0.29 & 0.48 & 0.57 \\
wob & 0.43 & 0.37 & 0.56 & 0.64 & wol & 0.45 & 0.52 & 0.69 & 0.76 & wos & 0.40 & 0.34 & 0.53 & 0.60 \\
wrs & 0.43 & 0.39 & 0.58 & 0.65 & wsk & 0.43 & 0.35 & 0.54 & 0.62 & wuv & 0.43 & 0.51 & 0.70 & 0.76 \\
wwa & 0.44 & 0.46 & 0.65 & 0.72 & xal & 0.46 & 0.49 & 0.67 & 0.74 & xav & 0.45 & 0.33 & 0.53 & 0.61 \\
xbr & 0.45 & 0.54 & 0.72 & 0.78 & xed & 0.43 & 0.49 & 0.68 & 0.75 & xho & 0.47 & 0.55 & 0.72 & 0.78 \\
xla & 0.42 & 0.37 & 0.56 & 0.64 & xon & 0.44 & 0.49 & 0.68 & 0.74 & xrb & 0.42 & 0.36 & 0.56 & 0.64 \\
xsi & 0.43 & 0.37 & 0.56 & 0.65 & xsm & 0.44 & 0.43 & 0.63 & 0.71 & xsu & 0.43 & 0.33 & 0.52 & 0.60 \\
xtd & 0.43 & 0.35 & 0.54 & 0.62 & xtm & 0.40 & 0.39 & 0.59 & 0.66 & xuo & 0.42 & 0.41 & 0.60 & 0.68 \\
yaa & 0.44 & 0.33 & 0.51 & 0.58 & yad & 0.45 & 0.42 & 0.60 & 0.67 & yal & 0.47 & 0.55 & 0.72 & 0.78 \\
yam & 0.42 & 0.39 & 0.59 & 0.67 & yan & 0.45 & 0.39 & 0.59 & 0.66 & yaq & 0.44 & 0.40 & 0.59 & 0.66 \\
yby & 0.43 & 0.34 & 0.53 & 0.61 & ycn & 0.44 & 0.32 & 0.51 & 0.59 & yle & 0.40 & 0.25 & 0.43 & 0.51 \\
yli & 0.42 & 0.35 & 0.54 & 0.62 & yml & 0.42 & 0.41 & 0.60 & 0.67 & yon & 0.40 & 0.44 & 0.63 & 0.70 \\
yor & 0.45 & 0.50 & 0.67 & 0.74 & yrb & 0.44 & 0.31 & 0.50 & 0.58 & yre & 0.45 & 0.36 & 0.56 & 0.64 \\
yss & 0.41 & 0.33 & 0.52 & 0.60 & yua & 0.46 & 0.54 & 0.72 & 0.78 & yuj & 0.38 & 0.39 & 0.57 & 0.65 \\
yut & 0.43 & 0.43 & 0.63 & 0.70 & yuw & 0.43 & 0.41 & 0.61 & 0.68 & yuz & 0.40 & 0.40 & 0.58 & 0.65 \\
yva & 0.47 & 0.45 & 0.65 & 0.72 & zaa & 0.41 & 0.40 & 0.61 & 0.68 & zab & 0.44 & 0.46 & 0.64 & 0.71 \\
zac & 0.40 & 0.42 & 0.62 & 0.69 & zad & 0.40 & 0.45 & 0.64 & 0.71 & zae & 0.45 & 0.45 & 0.64 & 0.71 \\
zai & 0.45 & 0.42 & 0.62 & 0.69 & zam & 0.43 & 0.35 & 0.55 & 0.63 & zao & 0.43 & 0.44 & 0.63 & 0.70 \\
zar & 0.43 & 0.47 & 0.66 & 0.73 & zas & 0.40 & 0.44 & 0.62 & 0.69 & zat & 0.44 & 0.46 & 0.65 & 0.72 \\
zav & 0.41 & 0.38 & 0.57 & 0.65 & zaw & 0.43 & 0.46 & 0.65 & 0.72 & zca & 0.44 & 0.36 & 0.55 & 0.63 \\
zho & 0.49 & 0.60 & 0.77 & 0.82 & zia & 0.43 & 0.35 & 0.54 & 0.62 & ziw & 0.50 & 0.56 & 0.73 & 0.79 \\
zom & 0.48 & 0.48 & 0.66 & 0.73 & zos & 0.44 & 0.44 & 0.62 & 0.69 & zpc & 0.41 & 0.39 & 0.58 & 0.65 \\
zpi & 0.45 & 0.47 & 0.66 & 0.73 & zpl & 0.45 & 0.41 & 0.60 & 0.68 & zpm & 0.42 & 0.34 & 0.53 & 0.60 \\
zpo & 0.46 & 0.41 & 0.60 & 0.68 & zpq & 0.37 & 0.39 & 0.58 & 0.66 & zpt & 0.43 & 0.43 & 0.63 & 0.71 \\
zpu & 0.43 & 0.38 & 0.57 & 0.65 & zpv & 0.43 & 0.40 & 0.59 & 0.67 & zpz & 0.46 & 0.38 & 0.57 & 0.65 \\
zsm & 0.47 & 0.56 & 0.73 & 0.79 & zsr & 0.46 & 0.48 & 0.67 & 0.73 & ztq & 0.39 & 0.43 & 0.63 & 0.71 \\
zty & 0.45 & 0.45 & 0.64 & 0.71 & zul & 0.46 & 0.55 & 0.72 & 0.78 & zyp & 0.45 & 0.49 & 0.68 & 0.74 \\
\bottomrule
\end{tabular}
    }
    \caption{Transfer performance using other languages as the train/query language (Part V).}\label{tab:table5}
\end{table*}

\end{document}